\documentclass{article} % For LaTeX2e
\usepackage[final,preprint]{neurips_2026}

\usepackage{amssymb}
\usepackage[utf8]{inputenc} % allow utf-8 input
\usepackage[T1]{fontenc}    % use 8-bit T1 fonts
\usepackage{hyperref}       % hyperlinks
\usepackage{url}            % simple URL typesetting
\usepackage{booktabs}       % professional-quality tables
\usepackage{amsfonts}       % blackboard math symbols
\usepackage{nicefrac}       % compact symbols for 1/2, etc.
\usepackage{microtype}      % microtypography
\usepackage{xcolor}         % colors
\usepackage{graphicx}         % colors
\usepackage[export]{adjustbox} % in preamble
\usepackage{multirow}
\usepackage{tcolorbox}
\usepackage{xcolor}
\usepackage{xspace}
\usepackage[shortlabels]{enumitem}
\usepackage{wrapfig}

\newcommand{\olmotwo}{\textsc{OLMo2}\xspace}
\newcommand{\llama}{\textsc{Llama}\xspace}
\newcommand{\llamathreetwo}{\textsc{Llama-3.2}\xspace}
\newcommand{\llamathreeone}{\textsc{Llama-3.1}\xspace}
\newcommand{\llamathreeonetwo}{\textsc{Llama-3.1/3.2}\xspace}
\newcommand{\gemmathree}{\textsc{Gemma-3}\xspace}

\newcommand{\olmotwooneb}{\textsc{OLMo2}-1B\xspace}
\newcommand{\olmotwosevenb}{\textsc{OLMo2-7B}\xspace}
\newcommand{\llamathreetwooneb}{\textsc{Llama-3.2}-1B\xspace}
\newcommand{\llamathreetwothreeb}{\textsc{Llama-3.2}-3B\xspace}
\newcommand{\llamathreeoneeightb}{\textsc{Llama-3.1}-8B\xspace}
\newcommand{\gemmathreeoneb}{\textsc{Gemma-3}-1B\xspace}

\title{Time-Incremental Continued Pretraining of LLMs: Knowledge Updates Without Catastrophic Forgetting}

\author{%
  \textbf{Fırat Öncel}$^{1,3}$ \quad
  \textbf{Salman Hussain Ali}$^{2,3}$ \\
  \textbf{Mirco Ravanelli}$^{1,3}$ \quad
  \textbf{Cem Subakan}$^{1,3,4}$ \quad
  \textbf{Çağatay Yıldız}$^{5,6}$ \\
    \normalfont
  $^{1}$Concordia University \quad
  $^{2}$Université de Montréal \quad
  $^{3}$Mila -- Quebec AI Institute \\
  $^{4}$Laval University \quad
  $^{5}$University of Tübingen \quad
  $^{6}$Tübingen AI Center
}

\begin{document}

\maketitle
\vspace{-1.5em}
\begin{abstract}

Large language models (LLMs) drift out of date the moment their pretraining ends, yet retraining from scratch is prohibitively expensive. Continued pretraining (CPT) is the natural remedy, but it is typically evaluated through a continual learning lens that assumes disjoint data streams. This is a poor fit for time-incremental updates on web-scale crawls, where successive snapshots share substantial URL overlap by design. We study time-incremental CPT in this realistic regime: continued pretraining on FineWeb-Edu dumps drawn strictly from after each model's knowledge cutoff, evaluated across six open-weight models spanning three families (\olmotwo, \llamathreeonetwo, \gemmathreeoneb) and four parameter scales (1B-3B-7B–8B).\\
We organize our findings around four practical questions. \textit{(i) Is knowledge acquired?} Yes, but heterogeneously, and without catastrophic forgetting: five of six models also improve on pre-cutoff factual recall, and the gains track pretraining saturation (driven primarily by token budget per parameters). \textit{(ii) What does it cost?} Almost nothing: the macro-average across a thirteen-task suite stays within $0.01$ of the base for every model \textit{(iii) What is the recipe?} Data quality dominates quantity (a curated 6B-token slice matches a 40B broader one); the optima for knowledge acquisition and general capability are separated by roughly an order of magnitude in learning rate; and LoRA at sufficient rank matches full CPT. \textit{(iv) Does it survive deployment?} CPT gains transfer through SFT, while DPO's effect is family-dependent. Together, these results paint an optimistic picture of time-incremental CPT than the prior continual learning literature suggests.
\begin{center}
\begin{tcolorbox}[colframe=blue!60!black, colback=blue!5, boxrule=0.8pt, arc=4pt, left=8pt, right=8pt, top=4pt, bottom=4pt, width=\linewidth]
\centering
\href{https://llm-cpt.github.io/Time-Incremental-Continued-Pretraining-of-LLMs/}{\textbf{\color{blue!60!black} Project Page}}
\end{tcolorbox}
\end{center}

\end{abstract}

\section{Introduction}
\vspace{-.5em}

Large language models (LLMs) are pretrained on internet-scale corpora, much of which is scraped from over a decade of Common Crawl dumps \citep{commoncrawl}, and subsequently refined via supervised fine-tuning (SFT) and reinforcement learning (RL) before deployment. Yet the web on which they were trained does not stand still: news, scientific literature, software, regulations, and cultural discourse continue to evolve every day. This creates a \emph{temporal knowledge gap}, where models begin to drift out of date once their training ends, and their performance is known to deteriorate on data postdating their cutoff \citep{cheng2024knowledge, dai2025dailyoracle}. Combined with the prohibitive cost of retraining frontier models from scratch, this raises a natural and pressing question:
\textbf{do LLMs efficiently and reliably absorb new knowledge as it emerges without losing existing knowledge and abilities?}

The field of continual learning offers the natural framing for this problem: equipping a model with new knowledge over time without retraining from scratch. The central concern in this literature is \emph{catastrophic forgetting}, whereby acquiring new information overwrites previously learned capabilities. Early continual learning research operated at small scales \citep{li2017learning, zenke_continual_2017, shin_continual_2017, rebuffi_icarl_2017, lopez-paz_gradient_2017, nguyen_variational_2017, rusu_progressive_2016, kirkpatrick2017ewc} and only recently has begun to engage with modern LLM regimes. Even within the recent literature, conclusions diverge sharply: reported levels of forgetting and knowledge acquisition vary substantially across studies, in part because most prior work focuses on single domains such as Wikipedia, news, or social media \citep{jang2022, liska2022, luu2022}, and the few web-scale efforts \citep{gururangan2020dont, ibrahim2024simple, gupta2023rewarm, yildiz2024investigating} do not always report the same metrics on the same downstream tasks.
Importantly, much of this literature was shaped by an assumption of \emph{disjoint, incremental data streams}. This setting is quite unlike the realistic regime in which a practitioner/researcher would actually update a deployed LLM since modern web-scale curation pipelines, such as Common Crawl, retain substantial cross-snapshot URL overlap.

In this work, we study \emph{time-incremental continued pretraining (CPT)}:
continued pretraining of LLMs on data drawn strictly from \emph{after} their pretraining cutoff. This setup is informative on two counts. First, the post-cutoff data necessarily contains novel information, letting us quantify whether new knowledge is acquired. 
Second, the inevitable overlap between post-cutoff crawls and the original pretraining distribution contrasts with the typical continual learning assumption of fully disjoint data streams, and better reflects the conditions under which LLMs would realistically be updated over time. Furthermore, for almost all open-weight LLMs the pretraining corpus is closed, so one cannot enumerate which content in a fresh dump is genuinely novel except by timestamp — making strictly disjoint CPT operationally infeasible in practice. To this end, we continue pretraining six models spanning three architectural families and scales, using two datasets of varying size and quality sourced exclusively from beyond each model's knowledge cutoff. We then evaluate the resulting models on \textit{(i)} questions extracted from daily news, which probes both retention of pre-cutoff facts and acquisition of post-cutoff ones, and \textit{(ii)} a broad suite of standard LLM benchmarks, which gauges whether general capabilities are preserved. We organise our findings around four practical questions:

\textbf{Is knowledge acquired?} (Sec~\ref{sec:findings-know-acq}). Across three model families, CPT consistently absorbs new knowledge, most prominently for \olmotwo and \gemmathreeoneb, while gains within the \llama family remain limited. Critically, accuracy on \emph{before-cutoff} questions also improves for five of six models, indicating no catastrophic forgetting of existing knowledge. The advantage grows for more recent events and is preserved on questions whose source articles are not in the training corpus.

\textbf{What does CPT cost?} (Sec~\ref{sec:findings-general-abilities}). Macro-averaged accuracy across thirteen downstream tasks remains within $0.01$ of the base for every model, with eleven of thirteen tasks staying within $\pm0.02$. So, catastrophic forgetting of general capabilities is not the dominant failure mode of CPT at this scale.

\textbf{What is the learning recipe?} (Sec~\ref{sec:findings-recipe}). We discover three relevant findings: data quality dominates quantity (a curated 6B-token slice captures most of the available signal); a single learning rate cannot jointly optimize knowledge acquisition and general capability, with the two optima separated by roughly an order of magnitude; and low rank adaptation (LoRA) at sufficient rank matches full CPT while saving substantial GPU memory. We also notice that performance plateaus within the first third of training, suggesting short and periodically repeated runs.

\textbf{Does the recipe survive deployment?} (Sec~\ref{sec:findings-deployment}). CPT gains transfer robustly through subsequent SFT. DPO's interaction with CPT is family-dependent — partially eroding gains in \llamathreeoneeightb (reminiscent of the alignment-tax phenomenon) but improving them in \olmotwosevenb.

Taken together, our results paint a more optimistic picture of continued pretraining than prior continual learning literature, which was shaped largely by disjoint-data-stream assumptions, would suggest. In the realistic time-incremental setting we study, forgetting is small, and the cost-benefit tradeoff is governed by data quality. We close with a concrete operational recipe for time-incremental updates.

\vspace{-.5em}
\section{Methodology}
\label{sec:method}
\vspace{-.75em}
In this section, we explain the methods we consider, the pretraining data and evaluation setup.

\begin{table}[t]
    \centering
    \caption{Summary of our experiments}
    \label{tab:summary}
    \begin{tabular}{p{0.16\textwidth} | p{0.76\textwidth}}
        \toprule
         Models & \olmotwooneb, \olmotwosevenb, \llamathreetwooneb, \llamathreetwothreeb, \llamathreeoneeightb, \gemmathreeoneb  \\ \midrule
         Pretraining data & FineWeb-Edu \texttt{score} $\geq\!4$ (6B tokens) and \texttt{score} $\geq\!3.5$ (40B tokens) \\ \midrule
         Data window & Post-cutoff dumps (CC week 10/2024 -- week 26/2025; beginning shifted to mid-August 2024 for \gemmathreeoneb) \\ \midrule
         Evaluation set & Daily Oracle (True/False \& Multiple Choice, before/after cutoff)  \\ \midrule
         Downstream tasks & PIQA, HellaSwag, WinoGrande, OpenBookQA, BoolQ, SciQ, ARC-Easy, ARC-Challenge, COPA, CommonsenseQA, SocialIQA, Basic Arith., MMLU \\ \midrule
         Learning rates & 2--9 rates per model; linear warmup (500 steps), linear decay to $0.1\times$ peak \\ \midrule
         Post-training & Tulu 3 dataset (SFT and DPO) \\ \midrule
         Hardware & Single node, 8$\times$H200 GPUs \\
         \bottomrule
    \end{tabular}
    % \vspace{-.5em}
\end{table}

\textbf{Models. } We continue pretraining six open-weight models with known knowledge cutoff dates, spanning three families and multiple parameter scales: \olmotwo~(1B, 7B)~\citep{olmo2}, \llamathreetwo~(1B, 3B)~\citep{grattafiori2024llama3}, \llamathreeone\citep{grattafiori2024llama3}, and \gemmathreeoneb \citep{gemma3}. This selection is deliberate: \olmotwo provides full data transparency; the \llamathreeone and \llamathreetwo families share architectural line but differ in training-token budget and scale, isolating the effect of scale within a single regime; \gemmathreeoneb adds an independent family with a different cutoff (August 2024 vs. December 2023 for the others). The selected models also allow us compare different model sizes (1B vs 3B vs 7-8B) with multiple models falling into the smallest and largest groups. The original pretraining-token budgets vary considerably across families --- \gemmathreeoneb at $\sim$2T tokens, \olmotwo at $\sim$4.2T, \llamathreetwo at up to 9T, and \llamathreeoneeightb at $\sim$15T --- a fact we will return to in Section~\ref{sec:findings-know-acq}.

\paragraph{Continued pre-training data.} 
We use FineWeb-Edu~\citep{lozhkov2024fineweb-edu} dumps for pretraining, which is a 1.3T token educational subset of FineWeb~\citep{penedo2024the}. ``Educational content'' is defined as content that would genuinely help a reader learn a topic, as opposed to commercial, entertainment, or purely navigational pages. In the following, we first describe how FineWeb is constructed, and then specify our slices.

FineWeb \citep{penedo2024the} is a 15T-token English web corpus derived from 96 Common Crawl (CC) dumps. Each dump is a sample of the web, and is designed to be a representative and diverse sample on its own. These dumps were processed  through URL-based deduplication, heuristic quality filters, and MinHash deduplication. Within each dump, near-duplicate document pairs are flagged using 5-gram signatures, followed by a transitive clustering step that keeps one document per cluster. 
% An additional filter removes documents with anomalously high ratios of repeated lines. 
Importantly, deduplication is applied \emph{per snapshot} rather than globally. \citet{penedo2024the} report that cross-snapshot deduplication degraded downstream performance: the globally deduplicated data was in fact of \emph{lower} quality (including more ads, incoherent keyword lists, poorly formatted text) than the duplicate part.
FineWeb-Edu, as well as our pretraining sets, therefore, retains cross-dump duplicates by design.
Concretely, roughly $30$--$40\%$ of URLs in the dumps we use are URLs that had never appeared in any prior crawl, with the remainder being revisits of previously-seen URLs (whose content may or may not have changed since) \footnote{Per-dump URL-status statistics are reported by Common Crawl at \href{https://commoncrawl.github.io/cc-crawl-statistics/plots/crawlsize}{this link}.}.
This property differs from the usual continual learning scenarios, where the data is typically completely new.
This is a deliberate property of the corpus, not an artifact, and it mirrors the regime a practitioner updating their model on fresh web data would actually face.

The pretraining data for each model is drawn from strictly post-cutoff FineWeb-Edu dumps relative to each model's original pre-training cutoff. For all models except \gemmathreeoneb, we use dumps from  week 10 of 2024, i.e.,\ early March, through the latest available checkpoint at the time of our experiments (week 26 of 2025). For \gemmathreeoneb, we shift the starting dump forward to mid-August 2024 to preserve the post-cutoff guarantee. We experiment with two quality-filtered slices of this window: a high-quality slice (\texttt{score} $\geq\!4$, 6B tokens) and a broader slice (\texttt{score} $\geq\!3.5$, 40B tokens), allowing us to disentangle the effect of corpus quality from that of corpus size.

\paragraph{Evaluation protocol and metrics.}
We evaluate the checkpoints on two benchmarks. The first Daily Oracle benchmark~\citep{dai2025dailyoracle} harvests question-answer (QA) pairs from daily news. The QA pairs are
generated daily, and consist of True/False (TF) and
Multiple Choice (MC) questions across various categories
such as business, politics, and arts. We partition these questions into \emph{before-cutoff} and \emph{after-cutoff} sets with respect to each model's own pre-training cutoff. We follow the OLMES framework~\citep{gu2025olmes} for model evaluation, which provides robust, meaningful comparisons across a wide range of models and tasks. We report both F1 scores and accuracies, accounting for class imbalance in the True/False split and the differing chance-level baselines of the two QA formats. Second, to assess models' downstream performance, we test them on the tasks at the \olmotwo repository (PIQA, HellaSwag, WinoGrande, OpenBookQA, BoolQ, SciQ, ARC-Easy, ARC-Challenge, COPA, CommonsenseQA, SocialIQA, Basic Arithmetic, MMLU)~\citep{olmo2}.

Additionally, we report sample-level \emph{acquisition} and \emph{forgetting}. For a given question set, acquisition is the fraction of questions the base model answers incorrectly and the CPT model answers correctly (wrong~$\to$~right), and forgetting is the fraction answered correctly before but incorrectly after CPT (right~$\to$~wrong). The two are defined on the same items, so their difference is exactly the accuracy change: any aggregate gain decomposes as acquisition minus forgetting. Evaluated on the before-cutoff split, forgetting quantifies loss of knowledge the model already had; on the after-cutoff split, acquisition quantifies uptake of genuinely new knowledge.

\paragraph{Training details.}
For each model, we sweep over a range of learning rates and report results with the best-performing setting. The breadth of the sweep varies with model size: larger models are much more expensive to continue pretraining, hence we conduct fewer runs at the larger scales.
Concretely, for each model, we experiment with 2 to 9 different learning rates.
The full per-model grid is reported in Appendix~\ref{tab:results-long}. The learning rate is linearly warmed up from zero to its peak value over the first 500 steps and then linear decayed to one-tenth of the peak by the end of training~\citep{olmo2}. We train the \olmotwo models with their official codebase and the remaining models with the HuggingFace \texttt{Trainer}~\citep{wolf2020transformers}, in both cases keeping optimizer and tokenizer choices aligned with each model's original pre-training setup. We further post-train SFT and DPO \llamathreeoneeightb and \olmotwosevenb models on Tulu 3~\citep{lambert2024tulu} using the original recipe and codebase. All experiments are run on a single node with 8$\times$H200 GPUs. We will release all checkpoints and evaluation outputs upon acceptance.

\vspace{-.5em}
\section{Findings}
\label{sec:findings}
\vspace{-.5em}

We organize this section around four practical questions: \emph{is knowledge acquired?} (Section~\ref{sec:findings-know-acq}), \emph{what does it cost?} (Section~\ref{sec:findings-general-abilities}), \emph{what is the learning recipe?} (Section~\ref{sec:findings-recipe}), and \emph{does the recipe survive deployment?} (Section~\ref{sec:findings-deployment}). Before discussing the findings, we briefly clarify how results are organized across the main text and the appendix.

\textbf{Macro-F1 vs. accuracy.} Both the TF and MC splits are class-imbalanced; therefore, accuracy disproportionately reflects performance on dominant classes. We therefore report Macro-F1 (the unweighted mean of per-class F1). Every figure in this section is accompanied by an accuracy plot in the appendix; model rankings and all qualitative conclusions are consistent across both metrics.

\textbf{Tables and figures.} Tables~\ref{tab:results-long} and~\ref{tab:results-long-acc} report the full Daily Oracle results (Macro-F1 and accuracy) for every model and learning-rate configuration we trained. Table~\ref{tab:downstream-results} reports the full thirteen-task downstream evaluation. Most downstream numbers move little under CPT. Per-class sample counts for the before- and after-cutoff TF and MC splits are reported in Appendix~\ref{app:split-stats}.

\textbf{Run-to-run variability.} To address run-to-run variability, we retrained \llamathreetwothreeb and \olmotwooneb. Resulting after-/before-cutoff Macro-F1 changes are \llamathreetwothreeb ($+0.015$, $-0.003$) and \olmotwooneb ($-0.012$, $+0.002$). These numbers are below the effect sizes reported in Section~\ref{sec:findings-know-acq}, indicating stable conclusions across re-runs.

%%%%%%%%%%%%%%%%%%%%%%%%%%%%%%%%%%%%%%%%%%%%%%%%%%%%%%%%%%
% \vspace{-0.5em}
\subsection{Heterogeneous knowledge acquisition without catastrophic forgetting}
\label{sec:findings-know-acq}
\vspace{-.75em}
\begin{figure}[h]
    \centering
    \includegraphics[width=0.49\linewidth]{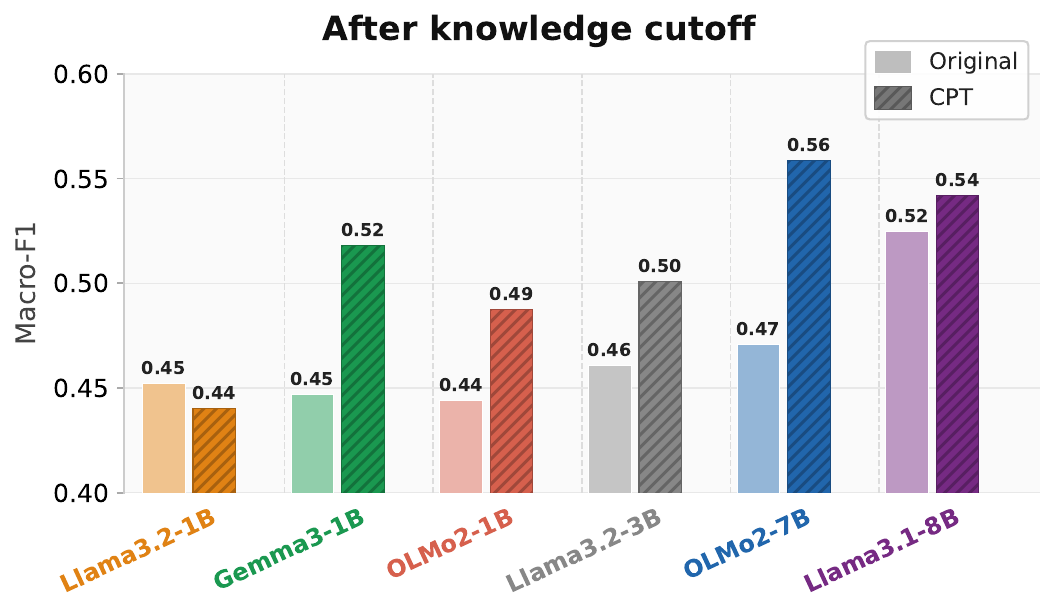}
    \hfill
    \includegraphics[width=0.49\linewidth]{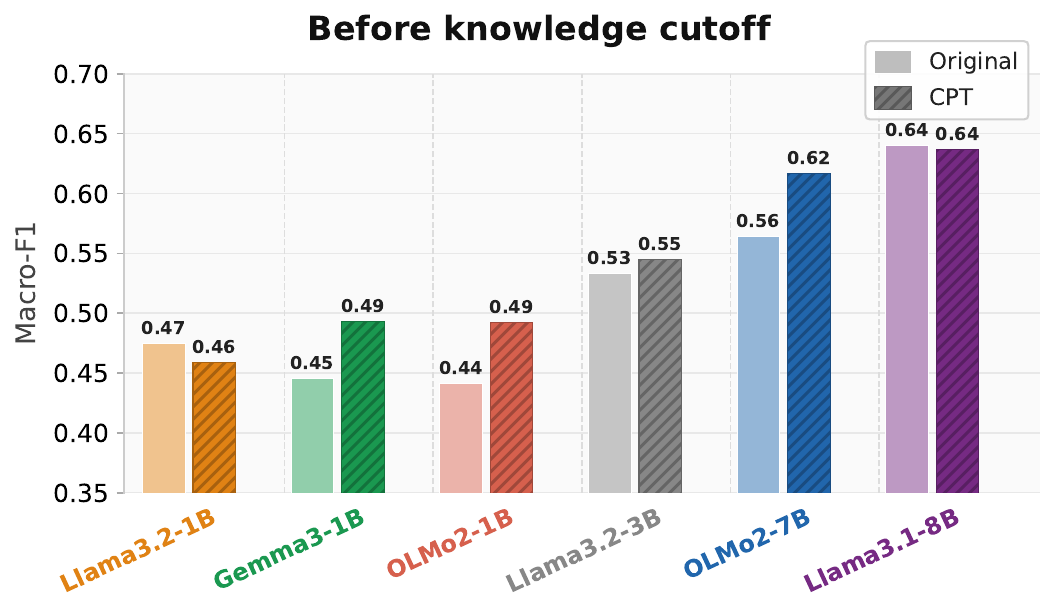}
    % \caption{Knowledge acquisition after CPT varies significantly across model families. Macro-F1 scores on questions from Daily Oracle that come after each model's knowledge cutoff show that \olmotwo and \gemmathreeoneb models achieve substantial improvements, while \llama models exhibit limited gains.}
    \caption{Macro-F1 on Daily Oracle questions split by each model's knowledge cutoff. \textbf{Left}: After-cutoff gains track the original token budget — \olmotwosevenb and \gemmathreeoneb absorb substantially more new knowledge than the more heavily pretrained \llama variants. \textbf{Right}: Five of six models also improve on before-cutoff questions, reflecting CC's re-crawl overlap with the original pretraining distribution rather than catastrophic forgetting.}
    \label{fig:main-avg}
\end{figure}

% Figure~\ref{fig:main-avg} shows Macro-F1 averaged across question types on the \emph{after-} and \emph{before-cutoff} partitions, probing genuine knowledge acquisition and forgetting respectively.

\textbf{After-cutoff knowledge acquisition is heterogeneous.}
On the \emph{after-cutoff} questions, we observe a striking spread across families: \olmotwosevenb and \gemmathreeoneb improve by $+0.09$ and $+0.07$ Macro-F1 respectively, whereas \llamathreetwooneb in fact \emph{loses} a small amount and the other \llama models gain less. This heterogeneity follows the original pretraining-token budget (with \llamathreeoneeightb being one exception): \gemmathreeoneb on $\sim$2T tokens, \olmotwo (1B and 7B) on $\sim$4.2T, while the more saturated \llamathreetwo family was trained on up to $9$T. Appendix~\ref{app:saturation} orders the six checkpoints by tokens-per-parameter, the after-cutoff gains fall monotonically once \llamathreeoneeightb is set aside (rank correlation $\rho=-0.63$). Plasticity therefore tracks how far past compute-optimal a checkpoint was trained, not how much compute it consumed. We caution that this pattern is correlative (pretraining recipes also differ in data composition, mixture, and architecture); but it is consistent across three independent families. 

The methodological implication is that \emph{model scale alone does not determine plasticity}. Studies that benchmark continued pretraining on a single model family, particularly a heavily-trained one, risk reporting findings that do not generalise. Multi-family evaluation should be a minimum standard for the field.

\textbf{CPT does not cause forgetting, and improves pre-cutoff recall.}
The \emph{pre-cutoff} results in Figure~\ref{fig:main-avg} tell a story that goes against the catastrophic-forgetting expectation. F1 score on questions whose answers were already encoded at the original cutoff consistently \emph{increases} after CPT (only \llamathreetwo-1B regresses by $0.01$ points). We attribute this primarily to the CPT data: FineWeb-Edu inherits FineWeb's per-snapshot deduplication policy, so a substantial fraction of post-cutoff pages are recrawls of URLs that already existed before each model's knowledge cutoff, implicitly replaying pre-cutoff content in training. The practical implication is that the cost--benefit tradeoff of CPT is more favourable than a catastrophic-forgetting framing would predict: in the worst case the model neither forgets nor learns, and in the typical case it improves on \emph{both} time slices simultaneously.

\textbf{Statistical consistency.} Because the per-model effects are small, we test the claims at the population level with paired tests across the six models (base vs.\ best CPT). \textit{(i)} After-cutoff Macro-F1 improves for five of six models (mean $+0.043$), significant under a paired $t$-test ($p=0.031$). \textit{(ii)} Before-cutoff Macro-F1 improves for five of six models (mean $+0.027$), also significant ($p=0.034$). A consistent pattern therefore holds despite model-to-model variation.

\textbf{The CPT advantage widens over time.}
When Daily Oracle results are plotted by the timestamp of the source news article rather than aggregated into a single before/after split, a clear temporal trend emerges (Figure~\ref{fig:over-time} in the appendix): the Macro-F1 gap between CPT and base models grows steadily as questions approach the present. This is most pronounced for the plastic families (\olmotwo, \gemmathreeoneb), and confirms that the aggregate gains reported above understate the benefit of CPT for the most recent events --- precisely the regime where temporal alignment matters most.

\textbf{Gains are not driven by dataset memorization}. A natural concern is that CPT gains might reflect memorization of source pages rather than knowledge acquisition. Daily Oracle as released has minimal overlap with our FineWeb-Edu slices: only $\approx1\%$ of its questions are extracted from articles present in the score $\geq$ 4 training set. To probe the memorization hypothesis directly, we generated an additional question set by sampling articles known to be in our training corpus and extracting QA pairs from them using the same protocol. If gains were driven by simple recall, we would expect substantially larger CPT improvements on this in-corpus set than on Daily Oracle. Instead, Macro-F1 deltas on the two sets are comparable across model families (Figure~\ref{fig:new-questions} in the appendix). We read this as evidence that CPT gains do not stem from verbatim page memorization, though we note this test does not rule out event-level memorization from related articles covering the same news. To close that gap, we additionally evaluate on a temporal holdout built with the same protocol from pages published \emph{after} our last training dump, so no source article --- related or not --- can be in the corpus: CPT improves five of six models there (mean $+0.028$), by margins comparable to the after-cutoff gains (Appendix~\ref{app:temporal-holdout}).

\textbf{A fourth family: \textsc{Qwen3}.} We exclude \textsc{Qwen} from our main pool because our design is anchored on a documented knowledge cutoff, which \textsc{Qwen} does not publish; adopting an assumed boundary of 2023-12-31 and running \emph{no} learning-rate search, we nonetheless find that CPT improves \textsc{Qwen3}-1.7B, 4B and 8B on both the before- and after-cutoff splits while leaving each model's downstream average within $0.005$ of its base. The picture is thus consistent with the other three families, but we report it in Appendix~\ref{app:qwen} rather than the main results since the cutoff is assumed and the hyperparameters untuned.

%%%%%%%%%%%%%%%%%%%%%%%%%%%%%%%%%%%%%%%%%%%%%%%%%%%%%%%%%%
\subsection{Sample-level view: acquisition and forgetting are coupled}
\label{sec:findings-sample-level}
\vspace{-0.5em}

Average Macro-F1 can hide simultaneous forgetting and learning, so we now decompose every aggregate change into its sample-level components (Section~\ref{sec:method}). Classifying each base~$\to$~CPT transition as acquisition, forgetting, retention (both correct) or neither (both wrong) makes the reported gain decompose exactly as $\mathrm{net} = \mathrm{acq} - \mathrm{forg}$. Table~\ref{tab:flips} reports these rates for all six models, both formats and both splits.

\begin{table}[t]
    \centering
    \caption{Sample-level transitions from base to best CPT checkpoint, as a percentage of questions in each split. Each cell reports acquisition / forgetting / net, where net $=$ acq $-$ forg is exactly the accuracy change. Forgetting is substantial even where the net is clearly positive.}
    \label{tab:flips}
    \small
    \setlength{\tabcolsep}{4.5pt}
    \begin{tabular}{l cc cc}
        \toprule
        & \multicolumn{2}{c}{\textbf{True/False}} & \multicolumn{2}{c}{\textbf{Multiple Choice}} \\
        \cmidrule(lr){2-3} \cmidrule(lr){4-5}
        Model & Before-cutoff & After-cutoff & Before-cutoff & After-cutoff \\
        \midrule
        \olmotwosevenb    & 26.0 / \phantom{0}9.3 / $+16.7$ & 25.3 / 13.1 / $+12.2$ & \phantom{0}7.8 / \phantom{0}5.7 / $+2.0$ & 11.8 / 6.0 / $+5.8$ \\
        \gemmathreeoneb   & 20.7 / 13.1 / $+7.7$\phantom{0} & 20.6 / 15.6 / $+5.1$\phantom{0} & \phantom{0}2.9 / \phantom{0}4.0 / $-1.0$ & \phantom{0}6.2 / 1.5 / $+4.7$ \\
        \olmotwooneb      & 20.0 / 12.3 / $+7.7$\phantom{0} & 21.5 / 16.7 / $+4.8$\phantom{0} & 15.1 / 11.7 / $+3.4$ & 11.3 / 8.2 / $+3.2$ \\
        \llamathreetwothreeb & 13.3 / 10.4 / $+3.0$\phantom{0} & 18.1 / 13.2 / $+4.9$\phantom{0} & \phantom{0}8.2 / 10.8 / $-2.6$ & \phantom{0}8.8 / 2.4 / $+6.4$ \\
        \llamathreeoneeightb & \phantom{0}8.2 / \phantom{0}8.5 / $-0.3$\phantom{0} & 16.0 / 14.2 / $+1.9$\phantom{0} & \phantom{0}4.8 / \phantom{0}3.7 / $+1.1$ & \phantom{0}5.2 / 2.1 / $+3.1$ \\
        \llamathreetwooneb   & \phantom{0}7.7 / 10.2 / $-2.5$\phantom{0} & 10.7 / 11.8 / $-1.1$\phantom{0} & \phantom{0}1.7 / \phantom{0}1.6 / $+0.2$ & \phantom{0}2.3 / 1.1 / $+1.2$ \\
        \bottomrule
    \end{tabular}
    \vspace{-1em}
\end{table}

\textbf{Forgetting is real at the sample level.} Even the most plastic models flip $9$--$16\%$ of the before-cutoff True/False answers they originally had right, so the flat-to-positive averages of Section~\ref{sec:findings-know-acq} indicate not an \emph{absence} of forgetting but forgetting outweighed by acquisition. The net stays positive for five of six models on the before-cutoff TF split, with \llamathreetwooneb the lone exception, which is the single regressing model in Figure~\ref{fig:main-avg}. Multiple-choice transitions are smaller, and there the before-cutoff net is negative for \gemmathreeoneb and \llamathreetwothreeb, whose aggregate before-cutoff gain is therefore carried by the TF format. The ``no catastrophic forgetting'' conclusion thus survives at the level of individual changed answers, but is visibly a net rather than an absolute statement.

\textbf{Acquisition and forgetting are always coupled.} Models that flip many wrong~$\to$~right also flip more right~$\to$~wrong, and no configuration achieves large acquisition at near-zero forgetting. Pooling all $191$ (acquisition, forgetting) pairs across models, learning rates, data slices and LoRA ranks, the two rates are strongly correlated (Pearson $r=0.68$, $p=1.3\mathrm{e}{-27}$), with a best fit of $\mathrm{forg} = 0.60\,\mathrm{acq} + 0.025$ (Figure~\ref{fig:coupling}). 
% The correlation is positive within every family ($r$ from $0.46$ to $0.95$, all $p<0.02$), so it is not an artifact of pooling models of differing plasticity. 
The slope below one is what makes CPT worthwhile ($82\%$ of configurations are net-positive), but slope and intercept together say roughly $0.6$ previously-correct answers are lost per additional answer gained, and that no configuration escapes the trade-off. This is the stability--plasticity trade-off expected when facts are stored in distributed, overlapping directions \citep{geva2021transformer, elhage2022toy}: plain CPT has no term distinguishing directions that carry retained knowledge from free ones, so the displacement that writes a new association perturbs old ones. Acquisition and forgetting are then not two competing processes but one process measured on two item sets.

\begin{figure}[t]
    \centering
    \includegraphics[width=\linewidth]{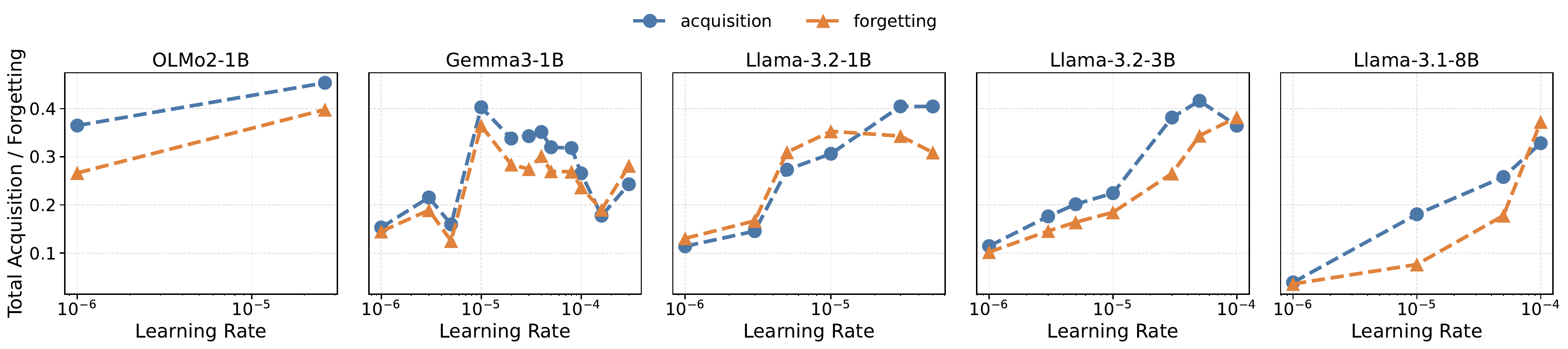}
    \caption{Total acquisition and forgetting (summed over splits and formats) as a function of peak learning rate. Both rise together with the learning rate; the useful \emph{net} is maximised at intermediate rates, since small rates buy little acquisition while large rates let forgetting catch up with (or overtake) it.}
    \label{fig:acq-forg-lr}
    \vspace{-1em}
\end{figure}

\textbf{There is a learning-rate sweet spot.} Figure~\ref{fig:acq-forg-lr} plots total acquisition and forgetting against the peak learning rate for the five models with multi-rate sweeps. Both grow with the learning rate, consistent with both being monotone in the effective update magnitude, but not at the same rate throughout: at the smallest rates the model barely moves and acquires little, while at the largest rates the forgetting curve closes on (\llamathreeoneeightb, \llamathreetwothreeb) or crosses (\gemmathreeoneb, \llamathreetwooneb) the acquisition curve. The gap between the curves is therefore widest at intermediate rates, in the same $\texttt{1e-5}$--$\texttt{5e-5}$ band identified in Section~\ref{sec:findings-lr}. 
% The trade-off is not exceptionless: the \llamathreeoneeightb \texttt{score}$\geq\!3.5$ variants at \texttt{1e-6} reach relatively high acquisition at near-zero forgetting, a high gradient signal-to-noise regime (informative data, small update) in which acquisition is bought without proportional displacement. Severe forgetting is thus a signature of an unregularised update at a high learning rate rather than an inevitable price of acquisition.

%%%%%%%%%%%%%%%%%%%%%%%%%%%%%%%%%%%%%%%%%%%%%%%%%%%%%%%%%%
\subsection{What does CPT cost? General abilities are largely preserved.}
\label{sec:findings-general-abilities}
\vspace{-0.5em}
\begin{figure}[h]
    \centering
    \includegraphics[width=.49\linewidth]{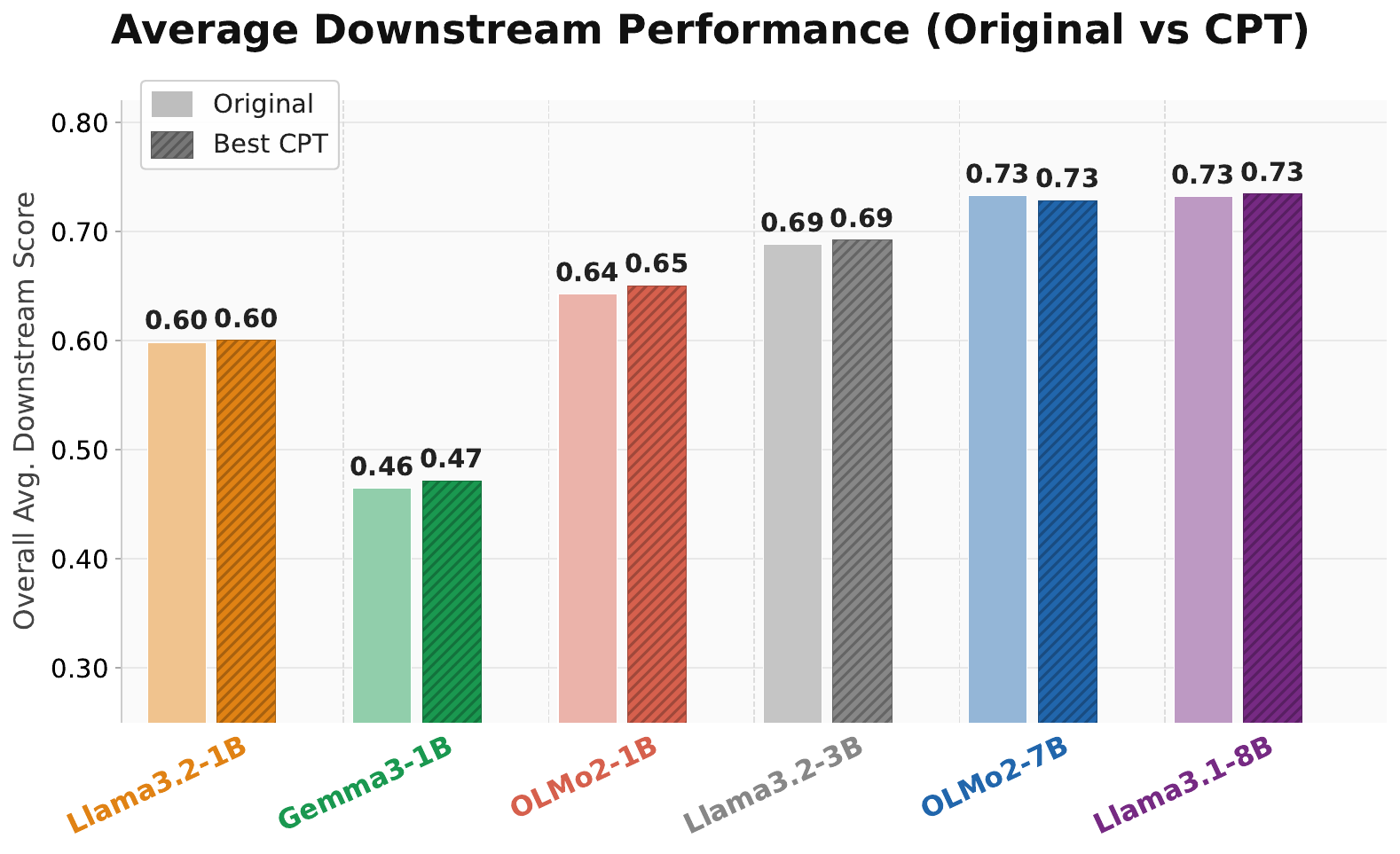} \hfill
    \includegraphics[width=.49\linewidth, trim=0 0 550 0, clip]{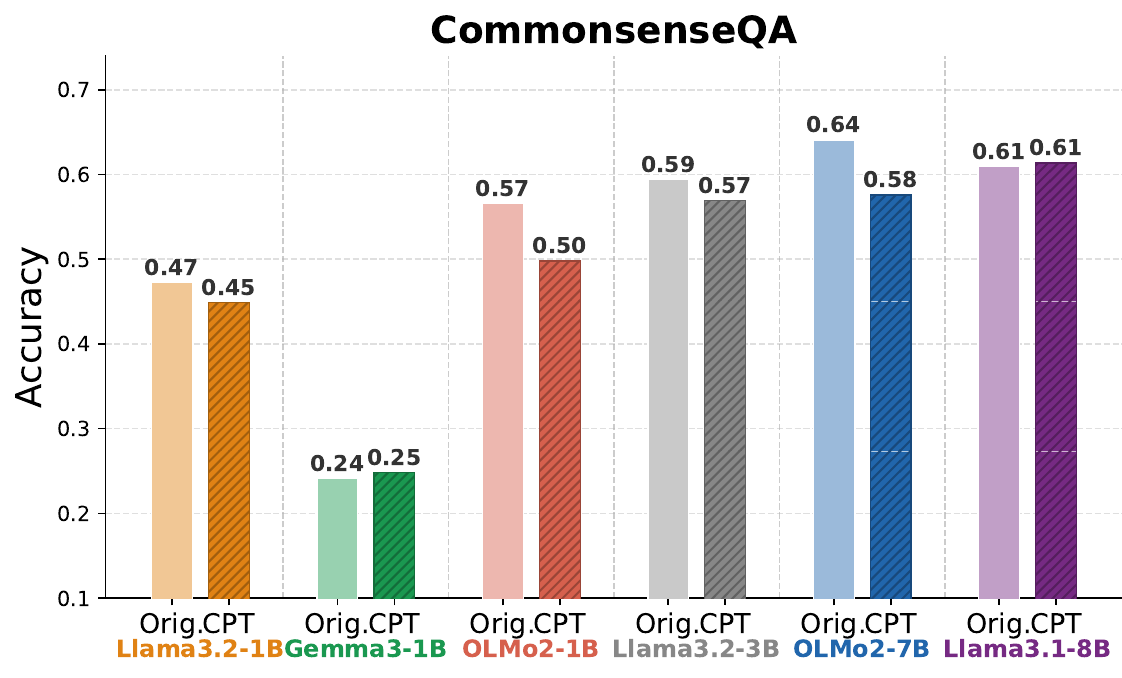}
    \caption{General capabilities are preserved under CPT. \textbf{Left:} macro-average accuracy across thirteen downstream tasks for each base model (light) and its best CPT variant (hatched); every model lands within $0.01$ of its base. \textbf{Right}: CommonsenseQA, the single benchmark on which any model loses more than $0.02$, with the largest drops concentrated on the \olmotwo family, which improves the most on Daily Oracle.}
    \label{fig:downstreams-extreme}
\end{figure}

We next ask whether continued pretraining erodes the general capabilities acquired during the original pretraining stage. Figure~\ref{fig:downstreams-extreme} (left) reports the macro-average across the thirteen downstream benchmarks for each base model and its best CPT variant. The picture is overwhelmingly stable: every model lands within $0.01$ of its base, and eleven of the thirteen individual tasks stay within $\pm 0.02$ across all models tested. A two one-sided test against a $\pm0.01$ margin gives a 90\% CI of $[-0.001, +0.007]$, inside the margin on both sides, so general capabilities are statistically \emph{equivalent} to base rather than merely indistinguishable from it. Full per-task results are in Table~\ref{tab:downstream-results}.

The two tasks that move beyond this band do so in opposite directions. ARC-Challenge improves consistently across all six models (e.g., $+0.04$ on \olmotwooneb, $+0.03$ on \llamathreetwothreeb), and is the only benchmark where CPT yields a uniform upward shift. CommonsenseQA, shown in Figure~\ref{fig:downstreams-extreme} (right), is the only benchmark on which any model loses more than two points, and the degradation is concentrated on the \olmotwo family --- precisely the family that gains the most on Daily Oracle (Section~\ref{sec:findings-know-acq}). We read this as a localised, interpretable trade-off rather than catastrophic forgetting: the models that absorb the most new factual knowledge are also the ones that pay the largest cost on commonsense reasoning. The worst-case loss is $0.06$ on a single task out of thirteen, set against substantial gains on Daily Oracle and a parallel improvement on ARC-Challenge.

%%%%%%%%%%%%%%%%%%%%%%%%%%%%%%%%%%%%%%%%%%%%%%%%%%%%%%%%%%
\subsection{What is the learning recipe?}
\label{sec:findings-recipe}
\vspace{-0.5em}

Having established that CPT acquires knowledge with no catastrophic forgetting on before-cutoff questions (Sec~\ref{sec:findings-know-acq}) and largely preserves general capabilities (Sec~\ref{sec:findings-general-abilities}), we now ask how CPT should be \emph{configured}. We isolate three important factors: data quality versus quantity, learning rate, and parameter efficiency via LoRA, and examine each in turn. We additionally note that performance plateaus within the first third of training (Figure~\ref{fig:acc-vs-checkpoints} in the appendix) for both \olmotwooneb and \llamathreeoneeightb, confirming that short, periodically-repeated runs are computationally tractable.

\subsubsection{Data quality dominates quantity}
\label{sec:findings-data-amount}
\vspace{-.5em}
\begin{figure}[h]
    \centering
    \includegraphics[width=0.995\linewidth]{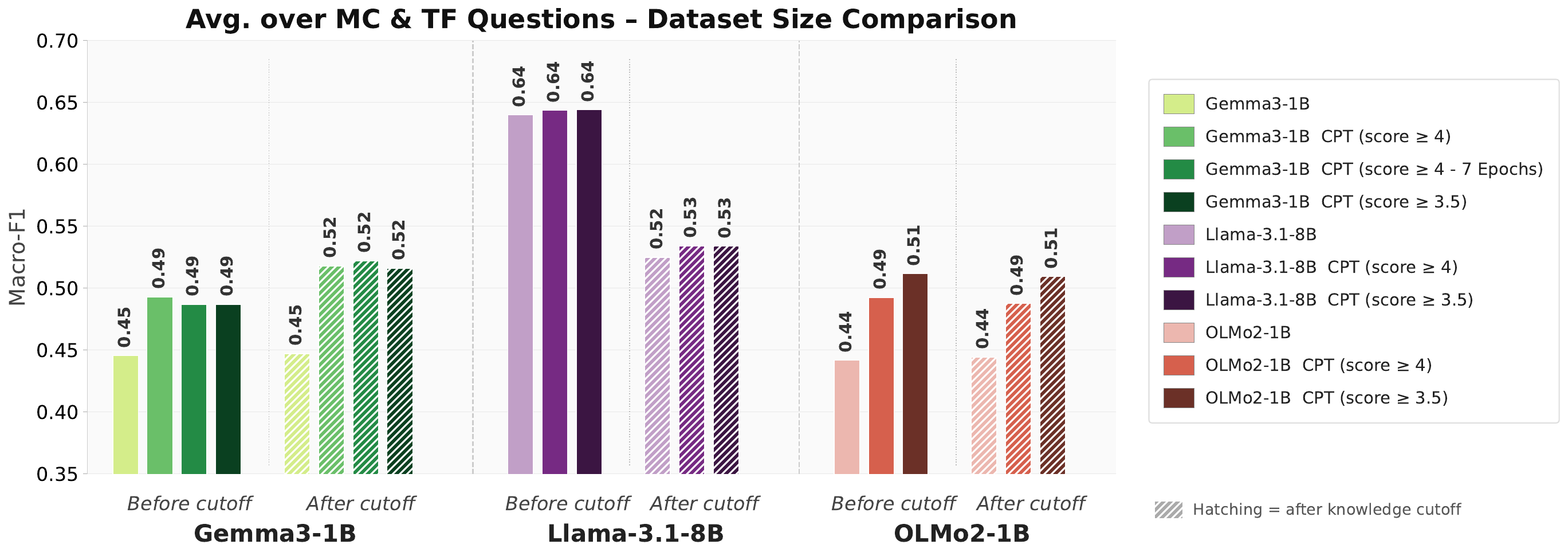} \hfill
    \caption{Data quality dominates quantity in CPT. Macro-F1 scores on Daily Oracle averaged across MC and TF questions show that a curated 6B-token high-quality slice (\texttt{score} $\geq$ 4) achieves most of the available CPT signal across all three model families. Expanding to a sevenfold larger but lower-quality 40B-token slice (\texttt{score} $\geq$ 3.5) yields only marginal additional gains, while training for 7 epochs on the high-quality slice similarly provides negligible improvement over a single pass.}
    \label{fig:data-size}
\end{figure}

Figure~\ref{fig:data-size} contrasts our two FineWeb-Edu slices across three model families: a high-quality $6$B-token slice (\texttt{score} $\geq\!4$) and a larger, lower-quality $40$B slice (\texttt{score} $\geq\!3.5$). For \llamathreeoneeightb and \gemmathreeoneb, the two slices are completely indistinguishable: both yield identical Macro-F1 scores across. Only \olmotwooneb benefits from the larger slice, improving from $0.49$ to $0.51$ on both splits --- a modest gain relative to what the high-quality slice had already contributed.

To further isolate the effect of token budget from data quality, we additionally train \gemmathreeoneb on the \texttt{score}$\geq\!4$ slice for $\approx 7$ epochs, matching the total token budget of the \texttt{score}$\geq\!3.5$ slice. The 7-epoch variant yields identical scores to the single-epoch run, confirming that additional passes over high-quality data provide no benefit once the signal is exhausted. Taken together, these results establish that data \emph{quality} dominates data \emph{quantity}: a curated $6$B-token slice captures the available CPT signal for most model families, and expanding the corpus sevenfold at lower quality yields gains only for the most plastic models, and even then only marginally.

\subsubsection{Knowledge acquisition and general capability retention require different learning rates}
\label{sec:findings-lr}
\begin{wrapfigure}[17]{r}{0.52\linewidth}
    % \vspace{-1.5em}
    \centering
    \includegraphics[width=\linewidth]{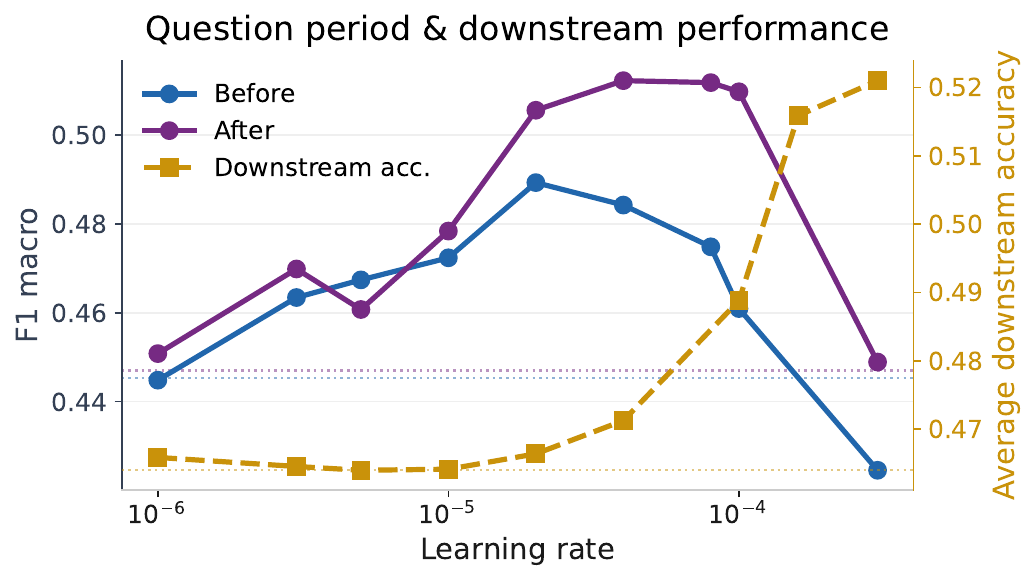}
    \vspace{-2em}
    \caption{How learning rate impacts knowledge acquisition and general capability for \textsc{Gemma-3-1B}. Daily Oracle Macro-F1 (left axis) peaks near \texttt{3e-5} and collapses at higher rates, while downstream accuracy (right) improves at rates that degrade factual recall. Dashed lines are base model performance.}
    \label{fig:lr}
    \vspace{-1em}
\end{wrapfigure}

To characterize how the learning rate influences knowledge acquisition and general capabilities, we train \gemmathreeoneb across nine learning rates (from \texttt{1e-6} to \texttt{3e-4}), and evaluate checkpoints on Daily Oracle and the downstream suite. Figure~\ref{fig:lr} summarizes the results.

The Daily Oracle curves reveal a clear monotonic trend up to a threshold: increasing the learning rate amplifies knowledge acquisition on both the after- and before-cutoff splits, with gains more pronounced on after-cutoff questions, as expected given that post-cutoff content is encountered for the first time during CPT. Performance peaks at \texttt{3e-5} and then degrades sharply at the highest rate tested, consistent with potential destabilisation from excessively large updates. The before-cutoff split follows the same trend but is more sensitive to over-shooting.

Average downstream accuracy tells a different story. Whereas QA performance peaks and collapses within $[\texttt{1e-5}, \texttt{1e-4}]$, downstream accuracy remains flat over the lower portion of that range and continues to rise even at \texttt{3e-4}, a rate that already substantially degrades Daily Oracle scores. 
% This divergence is consistent with prior observations that factual recall is mechanistically closer to memorization than downstream reasoning is \citep{wang2025generalization}, and that learning rate controls the memorization–generalization tradeoff: lower LRs commit to verbatim, episodic patterns (earlier in training), while higher LRs preferentially capture broader, generalizing structure \citep{kang2025learning}.
We attribute this divergence to a difference in what each surface measures: Daily Oracle probes retrieval of specific time-stamped facts, which is sensitive to weight perturbations that displace precise associative memories \citep{meng2022locating}; downstream benchmarks probe perhaps more distributed reasoning capacities that are harder to disrupt in a single training phase.

The practical implication is that the optimal learning rate is a function of the deployment objective. Practitioners targeting factual currency should use a moderate rate around \texttt{3e-5}; those targeting general capability gains can tolerate higher rates, but at the cost of factual precision. We recommend tuning against a held-out slice of the target evaluation surface rather than inheriting the learning rate from the original pretraining recipe. We caution that this analysis is conducted on a single model (\gemmathreeoneb); replicating the sweep on additional families is a natural extension.

\subsubsection{Low rank adaptation performs as well as full pretraining}
\label{sec:findings-lora}
\begin{wrapfigure}[21]{r}{0.5\linewidth}
    \vspace{-1.5em}
    \centering
    \includegraphics[width=\linewidth]{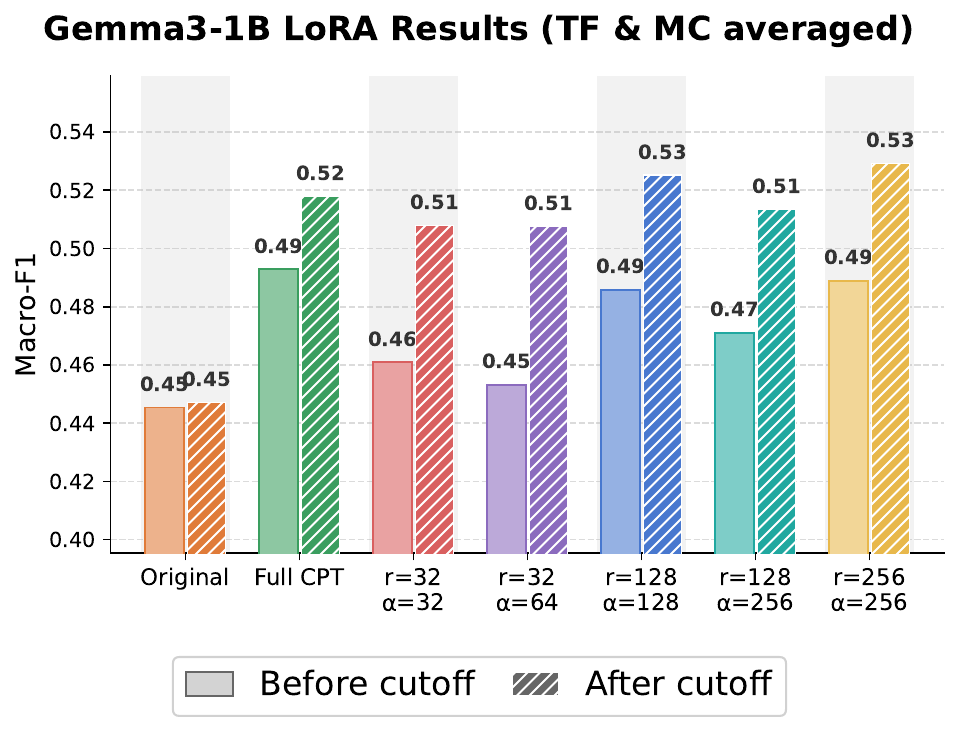}
    \caption{Low-rank adaptation (LoRA) performs comparably to full CPT. Comparison of Macro-F1 scores on Daily Oracle for models trained with LoRA versus full CPT. LoRA achieves similar knowledge acquisition with reduced computational overhead, particularly for larger models.}
    \label{fig:metric-lora}
    \vspace{-1em}
\end{wrapfigure}
Figure~\ref{fig:metric-lora} reports averaged Macro-F1 (TF and MC) on Daily Oracle for \gemmathreeoneb trained with LoRA across a range of rank--scaling configurations, alongside the original base model and its full-CPT counterpart. In LoRA, the weight updates to each layer are reparameterised as the product of two low-rank matrices of rank $r$, and $\alpha$ is a scalar that controls the magnitude of the resulting update; together, $r$ and $\alpha$ govern how much of the model's capacity is devoted to absorbing new information. We apply LoRA to both attention and MLP projection layers. The results show a clear rank-dependent trend on both splits. Both on before- and after-cutoff questions, the smallest configurations reach slightly below full CPT, while higher-rank variants match or slightly exceed full CPT. Together, these results suggest that sufficient rank is necessary both to absorb new  knowledge and to consolidate existing knowledge, but that once rank is adequate, LoRA matches full CPT on both dimensions. This conclusion is not specific to \gemmathreeoneb: repeating the experiment on \llamathreetwothreeb, LoRA at ranks $128$ and $256$ reaches after-cutoff Macro-F1 of $0.525$ and $0.534$ against $0.540$ for full CPT (base: $0.484$), with all four models within $0.530$--$0.537$ before cutoff (Appendix~\ref{app:lora-llama}).

Despite its theoretical promise of reduced computational overhead, LoRA does not yield a measurable speedup over full-model CPT in our experiments, with both approaches requiring comparable wall-clock time. This outcome aligns with the observation that LoRA, while updating only a small fraction of parameters, still requires full forward and backward passes through the frozen model. The primary advantage of LoRA in this setting is hence \emph{memory efficiency}, not training speed, making it a practical choice for scenarios where GPU memory is the constraint. For continual learning practitioners, this is an encouraging result: LoRA offers a parameter-efficient path to temporal updating that does not sacrifice knowledge acquisition and, at sufficient rank, matches the plasticity of full pretraining, while leaving the base weights untouched and thus trivially enabling rollback or multi-adapter merging.

\subsection{Does the recipe survive deployment?}
\label{sec:findings-deployment}
\vspace{-0.5em}
The previous sections established how CPT modifies \emph{base} model checkpoints. In practice, however, deployed LLMs have undergone post-training (typically SFT followed by preference optimisation such as DPO) on top of a base. We now ask whether CPT gains transfer through the post-training pipeline, or are they washed out by subsequent alignment. For this, we utilize the post-training recipes from \citep{lambert2024tulu}, and apply an identical SFT and DPO pipeline to five base checkpoints: the original \olmotwosevenb, \llamathreeoneeightb, our CPT variants trained on the standard 6B-token dataset, and finally \llamathreeoneeightb trained on the larger \texttt{score}$\geq\!3.5$ slice. The post-training data is held fixed across all three conditions, so any difference in the resulting models is attributable solely to whether CPT was performed beforehand. We follow the prompting format in \citep{dai2025dailyoracle} to have comparable results.

\begin{wrapfigure}[22]{r}{0.5\linewidth}
    \vspace{-1.0em}
    \centering
    \includegraphics[width=\linewidth]{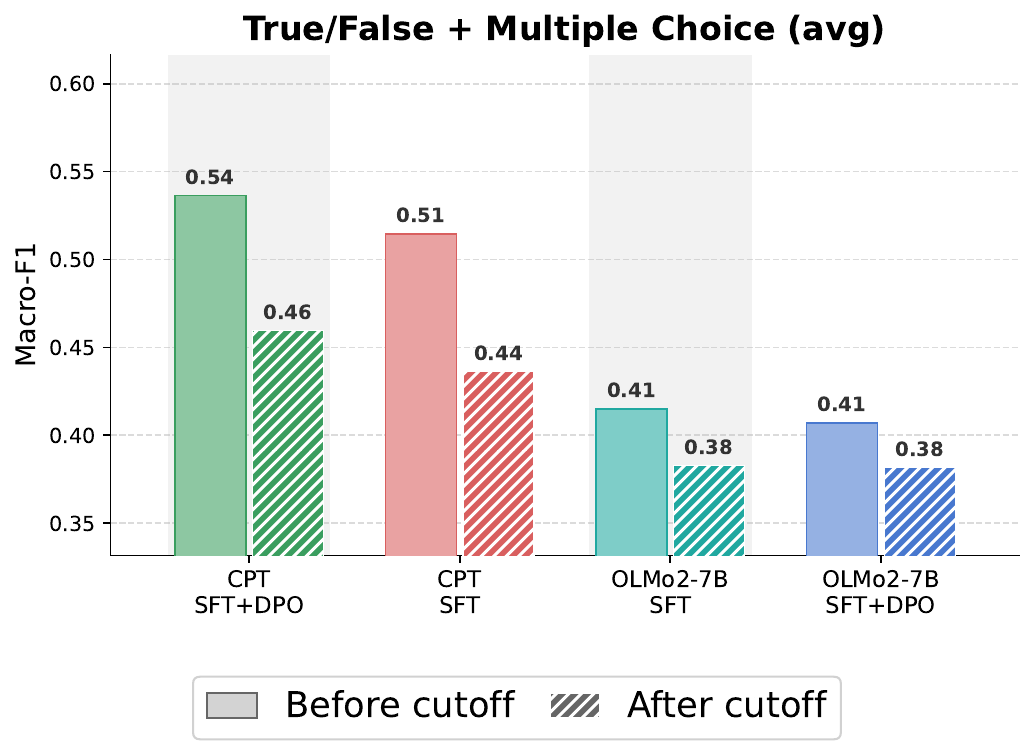}
    \caption{Macro-F1 scores on Daily Oracle for four \olmotwosevenb post-training configurations. Each group shows before-cutoff (solid) and after-cutoff (hatched) bars. CPT variants consistently outperform their non-CPT counterparts on after-cutoff questions, with DPO bringing additional benefits on top of CPT+SFT.}
    \label{fig:post-train}
    \vspace{-1em}
\end{wrapfigure}

Figure~\ref{fig:post-train} reports averaged Macro-F1 across the TF and MC splits for \olmotwosevenb. Importantly, we see that CPT knowledge is not washed out by subsequent instruction tuning: on both question portions, all CPT variants outperform the \olmotwosevenb baseline.
The same pattern holds for \llamathreeoneeightb (see Figure~\ref{fig:post-train-olmo-supp} for results). The gain is more pronounced for the checkpoints trained on the larger \texttt{score}$\geq\!3.5$ slice, suggesting that more CPT signal injected before SFT translates into more retained signal after.

A more nuanced picture emerges around DPO. For both base models, DPO has little effect when applied without CPT: \olmotwosevenb and \llamathreeoneeightb finish at the same score with or without DPO. On top of CPT, however, DPO's effect is family-dependent: it helps \olmotwosevenb (Figure~\ref{fig:post-train}) but slightly \emph{hurts} \llamathreeoneeightb. The latter direction is reminiscent of the \emph{alignment-tax} phenomenon, where RLHF and DPO erode SFT capabilities \citep{ouyang2022training, lin2024mitigating}. Here, it manifests in temporal recall, while the improvement in \olmotwosevenb suggests the tax is not universal. We acknowledge that with only two base models, we cannot reason about a robust family effect, and disentangling whether the divergence reflects base-model differences or interactions with the CPT signal is left to future work.

\section{Related Work}
\vspace{-.5em}
\textbf{Continued pretraining. }
Prior work has established that continued pretraining of LLMs on domain-shifted corpora can substantially improve downstream task performance~\citep{gururangan2020dont, roziere2023codellama, chen2023meditron}, but potentially at the cost of forgetting the original distribution when the new data have a different character~\citep{ibrahim2024simple, yildiz2024investigating}. Approaches to mitigate forgetting have been introduced and include careful learning-rate engineering~\citep{gupta2023rewarm, ibrahim2024simple} and moderate data replay. Additionally, smaller models ($\lesssim$1--3B parameters) were shown to exhibit the most significant rates of learning and forgetting~\citep{yildiz2024investigating}. However, the existing CPT literature and approaches focus primarily on \emph{domain} shifts (general $\to$ medical, code, finance), not on the \emph{temporal} shift that accumulates as the world evolves past a model's cutoff.

\textbf{Temporal continued pretraining. }
A line of research has emerged to better understand the problem of temporal misalignment and to propose approaches to update models. \citet{lazaridou2021mind} showed that perplexity on news and scientific text degrades systematically as the gap between training and test years grows. \citet{luu2022} demonstrated that this degradation extends to downstream tasks where temporal continued pretraining recovers only a fraction of what target-period finetuning achieves, and \citet{cheng2024knowledge} showed that the effective knowledge cutoffs of deployed LLMs often lag their reported cutoffs. \citet{jang2022} formalized the problem as \emph{continual knowledge learning}, with three goals: retaining time-invariant facts, updating outdated ones, and acquiring new ones. Methods to address these goals include conditioning LLMs on time as an explicit input~\citep{dhingra2022timeaware}, temporally aligning models to a target year through prompting and finetuning~\citep{zhao-etal-2024-set}, and scaling time-continual training to web-scale corpora through autoregressive update schedules and fixed-ratio replay~\citep{li2025ticlm}. However, they largely held the model fixed and varied the update procedure, leaving open how forgetting and knowledge acquisition vary across model families, post-training stages, and data quality regimes.

\textbf{Temporal benchmarking. }
Evaluating LLMs under a shifting data distribution requires moving beyond static benchmarks like MMLU~\citep{hendryckstest2021} and GSM8K~\citep{cobbe2021gsm8k}, lacking a temporal axis. Forecasting benchmarks address this by asking models to predict resolved events. Autocast~\citep{zou2022forecasting} and its successor Autocast++~\citep{yan2023autocast++} draw questions from forecasting tournaments such as Metaculus and Good Judgment Open, \citet{halawi2024approaching} compares LLM systems against humans, and MIRAI~\citep{ye2024mirai} extends the paradigm to agentic forecasting of international events. These benchmarks rely on curated question pools that do not continuously regenerate from the world's daily output. Daily Oracle~\citep{dai2025dailyoracle} closes this gap by sourcing the daily news and automatically generating TF and MC questions continuously. Additionally, \citet{dai2025dailyoracle} show that LLM accuracy degrades substantially from 2020 to 2024, and that the degradation accelerates after model knowledge cutoff and persists under retrieval augmentation.

\vspace{-.75em}
\section{Discussion}
\vspace{-.75em}
Our results reframe the prospect of keeping LLMs temporally aligned in a more optimistic light than the prior continual learning literature, developed largely under disjoint-data-stream assumptions, would suggest. We close by drawing out the implications and a concrete operational recipe.

\textbf{The disjoint-stream assumption does not hold at web scale.} A natural concern with our setup is that retaining cross-snapshot URL overlap is a form of replay, making the absence of catastrophic forgetting a foregone conclusion. Two structural facts argue against this. First, the training corpora of almost all LLMs are closed, so one cannot enumerate which content in a fresh dump is genuinely novel relative to the original pretraining distribution except using timestamps. Hence, constructing a \emph{strictly disjoint} corpus is infeasible in practice. Second, even when global cross-snapshot deduplication is technically possible, \citep{penedo2024the} show that it \emph{degrades} data quality. The disjoint-stream assumption that shaped earlier continual learning work is thus not a property that survives at web scale. Our findings quantify CPT under the conditions in which it would actually be deployed.

\textbf{Web-scale CPT does not cause catastrophic forgetting.} Across six models and three families, general capabilities are largely preserved — eleven of thirteen downstream benchmarks stay within $\pm 0.02$ of the base — and pre-cutoff Daily Oracle accuracy actively \emph{improves}. Given the cross-snapshot overlap discussed above, this is unsurprising: every post-cutoff dump implicitly replays a substantial fraction of the original training distribution. Catastrophic forgetting, in other words, is a property of disjoint regimes, not of CPT per se.

\textbf{Temporal plasticity tracks pretraining saturation more than scale.} All models in \llama family show similarly small gains, while \olmotwo and \gemmathreeoneb absorb more knowledge than any Llama. The most plastic models (\gemmathreeoneb at $\sim$2T, \olmotwo at $\sim$4.2T) saw substantially fewer pretraining tokens than the least plastic (\llamathreeoneeightb at $\sim$15T) — consistent with a saturation hypothesis. The methodological implication is that benchmarking CPT on a single heavily-trained family risks family-specific artefacts, and multi-family evaluation should be the minimum standard.

\textbf{Quality over quantity, and tuning the learning rate to the objective.} 
A curated 6B-token high-quality slice captures the bulk of available CPT signal; a $7 \times$ larger lower-quality slice yields only marginal gains. Separately, Daily Oracle peaks near \texttt{3e-5} while downstream accuracy keeps rising at rates that already degrade factual recall — the two optima differ by roughly an order of magnitude. The learning rate should therefore be treated as deployment-objective-specific: factual currency requires tuning against a held-out temporal slice; general-capability gains tolerate higher rates.

\textbf{An operational recipe for time-incremental updates.}
Synthesising our findings, we recommend the following recipe for keeping LLMs temporally-aligned: \textit{(i)} source CPT data from a curated, high-quality post-cutoff slice (FineWeb-Edu \texttt{score}$\geq\!4$ or equivalent); large lower-quality crawls offer poor return on compute; \textit{(ii)} train at a moderate learning rate ($\sim$\texttt{3e-5}) when factual recall is the primary objective, tuning against a held-out temporal slice; \textit{(iii) }use LoRA at rank $\geq128$ if memory-bound --- it matches full CPT on knowledge acquisition; \textit{(iv)} plan for short runs, since performance plateaus within the first third of training; and \textit{(v)} apply CPT before SFT in the post-training pipeline.
\newpage

\bibliographystyle{plainnat}

\bibliography{refs}

@inproceedings{dai2025dailyoracle, 
  title     = {Are {LLM}s Prescient? {A} Continuous Evaluation Using Daily News as the Oracle},
  author    = {Dai, Hui and Teehan, Ryan and Ren, Mengye},
  booktitle = {Proceedings of the International Conference on Machine Learning (ICML)},
  year      = {2025}
}

@inproceedings{cheng2024knowledge,
  title     = {Dated Data: Tracing Knowledge Cutoffs in Large Language Models},
  author    = {Cheng, Jeffrey and Marone, Marc and Weller, Orion and Lawrie, Dawn and Khashabi, Daniel and Van Durme, Benjamin},
  booktitle = {Conference on Language Modeling (COLM)},
  year      = {2024}
}

@inproceedings{
  penedo2024the,
  title={The FineWeb Datasets: Decanting the Web for the Finest Text Data at Scale},
  author={Guilherme Penedo and Hynek Kydl{\'\i}{\v{c}}ek and Loubna Ben allal and Anton Lozhkov and Margaret Mitchell and Colin Raffel and Leandro Von Werra and Thomas Wolf},
  booktitle={The Thirty-eight Conference on Neural Information Processing Systems Datasets and Benchmarks Track},
  year={2024},
  url={https://openreview.net/forum?id=n6SCkn2QaG}
}

@misc{lozhkov2024fineweb-edu,
    author       = { Lozhkov, Anton and Ben Allal, Loubna and von Werra, Leandro and Wolf, Thomas },  
    title        = { FineWeb-Edu: the Finest Collection of Educational Content }, 
    year         = 2024,  
    url          = { https://huggingface.co/datasets/HuggingFaceFW/fineweb-edu },  
    doi          = { 10.57967/hf/2497 },
    publisher    = { Hugging Face }
}

@inproceedings{gu2025olmes,
  title     = {{OLMES}: A Standard for Language Model Evaluations},
  author    = {Gu, Yuling and Tafjord, Oyvind and Kuehl, Bailey and Haddad, Dany and
               Dodge, Jesse and Hajishirzi, Hannaneh},
  booktitle = {Proceedings of the Annual Conference of the North American Chapter of the
               Association for Computational Linguistics (NAACL)},
  year      = {2025}
}

@article{
ibrahim2024simple,
title={Simple and Scalable Strategies to Continually Pre-train Large Language Models},
author={Adam Ibrahim and Benjamin Th{\'e}rien and Kshitij Gupta and Mats Leon Richter and Quentin Gregory Anthony and Eugene Belilovsky and Timoth{\'e}e Lesort and Irina Rish},
journal={Transactions on Machine Learning Research},
issn={2835-8856},
year={2024},
url={https://openreview.net/forum?id=DimPeeCxKO},
note={}
}

@inproceedings{gupta2023rewarm,
  title     = {Continual Pre-Training of Large Language Models: How to (Re)Warm Your Model?},
  author    = {Gupta, Kshitij and Thérien, Benjamin and Ibrahim, Adam and Richter, Mats L.
               and Anthony, Quentin and Belilovsky, Eugene and Rish, Irina and Lesort, Timothée},
  booktitle = {ICML Workshop on Efficient Systems for Foundation Models (ES-FoMo)},
  year      = {2023},
  url       = {https://arxiv.org/abs/2308.04014}
}

@article{
yildiz2024investigating,
title={Investigating Continual Pretraining in Large Language Models: Insights and Implications},
author={{\c{C}}a{\u{g}}atay Y{\i}ld{\i}z and Nishaanth Kanna Ravichandran and Nitin Sharma and Matthias Bethge and Beyza Ermis},
journal={Transactions on Machine Learning Research},
issn={2835-8856},
year={2025},
url={https://openreview.net/forum?id=aKjJoEVKgO},
note={}
}

@inproceedings{gururangan2020dont,
  title     = {Don't Stop Pretraining: Adapt Language Models to Domains and Tasks},
  author    = {Gururangan, Suchin and Marasović, Ana and Swayamdipta, Swabha and Lo, Kyle
               and Beltagy, Iz and Downey, Doug and Smith, Noah A.},
  booktitle = {Proceedings of the Annual Meeting of the Association for Computational
               Linguistics (ACL)},
  year      = {2020},
  url       = {https://arxiv.org/abs/2004.10964}
}

@inproceedings{jang2022,
  title     = {Towards Continual Knowledge Learning of Language Models},
  author    = {Jang, Joel and Ye, Seonghyeon and Yang, Sohee and Shin, Joongbo and Han, Janghoon and Kim, Gyeonghun and Choi, Stanley Jungkyu and Seo, Minjoon},
  booktitle = {International Conference on Learning Representations (ICLR)},
  year      = {2022}
}

@inproceedings{liska2022,
  title     = {{StreamingQA}: A Benchmark for Adaptation to New Knowledge over Time in Question Answering Models},
  author    = {Li{\v{s}}ka, Adam and Ko{\v{c}}isk{\'y}, Tom{\'a}{\v{s}} and Gribovskaya, Elena and Terzi, Tayfun and Sezener, Eren and Agrawal, Devang and de Masson d'Autume, Cyprien and Scholtes, Tim and Zaheer, Manzil and Young, Susannah and Gilsenan-McMahon, Ellen and Austin, Sophia and Blunsom, Phil and Lazaridou, Angeliki},
  booktitle = {International Conference on Machine Learning (ICML)},
  year      = {2022}
}

@inproceedings{luu2022,
  title     = {Time Waits for No One! Analysis and Challenges of Temporal Misalignment},
  author    = {Luu, Kelvin and Khashabi, Daniel and Gururangan, Suchin and Mandyam, Karishma and Smith, Noah A.},
  booktitle = {Proceedings of the 2022 Conference of the North American Chapter of the Association for Computational Linguistics (NAACL)},
  year      = {2022}
}

@misc{commoncrawl,
  author = {{Common Crawl Foundation}},
  title  = {Common Crawl},
  year   = {2025},
  howpublished = {\url{https://commoncrawl.org}},
  note   = {Accessed: 2026-04-30}
}

@article{grattafiori2024llama3,
  title   = {The {L}lama 3 Herd of Models},
  author  = {Grattafiori, Aaron and Dubey, Abhimanyu and Jauhri, Abhinav and Pandey, Abhinav
             and Kadian, Abhishek and Al-Dahle, Ahmad and Lesecq, Aiesha and Purujit, Aishwarya
             and others},
  journal = {arXiv preprint arXiv:2407.21783},
  year    = {2024}
}

@inproceedings{
olmo2,
title={2 {OLM}o 2 Furious},
author={Evan Pete Walsh and Luca Soldaini and Dirk Groeneveld and Kyle Lo and Shane Arora and Akshita Bhagia and Yuling Gu and Shengyi Huang and Matt Jordan and Nathan Lambert and Dustin Schwenk and Oyvind Tafjord and Taira Anderson and David Atkinson and Faeze Brahman and Christopher Clark and Pradeep Dasigi and Nouha Dziri and Allyson Ettinger and Michal Guerquin and David Heineman and Hamish Ivison and Pang Wei Koh and Jiacheng Liu and Saumya Malik and William Merrill and Lester James Validad Miranda and Jacob Morrison and Tyler Murray and Crystal Nam and Jake Poznanski and Valentina Pyatkin and Aman Rangapur and Michael Schmitz and Sam Skjonsberg and David Wadden and Christopher Wilhelm and Michael Wilson and Luke Zettlemoyer and Ali Farhadi and Noah A. Smith and Hannaneh Hajishirzi},
booktitle={Second Conference on Language Modeling},
year={2025},
url={https://openreview.net/forum?id=2ezugTT9kU}
}

@article{gemma3,
  title   = {Gemma 3 Technical Report},
  author  = {{Gemma Team}},
  journal = {arXiv preprint arXiv:2503.19786},
  year    = {2025}
}

@inproceedings{wolf2020transformers,
    title = "Transformers: State-of-the-Art Natural Language Processing",
    author = "Wolf, Thomas  and
      Debut, Lysandre  and
      Sanh, Victor  and
      Chaumond, Julien  and
      Delangue, Clement  and
      Moi, Anthony  and
      Cistac, Pierric  and
      Rault, Tim  and
      Louf, Remi  and
      Funtowicz, Morgan  and
      Davison, Joe  and
      Shleifer, Sam  and
      von Platen, Patrick  and
      Ma, Clara  and
      Jernite, Yacine  and
      Plu, Julien  and
      Xu, Canwen  and
      Le Scao, Teven  and
      Gugger, Sylvain  and
      Drame, Mariama  and
      Lhoest, Quentin  and
      Rush, Alexander",
    editor = "Liu, Qun  and
      Schlangen, David",
    booktitle = "Proceedings of the 2020 Conference on Empirical Methods in Natural Language Processing: System Demonstrations",
    month = oct,
    year = "2020",
    address = "Online",
    publisher = "Association for Computational Linguistics",
    url = "https://aclanthology.org/2020.emnlp-demos.6/",
    doi = "10.18653/v1/2020.emnlp-demos.6",
    pages = "38--45"
}

@article{rusu_progressive_2016,
  title={Progressive neural networks},
  author={Rusu, Andrei A and Rabinowitz, Neil C and Desjardins, Guillaume and Soyer, Hubert and Kirkpatrick, James and Kavukcuoglu, Koray and Pascanu, Razvan and Hadsell, Raia},
  journal={arXiv preprint arXiv:1606.04671},
  year={2016}
}

@inproceedings{nguyen_variational_2017,
title={Variational Continual Learning},
author={Cuong V. Nguyen and Yingzhen Li and Thang D. Bui and Richard E. Turner},
booktitle={International Conference on Learning Representations},
year={2018},
url={https://openreview.net/forum?id=BkQqq0gRb},
}

@inproceedings{lopez-paz_gradient_2017,
  title = {Gradient {Episodic} {Memory} for {Continual} {Learning}},
  volume = {30},
  url = {https://proceedings.neurips.cc/paper/2017/hash/f87522788a2be2d171666752f97ddebb-Abstract.html},
  urldate = {2022-10-04},
  booktitle = {Advances in {Neural} {Information} {Processing} {Systems}},
  author = {Lopez-Paz, David and Ranzato, Marc' Aurelio},
  year = {2017},
}

@InProceedings{rebuffi_icarl_2017,
author = {Rebuffi, Sylvestre-Alvise and Kolesnikov, Alexander and Sperl, Georg and Lampert, Christoph H.},
title = {iCaRL: Incremental Classifier and Representation Learning},
booktitle = {Proceedings of the IEEE Conference on Computer Vision and Pattern Recognition (CVPR)},
month = {July},
year = {2017}
}

@article{shin_continual_2017, 
  title={Continual learning with deep generative replay},
  author={Shin, Hanul and Lee, Jung Kwon and Kim, Jaehong and Kim, Jiwon},
  journal={Advances in neural information processing systems},
  volume={30},
  year={2017}
}

@inproceedings{zenke_continual_2017, 
  title={Continual learning through synaptic intelligence},
  author={Zenke, Friedemann and Poole, Ben and Ganguli, Surya},
  booktitle={International conference on machine learning},
  pages={3987--3995},
  year={2017},
  organization={PMLR}
}

@article{li2017learning, 
  title={Learning without forgetting},
  author={Li, Zhizhong and Hoiem, Derek},
  journal={IEEE Transactions on Pattern Analysis and Machine Intelligence},
  volume={40},
  number={12},
  pages={2935--2947},
  year={2018},
  publisher={IEEE}
}

@article{kirkpatrick2017ewc,
  title   = {Overcoming Catastrophic Forgetting in Neural Networks},
  author  = {Kirkpatrick, James and Pascanu, Razvan and Rabinowitz, Neil and Veness, Joel
             and Desjardins, Guillaume and Rusu, Andrei A. and Milan, Kieran and Quan, John
             and Ramalho, Tiago and Grabska-Barwinska, Agnieszka and others},
  journal = {Proceedings of the National Academy of Sciences (PNAS)},
  volume  = {114},
  number  = {13},
  pages   = {3521--3526},
  year    = {2017}
}

@article{roziere2023codellama,
  title   = {Code {L}lama: Open Foundation Models for Code},
  author  = {Rozière, Baptiste and Gehring, Jonas and Gloeckle, Fabian and Sootla, Sten
             and Gat, Itai and Tan, Xiaoqing Ellen and Adi, Yossi and Liu, Jingyu and
             Remez, Tal and Rapin, Jérémy and others},
  journal = {arXiv preprint arXiv:2308.12950},
  year    = {2023}
}

@article{chen2023meditron,
  title   = {{MEDITRON}-70{B}: Scaling Medical Pretraining for Large Language Models},
  author  = {Chen, Zeming and Cano, Alejandro Hernández and Romanou, Angelika and Bonnet,
             Antoine and Matoba, Kyle and Salvi, Francesco and Pagliardini, Matteo and
             Fan, Simin and Köpf, Andreas and Mohtashami, Amirkeivan and others},
  journal = {arXiv preprint arXiv:2311.16079},
  year    = {2023}
}

@article{hendryckstest2021,
  title={Measuring Massive Multitask Language Understanding},
  author={Dan Hendrycks and Collin Burns and Steven Basart and Andy Zou and Mantas Mazeika and Dawn Song and Jacob Steinhardt},
  journal={Proceedings of the International Conference on Learning Representations (ICLR)},
  year={2021}
}

@inproceedings{geva2021transformer,
    title = "Transformer Feed-Forward Layers Are Key-Value Memories",
    author = "Geva, Mor  and
      Schuster, Roei  and
      Berant, Jonathan  and
      Levy, Omer",
    editor = "Moens, Marie-Francine  and
      Huang, Xuanjing  and
      Specia, Lucia  and
      Yih, Scott Wen-tau",
    booktitle = "Proceedings of the 2021 Conference on Empirical Methods in Natural Language Processing",
    month = nov,
    year = "2021",
    address = "Online and Punta Cana, Dominican Republic",
    publisher = "Association for Computational Linguistics",
    url = "https://aclanthology.org/2021.emnlp-main.446/",
    doi = "10.18653/v1/2021.emnlp-main.446",
    pages = "5484--5495",
}

@article{elhage2022toy,
   title={Toy Models of Superposition},
   author={Elhage, Nelson and Hume, Tristan and Olsson, Catherine and Schiefer, Nicholas and Henighan, Tom and Kravec, Shauna and Hatfield-Dodds, Zac and Lasenby, Robert and Drain, Dawn and Chen, Carol and Grosse, Roger and McCandlish, Sam and Kaplan, Jared and Amodei, Dario and Wattenberg, Martin and Olah, Christopher},
   year={2022},
   journal={Transformer Circuits Thread},
    note = {\url{https://transformer-circuits.pub/2022/toy_model/index.html}},
}

@article{cobbe2021gsm8k,
  title={Training Verifiers to Solve Math Word Problems},
  author={Cobbe, Karl and Kosaraju, Vineet and Bavarian, Mohammad and Chen, Mark and Jun, Heewoo and Kaiser, Lukasz and Plappert, Matthias and Tworek, Jerry and Hilton, Jacob and Nakano, Reiichiro and Hesse, Christopher and Schulman, John},
  journal={arXiv preprint arXiv:2110.14168},
  year={2021}
}

@article{yan2023autocast++,
  title={Autocast++: Enhancing world event prediction with zero-shot ranking-based context retrieval},
  author={Yan, Qi and Seraj, Raihan and He, Jiawei and Meng, Lili and Sylvain, Tristan},
  journal={Proceedings of the International Conference on Learning Representations (ICLR)},
  year={2024}
}

@article{zou2022forecasting,
  title={Forecasting future world events with neural networks},
  author={Zou, Andy and Xiao, Tristan and Jia, Ryan and Kwon, Joe and Mazeika, Mantas and Li, Richard and Song, Dawn and Steinhardt, Jacob and Evans, Owain and Hendrycks, Dan},
  journal={Advances in Neural Information Processing Systems},
  volume={35},
  pages={27293--27305},
  year={2022}
}

@article{halawi2024approaching,
  title={Approaching human-level forecasting with language models},
  author={Halawi, Danny and Zhang, Fred and Yueh-Han, Chen and Steinhardt, Jacob},
  journal={Advances in Neural Information Processing Systems},
  volume={37},
  pages={50426--50468},
  year={2024}
}

@article{ye2024mirai,
  title={Mirai: Evaluating llm agents for event forecasting},
  author={Ye, Chenchen and Hu, Ziniu and Deng, Yihe and Huang, Zijie and Ma, Mingyu Derek and Zhu, Yanqiao and Wang, Wei},
  journal={arXiv preprint arXiv:2407.01231},
  year={2024}
}

@article{lazaridou2021mind,
  title={Mind the gap: Assessing temporal generalization in neural language models},
  author={Lazaridou, Angeliki and Kuncoro, Adhi and Gribovskaya, Elena and Agrawal, Devang and Liska, Adam and Terzi, Tayfun and Gimenez, Mai and de Masson d'Autume, Cyprien and Kocisky, Tomas and Ruder, Sebastian and others},
  journal={Advances in Neural Information Processing Systems},
  volume={34},
  pages={29348--29363},
  year={2021}
}

@article{dhingra2022timeaware,
   title={Time-Aware Language Models as Temporal Knowledge Bases},
   volume={10},
   ISSN={2307-387X},
   url={http://dx.doi.org/10.1162/tacl_a_00459},
   DOI={10.1162/tacl_a_00459},
   journal={Transactions of the Association for Computational Linguistics},
   publisher={MIT Press - Journals},
   author={Dhingra, Bhuwan and Cole, Jeremy R. and Eisenschlos, Julian Martin and Gillick, Daniel and Eisenstein, Jacob and Cohen, William W.},
   year={2022},
   pages={257–273} }

@inproceedings{li2025ticlm,
    title = "{T}i{C}-{LM}: A Web-Scale Benchmark for Time-Continual {LLM} Pretraining",
    author = "Li, Jeffrey  and
      Armandpour, Mohammadreza  and
      Mirzadeh, Seyed Iman  and
      Mehta, Sachin  and
      Shankar, Vaishaal  and
      Vemulapalli, Raviteja  and
      Bengio, Samy  and
      Tuzel, Oncel  and
      Farajtabar, Mehrdad  and
      Pouransari, Hadi  and
      Faghri, Fartash",
    editor = "Che, Wanxiang  and
      Nabende, Joyce  and
      Shutova, Ekaterina  and
      Pilehvar, Mohammad Taher",
    booktitle = "Proceedings of the 63rd Annual Meeting of the Association for Computational Linguistics (Volume 1: Long Papers)",
    month = jul,
    year = "2025",
    address = "Vienna, Austria",
    publisher = "Association for Computational Linguistics",
    url = "https://aclanthology.org/2025.acl-long.1551/",
    doi = "10.18653/v1/2025.acl-long.1551",
    pages = "32231--32273",
    ISBN = "979-8-89176-251-0"
}

@inproceedings{zhao-etal-2024-set,
    title = "Set the Clock: Temporal Alignment of Pretrained Language Models",
    author = "Zhao, Bowen  and
      Brumbaugh, Zander  and
      Wang, Yizhong  and
      Hajishirzi, Hannaneh  and
      Smith, Noah",
    editor = "Ku, Lun-Wei  and
      Martins, Andre  and
      Srikumar, Vivek",
    booktitle = "Findings of the Association for Computational Linguistics: ACL 2024",
    month = aug,
    year = "2024",
    address = "Bangkok, Thailand",
    publisher = "Association for Computational Linguistics",
    url = "https://aclanthology.org/2024.findings-acl.892/",
    doi = "10.18653/v1/2024.findings-acl.892",
    pages = "15015--15040"
}

@article{meng2022locating,
  title={Locating and editing factual associations in gpt},
  author={Meng, Kevin and Bau, David and Andonian, Alex and Belinkov, Yonatan},
  journal={Advances in neural information processing systems},
  volume={35},
  pages={17359--17372},
  year={2022}
}

@inproceedings{
lambert2024tulu,
title={Tulu 3: Pushing Frontiers in Open Language Model Post-Training},
author={Nathan Lambert and Jacob Morrison and Valentina Pyatkin and Shengyi Huang and Hamish Ivison and Faeze Brahman and Lester James Validad Miranda and Alisa Liu and Nouha Dziri and Xinxi Lyu and Yuling Gu and Saumya Malik and Victoria Graf and Jena D. Hwang and Jiangjiang Yang and Ronan Le Bras and Oyvind Tafjord and Christopher Wilhelm and Luca Soldaini and Noah A. Smith and Yizhong Wang and Pradeep Dasigi and Hannaneh Hajishirzi},
booktitle={Second Conference on Language Modeling},
year={2025},
url={https://openreview.net/forum?id=i1uGbfHHpH}
}

@article{ouyang2022training,
  title={Training language models to follow instructions with human feedback},
  author={Ouyang, Long and Wu, Jeffrey and Jiang, Xu and Almeida, Diogo and Wainwright, Carroll and Mishkin, Pamela and Zhang, Chong and Agarwal, Sandhini and Slama, Katarina and Ray, Alex and others},
  journal={Advances in neural information processing systems},
  volume={35},
  pages={27730--27744},
  year={2022}
}

@inproceedings{lin2024mitigating,
  title={Mitigating the alignment tax of rlhf},
  author={Lin, Yong and Lin, Hangyu and Xiong, Wei and Diao, Shizhe and Liu, Jianmeng and Zhang, Jipeng and Pan, Rui and Wang, Haoxiang and Hu, Wenbin and Zhang, Hanning and others},
  booktitle={Proceedings of the 2024 Conference on Empirical Methods in Natural Language Processing},
  pages={580--606},
  year={2024}
}

%%%%%%%%%%%%%%%%%%%%%%%%%%%%%%%%%%%%%%%%%%%%%%%%%%%%%%%%%%%%
\newpage
\appendix

\textbf{Limitations.}
Several scope limitations bound our conclusions. (i) Our pretraining corpus is English-only (FineWeb-Edu); whether the same recipe transfers to multilingual CPT is open. (ii) Our primary temporal evaluation is a single benchmark (Daily Oracle); temporal benchmarks vary in domain coverage and difficulty, and a multi-benchmark replication would strengthen the conclusions. (iii) Our learning-rate sweep is conducted in depth on \gemmathreeoneb only; the divergence of QA and downstream optima may take a different shape on other families. (iv) We study a single CPT cycle; whether iterative time-incremental updates compound gains or saturate over multiple rounds remains unresolved. (v) Our experiments are limited to models up to 7--8B parameters; whether the same trends hold at frontier scale is an open question.

\begin{table}[htbp]
    \centering
    \caption{Overview of appendix content.}
    \label{tab:appendix-overview}
    \setlength{\tabcolsep}{5pt}
    \renewcommand{\arraystretch}{1.4}
    \begin{tabular}{p{0.10\textwidth} p{0.58\textwidth} p{0.22\textwidth}}
        \toprule
        \textbf{Item} & \textbf{Description} & \textbf{Complements} \\
        \midrule
        Table~\ref{tab:results-long}       & Full Daily Oracle results (Macro-F1) for all models and learning-rate configurations & All experiments \\
        Table~\ref{tab:results-long-acc}   & Full Daily Oracle results (Accuracy) for all models and learning-rate configurations & All experiments \\
        Table~\ref{tab:downstream-results} & Full thirteen-task downstream evaluation (Accuracy) for all models and CPT variants & Sec.~\ref{sec:findings-general-abilities} \\
        \midrule
        Figure~\ref{fig:coupling}          & Sample-level acquisition and forgetting coupling & Sec.~\ref{sec:findings-sample-level} \\
        Figure~\ref{fig:over-time}          & CPT vs.\ base Macro-F1 over time (12-month EMA), TF and MC & Sec.~\ref{sec:findings-know-acq} \\
        Figure~\ref{fig:new-questions}      & CPT gains on questions grounded in the pretraining corpus & Sec.~\ref{sec:findings-know-acq} \\
        Figure~\ref{fig:acc-vs-checkpoints} & Knowledge acquisition and downstream accuracy across intermediate checkpoints & Sec.~\ref{sec:findings-recipe} \\
        \midrule
        Figure~\ref{fig:main-avg-supp}          & After-cutoff Accuracy counterpart to Figure~\ref{fig:main-avg} & Sec.~\ref{sec:findings-know-acq} \\
        Figure~\ref{fig:main-avg2-supp}      & Before-cutoff Accuracy counterpart to Figure~\ref{fig:main-avg} & Sec.~\ref{sec:findings-know-acq} \\
        Figure~\ref{fig:data-size-supp}    & MC and Accuracy counterparts to Figure~\ref{fig:data-size} (data quality vs.\ quantity) & Sec.~\ref{sec:findings-data-amount} \\
        Figure~\ref{fig:lr-supp}                 & Effect of learning rate on stability--plasticity trade-off & Sec.~\ref{sec:findings-lr} \\
        Figure~\ref{fig:metric-lora-supp}     & Separate TF and MC breakdowns of LoRA results, accompanying Figure~\ref{fig:metric-lora}  & Sec.~\ref{sec:findings-lora} \\
        Figure~\ref{fig:post-train-olmo-supp}   & Separate TF and MC breakdowns of post-training results, accompanying Figure~\ref{fig:post-train} & Sec.~\ref{sec:findings-deployment} \\
        Figure~\ref{fig:post-train-llama-supp}   & Daily Oracle performance of \llamathreeoneeightb model and its continued pretrained counterpart after SFT and DPO, accompanying Figure~\ref{fig:post-train} & Sec.~\ref{sec:findings-deployment} \\
        \midrule
        Sec.~\ref{app:split-stats}      & Per-class sample counts of the before-/after-cutoff TF and MC splits & Sec.~\ref{sec:findings} \\
        Sec.~\ref{app:saturation}       & Tokens-per-parameter $D/N$, pretraining compute $6ND$ and per-model CPT gains & Sec.~\ref{sec:findings-know-acq} \\
        Sec.~\ref{app:temporal-holdout} & Temporal holdout: questions drawn from pages postdating the training corpus & Sec.~\ref{sec:findings-know-acq} \\
        Sec.~\ref{app:lora-llama}       & LoRA replication on \llamathreetwothreeb at ranks 128 and 256 & Sec.~\ref{sec:findings-lora} \\
        Sec.~\ref{app:qwen}             & \textsc{Qwen3} experiments and sequential (multi-round) updating & Sec.~\ref{sec:findings-know-acq} \\
        \bottomrule
    \end{tabular}
\end{table}

\begin{table}[t]
\centering
\caption{Combined MC and TF Model Evaluation Results (Macro-F1). Instruct models are evaluated with generation.}
\label{tab:results-long}
\begin{adjustbox}{max width=\linewidth, max totalheight=\dimexpr\textheight-4\baselineskip\relax, keepaspectratio}
\begin{tabular}{llcccccc}
\toprule
 & & & \multicolumn{2}{c}{\textbf{MC}} & \multicolumn{2}{c}{\textbf{TF}} & \\
\cmidrule(lr){4-5}\cmidrule(lr){6-7}
\textbf{Model} & \textbf{Pretrain LR} & \textbf{Score} & \textbf{Before} & \textbf{After}  & \textbf{Before} & \textbf{After}  & \textbf{All Avg} \\
\midrule
Gemma3-1B & --- & --- & 0.38 & 0.43  & 0.53 & 0.49  & 0.46 \\
Llama-3.1-8B & --- & --- & 0.55 & 0.49  & 0.74 & 0.63  & 0.60 \\
Llama-3.2-1B & --- & --- & 0.36 & 0.37  & 0.60 & 0.56  & 0.47 \\
Llama-3.2-3B & --- & --- & 0.47 & 0.46  & 0.60 & 0.53  & 0.51 \\
OLMo2-1B & --- & --- & 0.36 & 0.42  & 0.53 & 0.49  & 0.45 \\
OLMo2-7B & --- & --- & 0.53 & 0.49  & 0.61 & 0.50  & 0.53 \\
Llama-3.1-8B-Instruct-3 & --- & --- & 0.43 & 0.37  & 0.65 & 0.55  & 0.50 \\
Llama-3.1-8B-Tulu-3 & --- & --- & 0.52 & 0.38  & 0.63 & 0.55  & 0.52 \\
Llama-3.1-8B-Tulu-3-SFT-DPO & --- & --- & 0.51 & 0.40  & 0.60 & 0.51  & 0.51 \\
Llama-3.1-8B-Tulu-3-SFT & --- & --- & 0.49 & 0.37  & 0.59 & 0.52  & 0.49 \\
OLMo-2-1124-7B-Tulu-3 & --- & --- & 0.44 & 0.44  & 0.39 & 0.37  & 0.41 \\
OLMo-2-1124-7B-Tulu-3-SFT-DPO & --- & --- & 0.43 & 0.43  & 0.38 & 0.37  & 0.40 \\
OLMo-2-1124-7B-Tulu-3-SFT & --- & --- & 0.45 & 0.43  & 0.39 & 0.37  & 0.41 \\
\midrule
gemma3-1b-cpt & $10^{-4}$ & 4.0 & 0.35 & 0.51  & 0.58 & 0.54  & 0.49 \\
gemma3-1b-cpt & $10^{-5}$ & 4.0 & 0.37 & 0.41  & 0.58 & 0.56  & 0.48 \\
gemma3-1b-cpt & $10^{-6}$ & 4.0 & 0.38 & 0.43  & 0.53 & 0.50  & 0.46 \\
gemma3-1b-cpt & $2 \times 10^{-5}$ & 4.0 & 0.37 & 0.43  & 0.61 & 0.59  & 0.50 \\
gemma3-1b-cpt & $3 \times 10^{-4}$ & 4.0 & 0.33 & 0.43  & 0.53 & 0.49  & 0.45 \\
gemma3-1b-cpt & $3 \times 10^{-5}$ & 4.0 & 0.37 & 0.44  & 0.63 & 0.60  & 0.51 \\
gemma3-1b-cpt (7 epochs) & $3 \times 10^{-5}$ & 4.0 & 0.36 & 0.47  & 0.62 & 0.60  & 0.52 \\
gemma3-1b-cpt & $3 \times 10^{-5}$ & 3.5 & 0.37 & 0.47  & 0.62 & 0.58  & 0.51 \\
gemma3-1b-cpt & $3 \times 10^{-6}$ & 4.0 & 0.38 & 0.44  & 0.56 & 0.54  & 0.48 \\
gemma3-1b-cpt & $4 \times 10^{-5}$ & 4.0 & 0.37 & 0.46  & 0.61 & 0.58  & 0.50 \\
gemma3-1b-cpt & $5 \times 10^{-5}$ & 4.0 & 0.36 & 0.45  & 0.60 & 0.59  & 0.50 \\
gemma3-1b-cpt & $5 \times 10^{-6}$ & 4.0 & 0.37 & 0.42  & 0.57 & 0.54  & 0.48 \\
gemma3-1b-cpt & $8 \times 10^{-5}$ & 4.0 & 0.36 & 0.48  & 0.59 & 0.57  & 0.50 \\
gemma3-1b-cpt (LoRA, r=128, a=128) & $10^{-4}$ & 4.0 & 0.36 & 0.46  & 0.61 & 0.59  & 0.51 \\
gemma3-1b-cpt (LoRA, r=128, a=128)  & $3 \times 10^{-4}$ & 4.0 & 0.36 & 0.49  & 0.62 & 0.60  & 0.52 \\
gemma3-1b-cpt (LoRA, r=128, a=256)  & $10^{-4}$ & 4.0 & 0.37 & 0.48  & 0.55 & 0.55  & 0.49 \\
gemma3-1b-cpt (LoRA, r=128, a=256)  & $3 \times 10^{-4}$ & 4.0 & 0.36 & 0.50  & 0.59 & 0.56  & 0.50 \\
gemma3-1b-cpt (LoRA, r=128, a=128)  & $6 \times 10^{-4}$ & 4.0 & 0.35 & 0.45  & 0.59 & 0.52  & 0.48 \\
gemma3-1b-cpt (LoRA, r=256, a=256)  & $10^{-4}$ & 4.0 & 0.36 & 0.50  & 0.53 & 0.53  & 0.48 \\
gemma3-1b-cpt (LoRA, r=256, a=256)  & $3 \times 10^{-4}$ & 4.0 & 0.35 & 0.49  & 0.61 & 0.58  & 0.51 \\
gemma3-1b-cpt (LoRA, r=256, a=256)  & $6 \times 10^{-4}$ & 4.0 & 0.34 & 0.50  & 0.57 & 0.56  & 0.49 \\
gemma3-1b-cpt (LoRA, r=32, a=32)  & $10^{-4}$ & 4.0 & 0.36 & 0.47  & 0.57 & 0.55  & 0.49 \\
gemma3-1b-cpt (LoRA, r=32, a=32)  & $3 \times 10^{-4}$ & 4.0 & 0.36 & 0.48  & 0.56 & 0.56  & 0.49 \\
gemma3-1b-cpt (LoRA, r=32, a=64)   & $10^{-4}$ & 4.0 & 0.36 & 0.46  & 0.55 & 0.55  & 0.48 \\
gemma3-1b-cpt (LoRA, r=32, a=64)  & $3 \times 10^{-4}$ & 4.0 & 0.36 & 0.48  & 0.50 & 0.49  & 0.46 \\
\midrule
llama31-8b-cpt & $10^{-5}$ & 4.0 & 0.56 & 0.51  & 0.72 & 0.65  & 0.61 \\
llama31-8b-cpt & $10^{-6}$ & 4.0 & 0.55 & 0.48  & 0.74 & 0.64  & 0.61 \\
llama31-8b-cpt & $10^{-6}$ & 3.5 & 0.55 & 0.49  & 0.74 & 0.64  & 0.61 \\
\midrule
llama32-1b-cpt & $10^{-5}$ & 4.0 & 0.37 & 0.41  & 0.47 & 0.44  & 0.42 \\
llama32-1b-cpt & $10^{-6}$ & 4.0 & 0.36 & 0.38  & 0.57 & 0.53  & 0.46 \\
llama32-1b-cpt & $3 \times 10^{-6}$ & 4.0 & 0.36 & 0.39  & 0.54 & 0.52  & 0.45 \\
llama32-1b-cpt & $5 \times 10^{-6}$ & 4.0 & 0.37 & 0.39  & 0.49 & 0.47  & 0.43 \\
\midrule
llama32-3b-cpt & $10^{-5}$ & 4.0 & 0.47 & 0.49  & 0.61 & 0.54  & 0.53 \\
llama32-3b-cpt & $10^{-6}$ & 4.0 & 0.47 & 0.47  & 0.61 & 0.55  & 0.53 \\
llama32-3b-cpt & $3 \times 10^{-6}$ & 4.0 & 0.47 & 0.48  & 0.63 & 0.56  & 0.54 \\
llama32-3b-cpt & $5 \times 10^{-6}$ & 4.0 & 0.47 & 0.49  & 0.63 & 0.56  & 0.54 \\
\midrule
olmo2-1b-cpt & $10^{-6}$ & 3.5 & 0.40 & 0.46  & 0.63 & 0.59  & 0.52 \\
olmo2-1b-cpt & $10^{-6}$ & 4.0 & 0.39 & 0.44  & 0.60 & 0.56  & 0.50 \\
olmo2-1b-cpt & $2.63 \times 10^{-5}$ & 4.0 & 0.35 & 0.45  & 0.49 & 0.45  & 0.44 \\
\midrule
olmo2-7b-cpt & $10^{-6}$ & 4.0 & 0.55 & 0.53  & 0.69 & 0.62  & 0.60 \\
\midrule
llama31-8b-cpt-sft-dpo & $1 \times 10^{-6}$ & 4.0 & 0.51 & 0.40  & 0.58 & 0.44  & 0.48 \\
llama31-8b-cpt-sft & $1 \times 10^{-6}$ & 4.0 & 0.51 & 0.41  & 0.59 & 0.43  & 0.48 \\
llama31-8b-cpt-sft-dpo & $1 \times 10^{-5}$ & 4.0 & 0.53 & 0.47  & 0.57 & 0.47  & 0.51 \\
llama31-8b-cpt-sft & $1 \times 10^{-5}$ & 4.0 & 0.53 & 0.47  & 0.59 & 0.47  & 0.52 \\
llama31-8b-cpt-sft-dpo & $1 \times 10^{-6}$ & 3.5 & 0.51 & 0.38  & 0.62 & 0.49  & 0.50 \\
llama31-8b-cpt-sft & $1 \times 10^{-6}$ & 3.5 & 0.52 & 0.38  & 0.62 & 0.51  & 0.51 \\
olmo2-7b-cpt-sft+dpo & $1 \times 10^{-6}$ & 4.0 & 0.48 & 0.46  & 0.60 & 0.54  & 0.52 \\
olmo2-7b-cpt-sft & $1 \times 10^{-6}$ & 4.0 & 0.47 & 0.45  & 0.56 & 0.51  & 0.50 \\

\bottomrule
\end{tabular}
\end{adjustbox}
\end{table}

\begin{table}[t]
\centering
\caption{Combined MC and TF Model Evaluation Results (Accuracy). Results with $^\star$ are taken from \citet{dai2025dailyoracle}.}
\label{tab:results-long-acc}
\begin{adjustbox}{max width=\linewidth, max totalheight=\dimexpr\textheight-4\baselineskip\relax, keepaspectratio}
\begin{tabular}{llcccccc}
\toprule
 & & & \multicolumn{2}{c}{\textbf{MC}} & \multicolumn{2}{c}{\textbf{TF}} & \\
\cmidrule(lr){4-5}\cmidrule(lr){6-7}
\textbf{Model} & \textbf{Pretraining LR} & \textbf{Score} & \textbf{Before} & \textbf{After}  & \textbf{Before} & \textbf{After}  & \textbf{All Avg} \\
\midrule
Claude-3.5-Sonnet$^\star$ & --- & --- & 0.74 & 0.61  & 0.78 & 0.59  & 0.68 \\
GPT-3.5$^\star$ & --- & --- & 0.50 & 0.44  & 0.62 & 0.57  & 0.53 \\
GPT-4$^\star$ & --- & --- & 0.68 & 0.53  & 0.65 & 0.57  & 0.61 \\
Gemma2-2B-Instruct$^\star$ & --- & --- & 0.47 & 0.44  & 0.58 & 0.56  & 0.51 \\
Llama-3-8B-Instruct$^\star$ & --- & --- & 0.52 & 0.45  & 0.64 & 0.57  & 0.55 \\
Mistral-7B-Instruct$^\star$ & --- & --- & 0.50 & 0.43  & 0.53 & 0.40  & 0.47 \\
Mixtral-8x7B-Instruct$^\star$ & --- & --- & 0.55 & 0.48  & 0.51 & 0.37  & 0.48 \\
Qwen2-7B-Instruct$^\star$ & --- & --- & 0.54 & 0.51  & 0.59 & 0.53  & 0.54 \\
Gemma3-1B & --- & --- & 0.38 & 0.44  & 0.59 & 0.56  & 0.49 \\
Llama-3.2-1B & --- & --- & 0.37 & 0.38  & 0.60 & 0.57  & 0.48 \\
Llama-3.2-3B & --- & --- & 0.48 & 0.46  & 0.61 & 0.58  & 0.53 \\
OLMo2-1B & --- & --- & 0.37 & 0.43  & 0.55 & 0.54  & 0.47 \\
OLMo2-7B & --- & --- & 0.53 & 0.50  & 0.62 & 0.54  & 0.55 \\
Llama-3.1-8B & --- & --- & 0.55 & 0.49  & 0.74 & 0.64  & 0.60 \\
Llama-3.1-8B-Instruct & --- & --- & 0.53 & 0.46  & 0.66 & 0.58  & 0.56 \\
Llama-3.1-8B-Tulu-3 & --- & --- & 0.52 & 0.40  & 0.63 & 0.55  & 0.53 \\
Llama-3.1-8B-Tulu-3-SFT-DPO & --- & --- & 0.52 & 0.41  & 0.61 & 0.56  & 0.53 \\
Llama-3.1-8B-Tulu-3-SFT & --- & --- & 0.50 & 0.38  & 0.61 & 0.53  & 0.51 \\
OLMo-2-1124-7B-Tulu-3 & --- & --- & 0.46 & 0.45  & 0.50 & 0.52  & 0.48 \\
OLMo-2-1124-7B-SFT-DPO & --- & --- & 0.46 & 0.45  & 0.50 & 0.52  & 0.48 \\
OLMo-2-1124-7B-SFT & --- & --- & 0.46 & 0.44  & 0.50 & 0.52  & 0.48 \\
\midrule
gemma3-1b-cpt & $10^{-4}$ & 4.0 & 0.35 & 0.51  & 0.61 & 0.56  & 0.51 \\
gemma3-1b-cpt & $10^{-5}$ & 4.0 & 0.37 & 0.42  & 0.61 & 0.58  & 0.50 \\
gemma3-1b-cpt & $10^{-6}$ & 4.0 & 0.38 & 0.44  & 0.60 & 0.56  & 0.49 \\
gemma3-1b-cpt & $2 \times 10^{-5}$ & 4.0 & 0.37 & 0.43  & 0.62 & 0.59  & 0.50 \\
gemma3-1b-cpt & $3 \times 10^{-4}$ & 4.0 & 0.33 & 0.44  & 0.55 & 0.55  & 0.47 \\
gemma3-1b-cpt & $3 \times 10^{-5}$ & 4.0 & 0.37 & 0.45  & 0.63 & 0.60  & 0.51 \\
gemma3-1b-cpt (7 epochs) & $3 \times 10^{-5}$ & 4.0 & 0.37 & 0.47  & 0.63 & 0.60  & 0.52 \\
gemma3-1b-cpt & $3 \times 10^{-5}$ & 3.5 & 0.37 & 0.48  & 0.63 & 0.59  & 0.52 \\
gemma3-1b-cpt & $3 \times 10^{-6}$ & 4.0 & 0.38 & 0.44  & 0.61 & 0.57  & 0.50 \\
gemma3-1b-cpt & $4 \times 10^{-5}$ & 4.0 & 0.37 & 0.46  & 0.61 & 0.59  & 0.51 \\
gemma3-1b-cpt & $5 \times 10^{-5}$ & 4.0 & 0.36 & 0.45  & 0.61 & 0.59  & 0.50 \\
gemma3-1b-cpt & $5 \times 10^{-6}$ & 4.0 & 0.38 & 0.43  & 0.61 & 0.58  & 0.50 \\
gemma3-1b-cpt & $8 \times 10^{-5}$ & 4.0 & 0.36 & 0.48  & 0.62 & 0.58  & 0.51 \\
gemma3-1b-cpt (LoRA, r=128, a=128) & $10^{-4}$ & 4.0 & 0.36 & 0.47  & 0.62 & 0.60  & 0.51 \\
gemma3-1b-cpt (LoRA, r=128, a=128) & $3 \times 10^{-4}$ & 4.0 & 0.36 & 0.49  & 0.62 & 0.61  & 0.52 \\
gemma3-1b-cpt (LoRA, r=128, a=256) & $10^{-4}$ & 4.0 & 0.37 & 0.49  & 0.57 & 0.59  & 0.50 \\
gemma3-1b-cpt (LoRA, r=128, a=256) & $3 \times 10^{-4}$ & 4.0 & 0.36 & 0.50  & 0.59 & 0.56  & 0.50 \\
gemma3-1b-cpt (LoRA, r=128, a=128) & $6 \times 10^{-4}$ & 4.0 & 0.35 & 0.46  & 0.63 & 0.56  & 0.50 \\
gemma3-1b-cpt (LoRA, r=256, a=256) & $10^{-4}$ & 4.0 & 0.36 & 0.50  & 0.56 & 0.57  & 0.50 \\
gemma3-1b-cpt (LoRA, r=256, a=256) & $3 \times 10^{-4}$ & 4.0 & 0.36 & 0.49  & 0.63 & 0.59  & 0.52 \\
gemma3-1b-cpt (LoRA, r=256, a=256) & $6 \times 10^{-4}$ & 4.0 & 0.35 & 0.50  & 0.58 & 0.57  & 0.50 \\
gemma3-1b-cpt (LoRA, r=32, a=32) & $10^{-4}$ & 4.0 & 0.36 & 0.47  & 0.58 & 0.57  & 0.50 \\
gemma3-1b-cpt (LoRA, r=32, a=32) & $3 \times 10^{-4}$ & 4.0 & 0.36 & 0.49  & 0.57 & 0.58  & 0.50 \\
gemma3-1b-cpt (LoRA, r=32, a=64) & $10^{-4}$ & 4.0 & 0.37 & 0.47  & 0.56 & 0.57  & 0.49 \\
gemma3-1b-cpt (LoRA, r=32, a=64) & $3 \times 10^{-4}$ & 4.0 & 0.37 & 0.48  & 0.56 & 0.56  & 0.49 \\
\midrule
llama31-8b-cpt & $10^{-5}$ & 4.0 & 0.56 & 0.51  & 0.73 & 0.66  & 0.62 \\
llama31-8b-cpt & $10^{-6}$ & 4.0 & 0.56 & 0.49  & 0.74 & 0.65  & 0.61 \\
llama31-8b-cpt & $10^{-6}$ & 3.5 & 0.56 & 0.49  & 0.74 & 0.65  & 0.61 \\
\midrule
llama32-1b-cpt & $10^{-5}$ & 4.0 & 0.37 & 0.41  & 0.57 & 0.55  & 0.47 \\
llama32-1b-cpt & $10^{-6}$ & 4.0 & 0.37 & 0.39  & 0.59 & 0.56  & 0.48 \\
llama32-1b-cpt & $3 \times 10^{-6}$ & 4.0 & 0.37 & 0.40  & 0.58 & 0.56  & 0.48 \\
llama32-1b-cpt & $5 \times 10^{-6}$ & 4.0 & 0.37 & 0.40  & 0.56 & 0.55  & 0.47 \\
\midrule
llama32-3b-cpt & $10^{-5}$ & 4.0 & 0.48 & 0.50  & 0.62 & 0.58  & 0.54 \\
llama32-3b-cpt & $10^{-6}$ & 4.0 & 0.48 & 0.47  & 0.62 & 0.58  & 0.54 \\
llama32-3b-cpt & $3 \times 10^{-6}$ & 4.0 & 0.48 & 0.48  & 0.63 & 0.59  & 0.55 \\
llama32-3b-cpt & $5 \times 10^{-6}$ & 4.0 & 0.48 & 0.49  & 0.63 & 0.59  & 0.55 \\
\midrule
olmo2-1b-cpt & $10^{-6}$ & 3.5 & 0.41 & 0.47  & 0.65 & 0.60  & 0.53 \\
olmo2-1b-cpt & $10^{-6}$ & 4.0 & 0.41 & 0.45  & 0.62 & 0.58  & 0.51 \\
olmo2-1b-cpt & $2.63 \times 10^{-5}$ & 4.0 & 0.38 & 0.46  & 0.59 & 0.55  & 0.49 \\
\midrule
olmo2-7b-cpt & $10^{-6}$ & 4.0 & 0.55 & 0.53  & 0.69 & 0.63  & 0.60 \\
\midrule
llama31-8b-cpt-sft-dpo & $10^{-6}$ & 4.0 & 0.52 & 0.42  & 0.60 & 0.53  & 0.51 \\
llama31-8b-cpt-sft & $10^{-6}$ & 4.0 & 0.51 & 0.42  & 0.60 & 0.53  & 0.52 \\
llama31-8b-cpt-sft-dpo & $10^{-5}$ & 4.0 & 0.53 & 0.48  & 0.59 & 0.54  & 0.54 \\
llama31-8b-cpt-sft & $10^{-5}$ & 4.0 & 0.53 & 0.47  & 0.61 & 0.55  & 0.54 \\
llama31-8b-cpt-sft-dpo & $10^{-6}$ & 3.5 & 0.52 & 0.39  & 0.62 & 0.53  & 0.52 \\
llama31-8b-cpt-sft & $10^{-6}$ & 3.5 & 0.52 & 0.39  & 0.62 & 0.54  & 0.52 \\
olmo2-7b-cpt-sft+dpo & $10^{-6}$ & 4.0 & 0.49 & 0.46  & 0.61 & 0.59  & 0.54 \\
olmo2-7b-cpt-sft & $10^{-6}$ & 4.0 & 0.48 & 0.46  & 0.59 & 0.57  & 0.53 \\
\bottomrule
\end{tabular}
\end{adjustbox}
\end{table}

\begin{table}[t]
\centering
\setlength{\tabcolsep}{4pt}
\renewcommand{\arraystretch}{0.88}
\caption{Downstream evaluation results (accuracy, $\uparrow$). MMLU Avg is the mean over all 12 MMLU sub-tasks. Avg is the macro-average over all reported metrics. \textbf{Bold} marks the best CPT variant within each model group.}
\label{tab:downstream-results}
\resizebox{\linewidth}{!}{%
\begin{tabular}{l|l|@{\;}c|c|c|c|c|c|c|c|c|c|c|c|c|c|@{\quad}c}
\toprule
\multirow{2}{*}{\textbf{Model}} & \multirow{2}{*}{\textbf{LR}} & \multirow{2}{*}{\textbf{Score}} & \multicolumn{13}{c}{\textbf{Accuracy ($\uparrow$)}} & \multirow{2}{*}{\rotatebox{90}{\textbf{Avg}\rule{0pt}{2.2em}}} \\
\cmidrule(lr){4-16}
 &  &  & \rotatebox{90}{\textbf{PIQA}\rule{0pt}{2.2em}} & \rotatebox{90}{\textbf{HellaSwag}\rule{0pt}{2.2em}} & \rotatebox{90}{\textbf{WinoGrande}\rule{0pt}{2.2em}} & \rotatebox{90}{\textbf{OpenBookQA}\rule{0pt}{2.2em}} & \rotatebox{90}{\textbf{BoolQ}\rule{0pt}{2.2em}} & \rotatebox{90}{\textbf{SciQ}\rule{0pt}{2.2em}} & \rotatebox{90}{\textbf{ARC-Easy}\rule{0pt}{2.2em}} & \rotatebox{90}{\textbf{ARC-Challenge}\rule{0pt}{2.2em}} & \rotatebox{90}{\textbf{COPA}\rule{0pt}{2.2em}} & \rotatebox{90}{\textbf{CommonsenseQA}\rule{0pt}{2.2em}} & \rotatebox{90}{\textbf{SocialIQA}\rule{0pt}{2.2em}} & \rotatebox{90}{\textbf{Basic Arith.}\rule{0pt}{2.2em}} & \rotatebox{90}{\textbf{MMLU Avg}\rule{0pt}{2.2em}} &  \\
\midrule
Llama-3.1-8B & --- & --- & 0.81 & 0.79 & 0.74 & 0.45 & 0.83 & 0.96 & 0.82 & 0.53 & 0.88 & 0.61 & 0.53 & 0.97 & 0.60 & 0.73 \\
\;\;CPT & $10^{-6}$ & 3.5 & $\mathbf{0.81}$ & 0.79 & $\mathbf{0.74}$ & $\mathbf{0.46}$ & 0.82 & $\mathbf{0.96}$ & $\mathbf{0.83}$ & $\mathbf{0.55}$ & $\mathbf{0.87}$ & 0.61 & 0.53 & $\mathbf{0.98}$ & $\mathbf{0.60}$ & $\mathbf{0.73}$ \\
\;\;CPT (cont) & $10^{-6}$ & cont & 0.81 & $\mathbf{0.79}$ & 0.74 & 0.45 & $\mathbf{0.82}$ & $\mathbf{0.96}$ & 0.83 & $\mathbf{0.55}$ & $\mathbf{0.87}$ & $\mathbf{0.61}$ & $\mathbf{0.53}$ & 0.98 & 0.60 & 0.73 \\
\midrule
Llama-3.2-1B & --- & --- & 0.74 & 0.64 & 0.61 & 0.38 & 0.65 & 0.92 & 0.65 & 0.37 & 0.78 & 0.47 & 0.46 & 0.77 & 0.33 & 0.60 \\
\;\;CPT & $10^{-6}$ & 4.0 & 0.74 & $\mathbf{0.64}$ & 0.60 & 0.38 & 0.65 & 0.92 & 0.65 & 0.36 & 0.77 & 0.47 & $\mathbf{0.46}$ & $\mathbf{0.78}$ & 0.33 & 0.60 \\
\;\;CPT & $3\times10^{-6}$ & 4.0 & 0.75 & 0.64 & 0.60 & $\mathbf{0.39}$ & $\mathbf{0.65}$ & 0.92 & 0.66 & 0.35 & 0.77 & $\mathbf{0.47}$ & $\mathbf{0.46}$ & 0.76 & 0.33 & 0.60 \\
\;\;CPT & $5\times10^{-6}$ & 4.0 & 0.74 & 0.64 & $\mathbf{0.61}$ & 0.39 & 0.65 & $\mathbf{0.93}$ & 0.66 & 0.35 & $\mathbf{0.78}$ & 0.46 & 0.46 & 0.77 & 0.34 & 0.60 \\
\;\;CPT & $10^{-5}$ & 4.0 & $\mathbf{0.75}$ & 0.63 & 0.61 & 0.38 & 0.65 & 0.93 & $\mathbf{0.69}$ & $\mathbf{0.37}$ & $\mathbf{0.78}$ & 0.45 & 0.45 & 0.77 & $\mathbf{0.34}$ & $\mathbf{0.60}$ \\
\midrule
Llama-3.2-3B & --- & --- & 0.77 & 0.74 & 0.69 & 0.41 & 0.74 & 0.95 & 0.75 & 0.43 & 0.85 & 0.59 & 0.52 & 0.97 & 0.52 & 0.69 \\
\;\;CPT & $10^{-6}$ & 4.0 & 0.77 & 0.73 & 0.69 & 0.42 & 0.74 & $\mathbf{0.95}$ & 0.75 & 0.45 & 0.85 & $\mathbf{0.60}$ & $\mathbf{0.52}$ & 0.97 & $\mathbf{0.52}$ & 0.69 \\
\;\;CPT & $3\times10^{-6}$ & 4.0 & $\mathbf{0.78}$ & 0.73 & $\mathbf{0.70}$ & 0.42 & 0.75 & 0.95 & 0.76 & 0.46 & $\mathbf{0.86}$ & 0.59 & 0.52 & $\mathbf{0.97}$ & 0.52 & $\mathbf{0.69}$ \\
\;\;CPT & $5\times10^{-6}$ & 4.0 & 0.78 & $\mathbf{0.73}$ & 0.69 & 0.41 & 0.75 & 0.95 & $\mathbf{0.77}$ & 0.46 & 0.85 & 0.58 & 0.52 & 0.96 & 0.52 & 0.69 \\
\;\;CPT & $10^{-5}$ & 4.0 & 0.77 & 0.73 & 0.69 & $\mathbf{0.43}$ & $\mathbf{0.75}$ & 0.95 & $\mathbf{0.77}$ & $\mathbf{0.46}$ & $\mathbf{0.86}$ & 0.57 & 0.51 & 0.96 & 0.52 & 0.69 \\
\midrule
Gemma-3-1B & --- & --- & 0.70 & 0.54 & 0.59 & 0.39 & 0.38 & 0.43 & 0.46 & 0.33 & 0.76 & 0.24 & 0.40 & 0.55 & 0.26 & 0.46 \\
\;\;CPT & $10^{-6}$ & 4.0 & $\mathbf{0.70}$ & 0.54 & 0.58 & 0.40 & 0.38 & 0.44 & 0.46 & 0.34 & 0.78 & 0.24 & 0.40 & 0.55 & 0.26 & 0.47 \\
\;\;CPT & $3\times10^{-6}$ & 4.0 & 0.70 & 0.54 & 0.58 & 0.39 & 0.38 & 0.44 & 0.45 & 0.34 & 0.76 & 0.24 & 0.40 & 0.55 & 0.26 & 0.46 \\
\;\;CPT & $5\times10^{-6}$ & 4.0 & 0.70 & 0.54 & 0.58 & 0.39 & 0.38 & 0.43 & 0.46 & 0.33 & 0.77 & 0.24 & 0.40 & 0.55 & 0.26 & 0.46 \\
\;\;CPT & $10^{-5}$ & 4.0 & 0.70 & 0.54 & 0.57 & 0.39 & 0.38 & 0.43 & 0.46 & 0.34 & 0.77 & 0.24 & 0.40 & $\mathbf{0.56}$ & 0.26 & 0.46 \\
\;\;CPT & $2\times10^{-5}$ & 4.0 & 0.70 & 0.54 & 0.59 & 0.39 & 0.38 & 0.44 & 0.47 & 0.34 & 0.78 & 0.24 & 0.40 & 0.54 & 0.26 & 0.47 \\
\;\;CPT & $3\times10^{-5}$ & 4.0 & 0.70 & 0.54 & 0.58 & 0.40 & 0.38 & 0.47 & 0.47 & 0.34 & 0.79 & 0.24 & 0.40 & 0.53 & 0.26 & 0.47 \\
\;\;CPT & $4\times10^{-5}$ & 4.0 & 0.70 & 0.54 & 0.56 & 0.40 & 0.38 & 0.49 & 0.48 & 0.34 & $\mathbf{0.81}$ & 0.25 & 0.40 & 0.51 & 0.26 & 0.47 \\
\;\;CPT & $5\times10^{-5}$ & 4.0 & 0.70 & 0.54 & 0.56 & 0.41 & 0.38 & 0.55 & 0.49 & 0.35 & 0.78 & 0.25 & 0.40 & 0.50 & 0.26 & 0.47 \\
\;\;CPT & $8\times10^{-5}$ & 4.0 & 0.70 & 0.54 & $\mathbf{0.60}$ & 0.41 & 0.38 & 0.63 & 0.54 & 0.35 & 0.76 & 0.25 & 0.40 & 0.51 & 0.26 & 0.49 \\
\;\;CPT & $10^{-4}$ & 4.0 & 0.70 & 0.53 & 0.57 & $\mathbf{0.43}$ & 0.38 & 0.68 & 0.55 & 0.34 & 0.76 & 0.25 & 0.41 & 0.50 & 0.25 & 0.49 \\
\;\;CPT & $3\times10^{-4}$ & 4.0 & 0.68 & 0.49 & 0.58 & 0.42 & $\mathbf{0.62}$ & $\mathbf{0.86}$ & $\mathbf{0.64}$ & 0.33 & 0.74 & $\mathbf{0.27}$ & 0.40 & 0.47 & $\mathbf{0.26}$ & $\mathbf{0.52}$ \\
\;\;CPT (7ep) & $3\times10^{-5}$ & 4.0 & 0.70 & 0.54 & 0.57 & 0.40 & 0.38 & 0.46 & 0.46 & 0.35 & 0.78 & 0.25 & 0.41 & 0.50 & 0.26 & 0.47 \\
\;\;CPT & $3\times10^{-5}$ & 3.5 & 0.70 & 0.54 & 0.58 & 0.41 & 0.38 & 0.49 & 0.48 & 0.34 & 0.79 & 0.25 & 0.40 & 0.52 & 0.26 & 0.47 \\
\;\;LoRA $r$32$\alpha$32 & $10^{-4}$ & 4.0 & 0.69 & 0.54 & 0.57 & 0.39 & 0.38 & 0.42 & 0.48 & 0.34 & 0.79 & 0.24 & 0.40 & 0.51 & 0.26 & 0.46 \\
\;\;LoRA $r$32$\alpha$32 & $3\times10^{-4}$ & 4.0 & 0.69 & $\mathbf{0.54}$ & 0.57 & 0.40 & 0.38 & 0.49 & 0.48 & 0.33 & 0.74 & 0.25 & 0.41 & 0.47 & 0.26 & 0.46 \\
\;\;LoRA $r$32$\alpha$64 & $10^{-4}$ & 4.0 & 0.69 & 0.54 & 0.57 & 0.39 & 0.38 & 0.44 & 0.47 & 0.32 & 0.73 & 0.24 & 0.40 & 0.51 & 0.26 & 0.46 \\
\;\;LoRA $r$32$\alpha$64 & $3\times10^{-4}$ & 4.0 & 0.70 & 0.54 & 0.58 & 0.41 & 0.38 & 0.55 & 0.47 & 0.34 & 0.78 & 0.25 & 0.40 & 0.48 & 0.26 & 0.47 \\
\;\;LoRA $r$128$\alpha$128 & $10^{-4}$ & 4.0 & 0.70 & 0.54 & 0.58 & 0.40 & 0.38 & 0.45 & 0.47 & 0.33 & 0.80 & 0.24 & 0.40 & 0.50 & 0.26 & 0.47 \\
\;\;LoRA $r$128$\alpha$128 & $3\times10^{-4}$ & 4.0 & 0.70 & 0.54 & 0.58 & 0.41 & 0.38 & 0.62 & 0.49 & 0.34 & $\mathbf{0.81}$ & 0.25 & 0.40 & 0.51 & 0.26 & 0.48 \\
\;\;LoRA $r$128$\alpha$128 & $6\times10^{-4}$ & 4.0 & 0.70 & 0.54 & 0.59 & 0.41 & 0.38 & 0.65 & 0.55 & 0.35 & 0.79 & 0.25 & 0.40 & 0.42 & 0.26 & 0.48 \\
\;\;LoRA $r$128$\alpha$256 & $10^{-4}$ & 4.0 & 0.70 & 0.54 & 0.58 & 0.41 & 0.38 & 0.53 & 0.48 & 0.33 & 0.78 & 0.25 & 0.40 & 0.50 & 0.26 & 0.47 \\
\;\;LoRA $r$128$\alpha$256 & $3\times10^{-4}$ & 4.0 & 0.70 & 0.54 & 0.58 & 0.41 & 0.38 & 0.63 & 0.52 & 0.34 & 0.78 & 0.25 & $\mathbf{0.41}$ & 0.44 & 0.26 & 0.48 \\
\;\;LoRA $r$256$\alpha$256 & $10^{-4}$ & 4.0 & 0.70 & 0.54 & 0.58 & 0.40 & 0.38 & 0.53 & 0.49 & 0.34 & 0.80 & 0.25 & 0.40 & 0.50 & 0.26 & 0.47 \\
\;\;LoRA $r$256$\alpha$256 & $3\times10^{-4}$ & 4.0 & 0.69 & 0.54 & 0.57 & 0.42 & 0.38 & 0.65 & 0.52 & 0.34 & $\mathbf{0.81}$ & 0.25 & 0.40 & 0.48 & 0.26 & 0.49 \\
\;\;LoRA $r$256$\alpha$256 & $6\times10^{-4}$ & 4.0 & 0.69 & 0.53 & 0.58 & 0.43 & 0.38 & 0.70 & 0.54 & $\mathbf{0.38}$ & 0.80 & 0.25 & 0.40 & 0.51 & 0.26 & 0.50 \\
\midrule
OLMo2-1B & --- & --- & 0.76 & 0.69 & 0.64 & 0.39 & 0.62 & 0.96 & 0.73 & 0.41 & 0.81 & 0.57 & 0.50 & 0.85 & 0.43 & 0.64 \\
\;\;CPT & $10^{-6}$ & 4.0 & $\mathbf{0.76}$ & $\mathbf{0.68}$ & $\mathbf{0.64}$ & 0.39 & $\mathbf{0.72}$ & $\mathbf{0.94}$ & 0.75 & $\mathbf{0.45}$ & 0.80 & $\mathbf{0.50}$ & $\mathbf{0.49}$ & $\mathbf{0.85}$ & 0.43 & $\mathbf{0.65}$ \\
\;\;CPT & $2.63\times10^{-5}$ & 4.0 & 0.74 & 0.66 & 0.61 & 0.41 & 0.70 & 0.94 & $\mathbf{0.77}$ & 0.44 & $\mathbf{0.81}$ & 0.47 & 0.48 & 0.79 & 0.40 & 0.63 \\
\;\;CPT & $10^{-6}$ & 3.5 & 0.76 & 0.67 & 0.64 & $\mathbf{0.41}$ & 0.71 & 0.94 & 0.75 & $\mathbf{0.45}$ & 0.80 & 0.48 & 0.48 & 0.82 & $\mathbf{0.43}$ & 0.64 \\
\midrule
OLMo2-7B & --- & --- & 0.81 & 0.80 & 0.74 & 0.46 & 0.79 & 0.97 & 0.82 & 0.54 & 0.90 & 0.64 & 0.56 & 0.92 & 0.58 & 0.73 \\
\;\;CPT & $10^{-6}$ & 4.0 & $\mathbf{0.81}$ & $\mathbf{0.79}$ & $\mathbf{0.75}$ & $\mathbf{0.48}$ & $\mathbf{0.84}$ & $\mathbf{0.96}$ & $\mathbf{0.83}$ & $\mathbf{0.54}$ & $\mathbf{0.90}$ & $\mathbf{0.58}$ & $\mathbf{0.54}$ & $\mathbf{0.88}$ & $\mathbf{0.57}$ & $\mathbf{0.73}$ \\
\bottomrule
\end{tabular}%
}
\end{table}

\begin{figure}[t]
    \centering
    \includegraphics[width=0.99\linewidth]{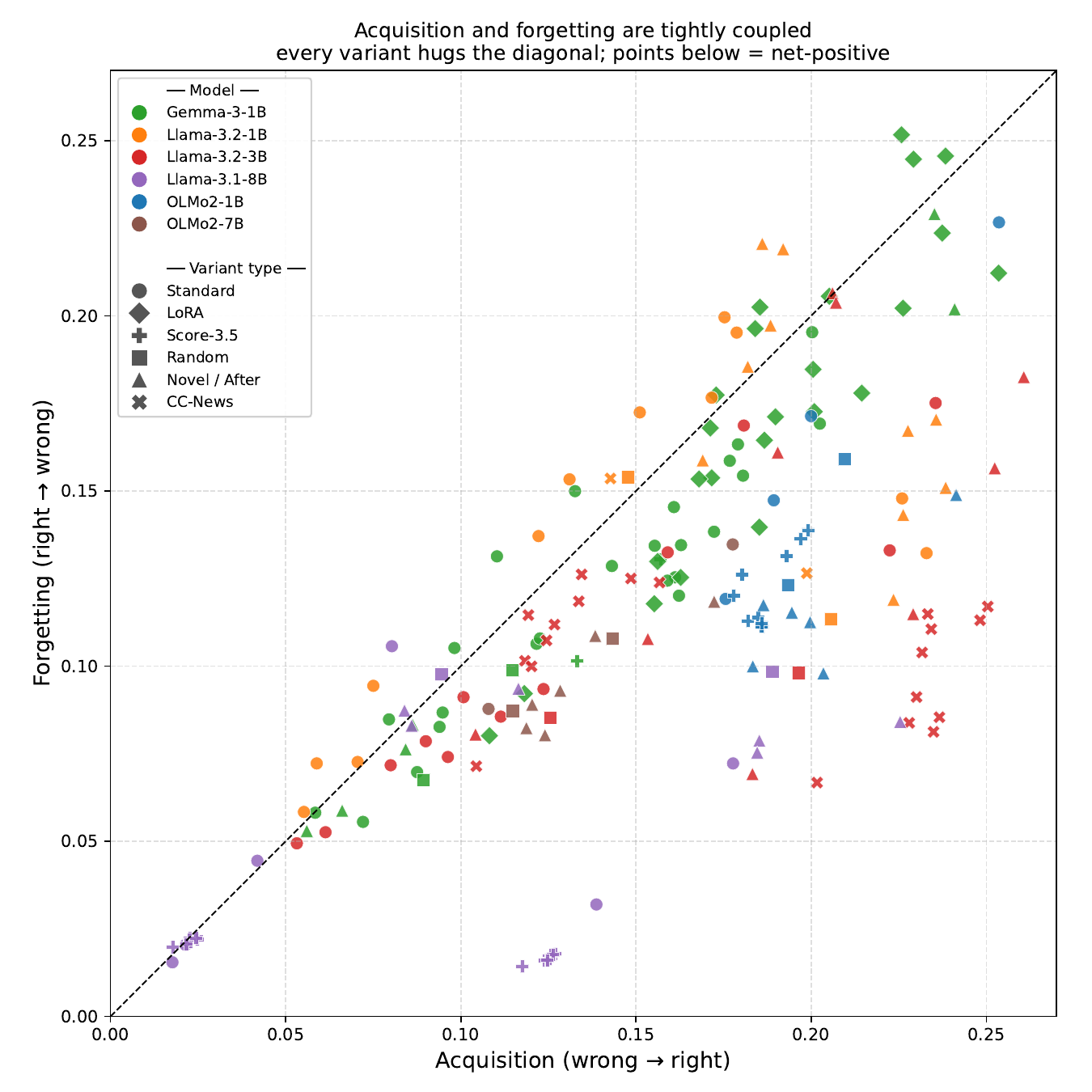}
    \caption{Acquisition and forgetting are coupled. Each point is one trained checkpoint, spanning models, learning rates, data slices and LoRA ranks; every variant hugs the diagonal, and points below it are net-positive.}
    \label{fig:coupling}
    \vspace{-1.5em}
\end{figure}

\begin{figure}
    \centering
    \includegraphics[width=0.995\linewidth]{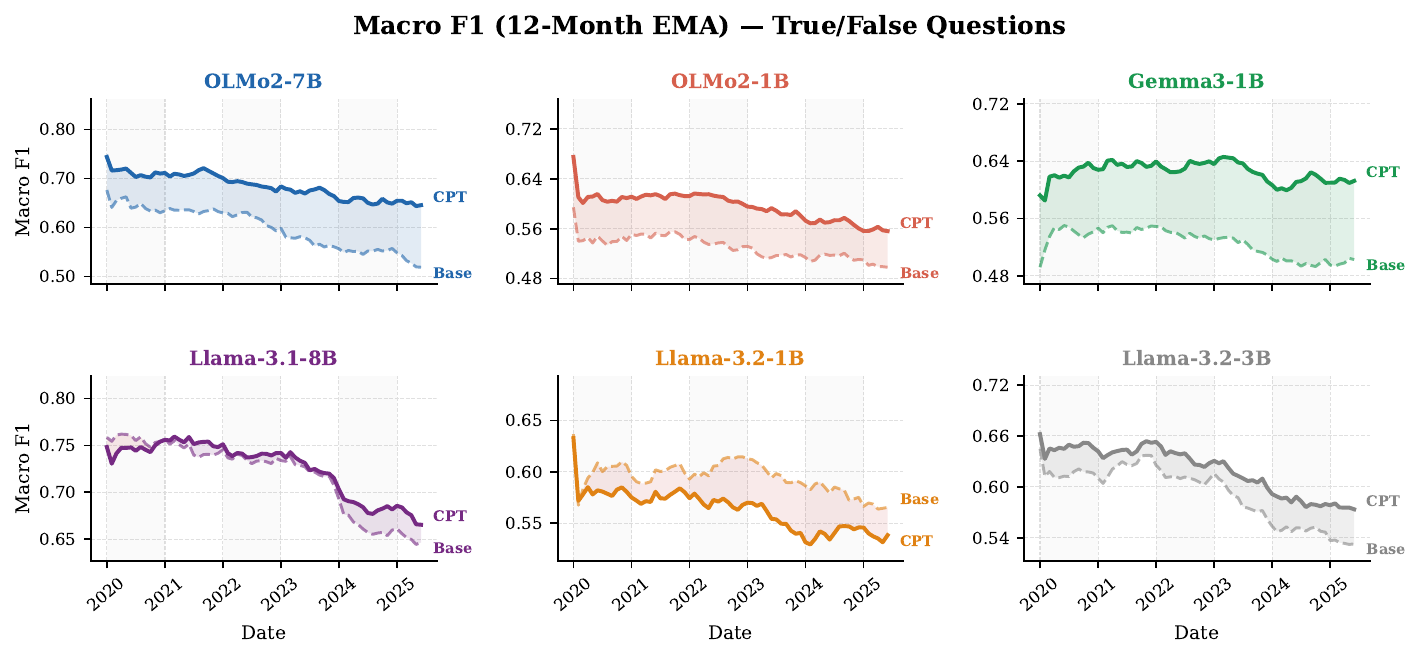}
    \includegraphics[width=0.995\linewidth]{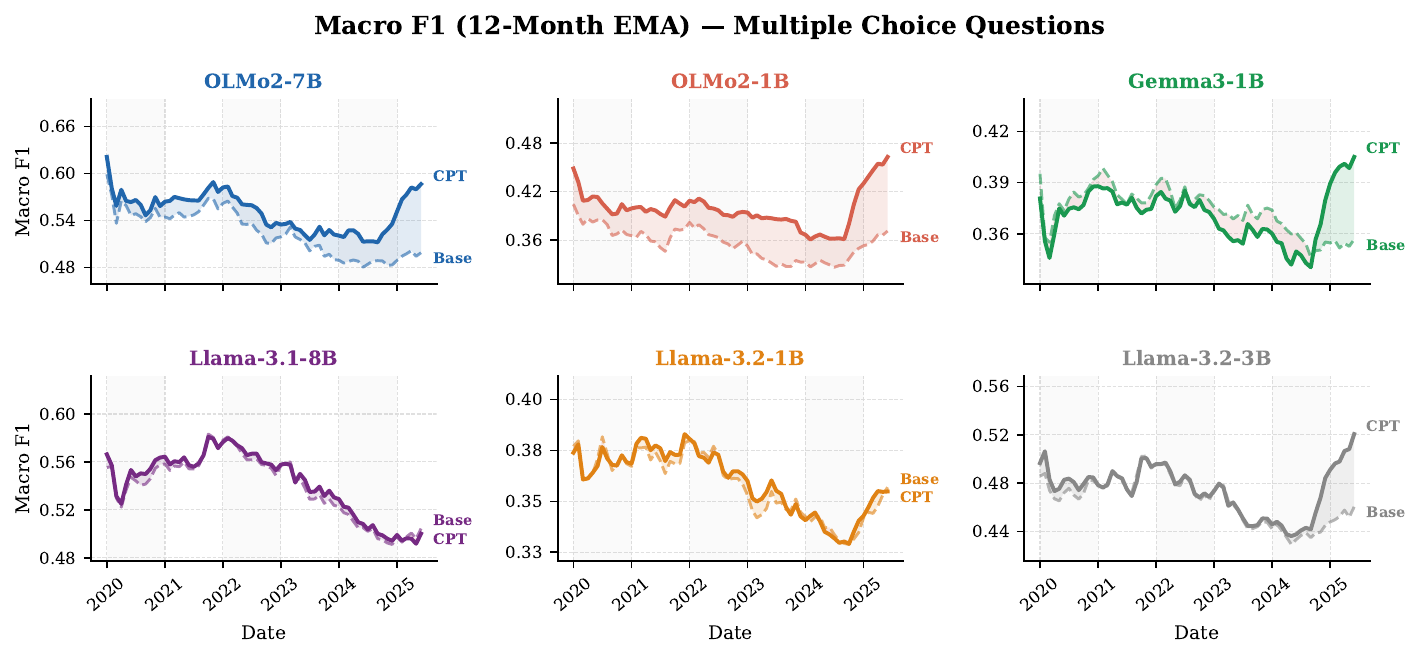}
    \caption{The performance advantage of CPT over the base model widens as questions approach the present. Macro-F1 scores on Daily Oracle (12-month EMA) plotted by the timestamp of the source news article show that CPT models increasingly outperform their base counterparts over time, with the gap most pronounced for plastic families (\olmotwo, \gemmathreeoneb). Complements Section~\ref{sec:findings-know-acq}.}
    \label{fig:over-time}
\end{figure}

\begin{figure}
    \centering
    \includegraphics[width=0.995\linewidth]{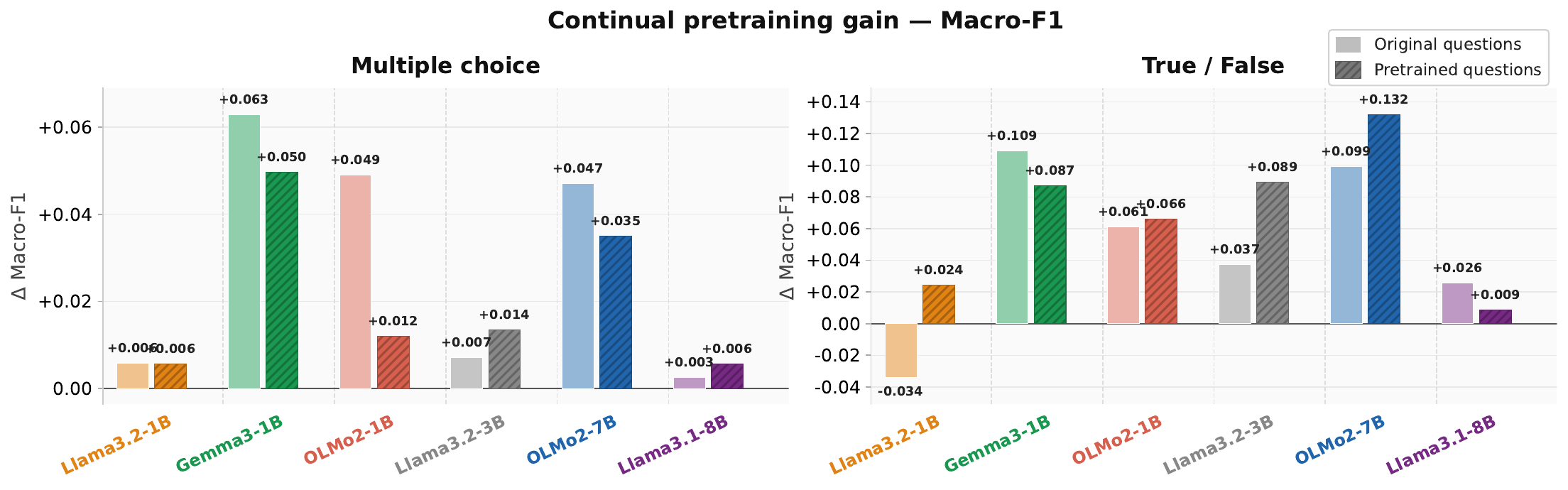}
    \includegraphics[width=0.995\linewidth]{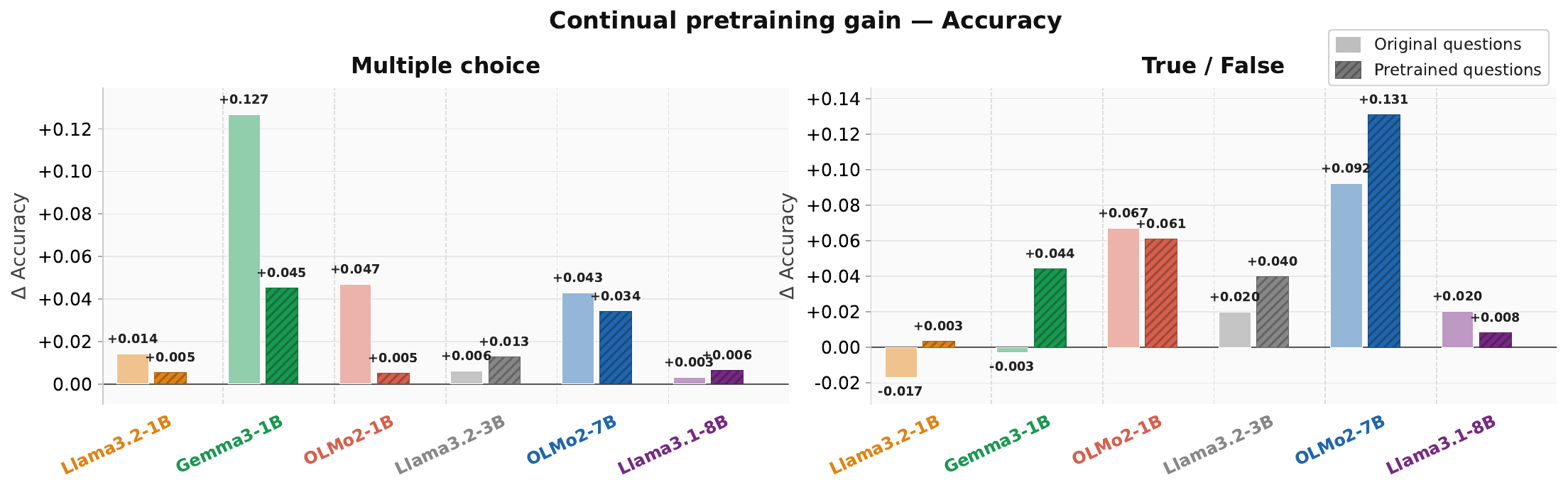}
    \caption{CPT improves performance on questions that come solely from the news articles from the pretraining corpus. Macro-F1 scores on these questions show similar performance to the questions extracted from news that are not necessarily in the corpus. We conclude that CPT models generalize well to unseen temporal knowledge.}
    \label{fig:new-questions}
\end{figure}

\begin{figure}
    \centering
    \includegraphics[width=0.49\linewidth]{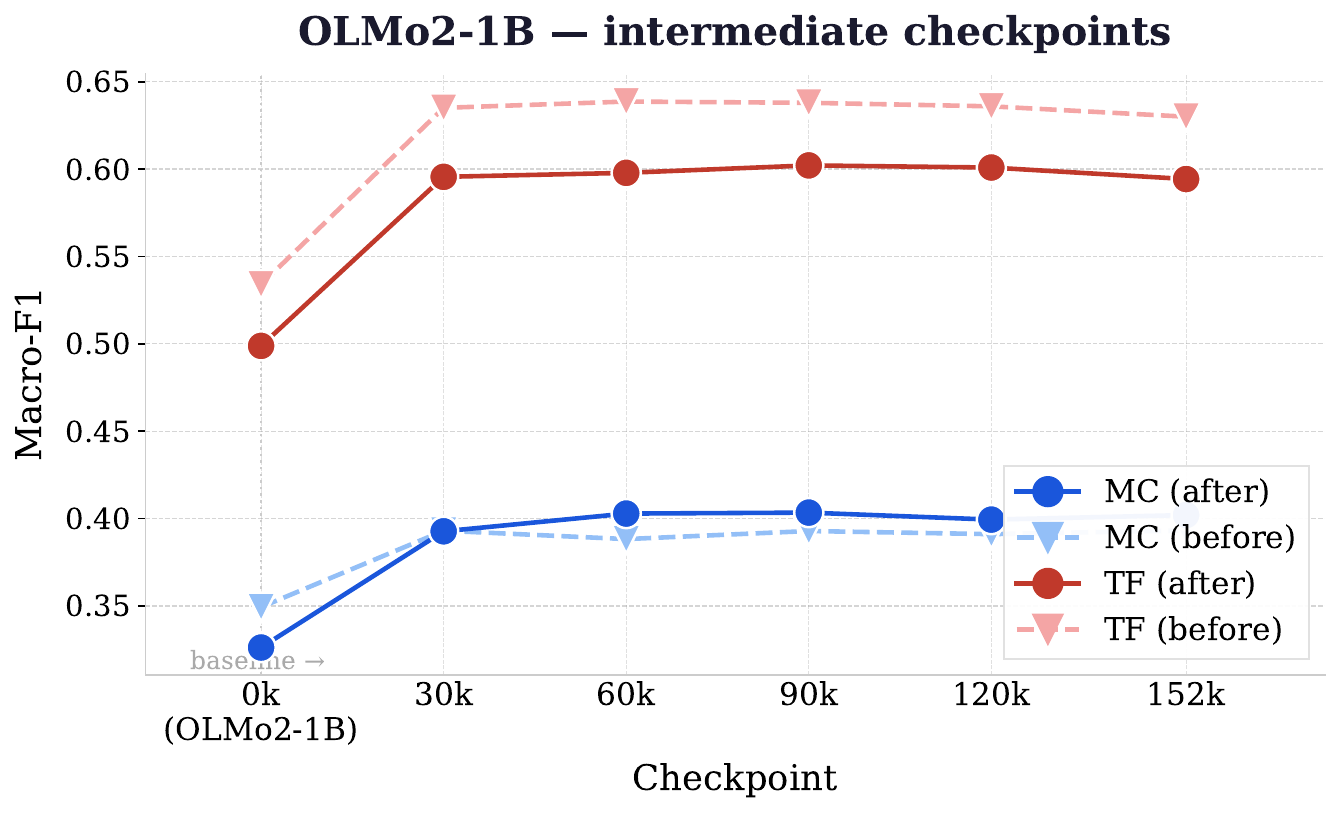} \hfill
    \includegraphics[width=0.49\linewidth]{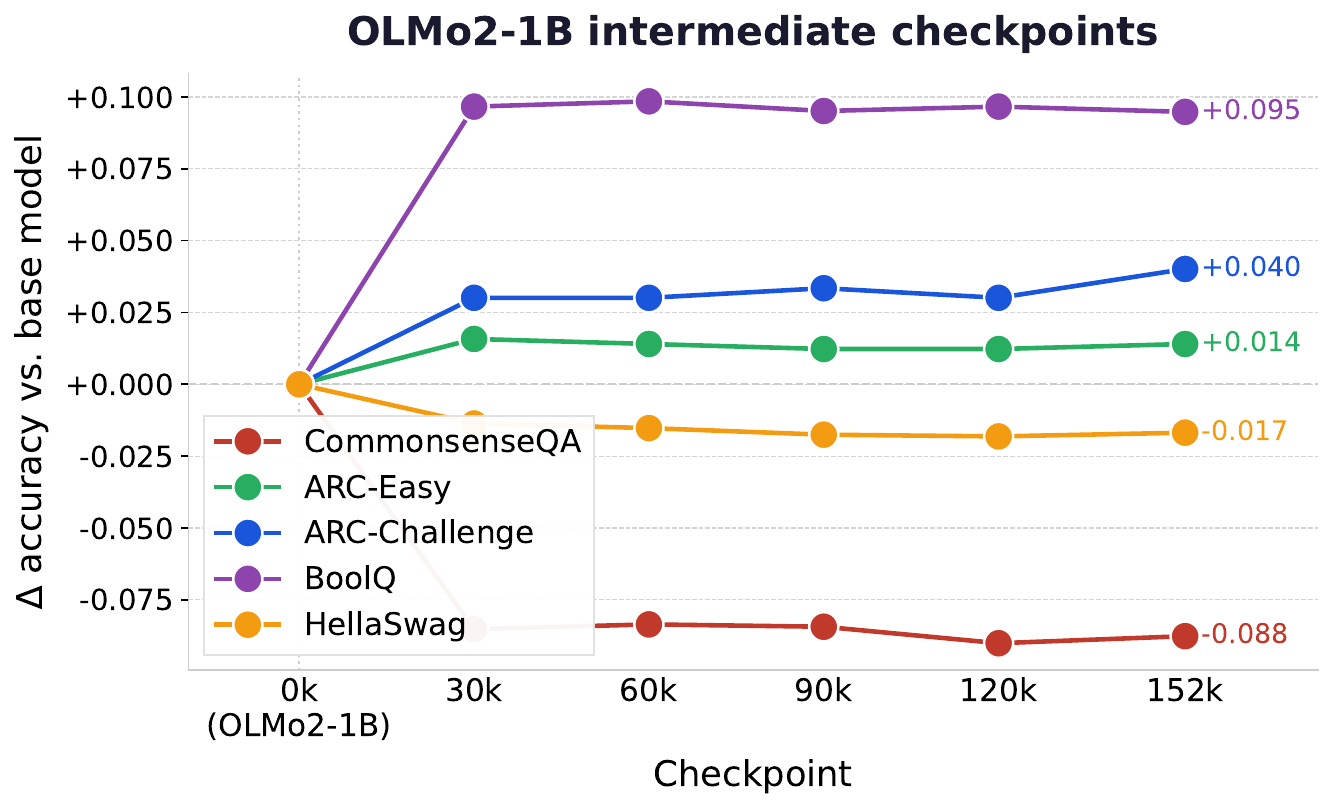}
    \includegraphics[width=0.49\linewidth]{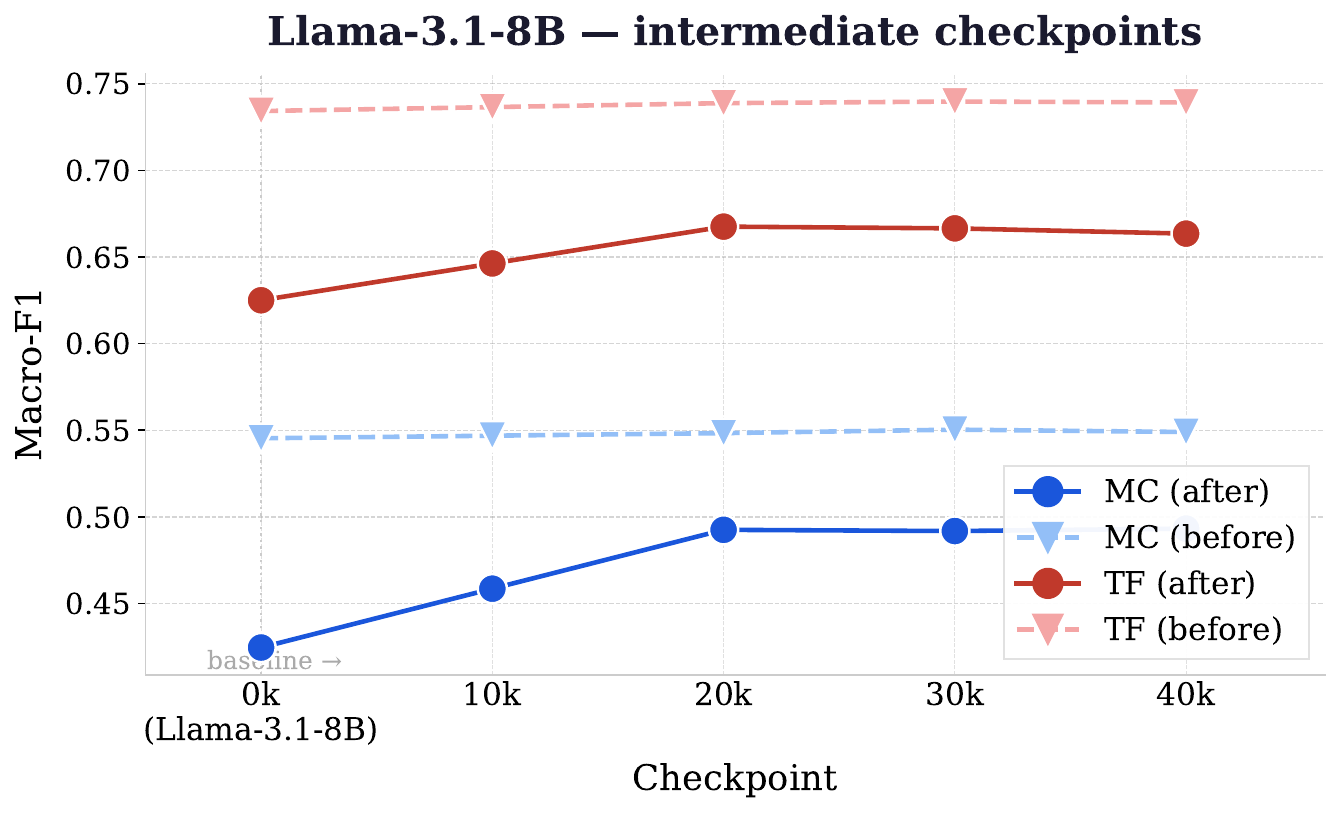} \hfill
    \includegraphics[width=0.49\linewidth]{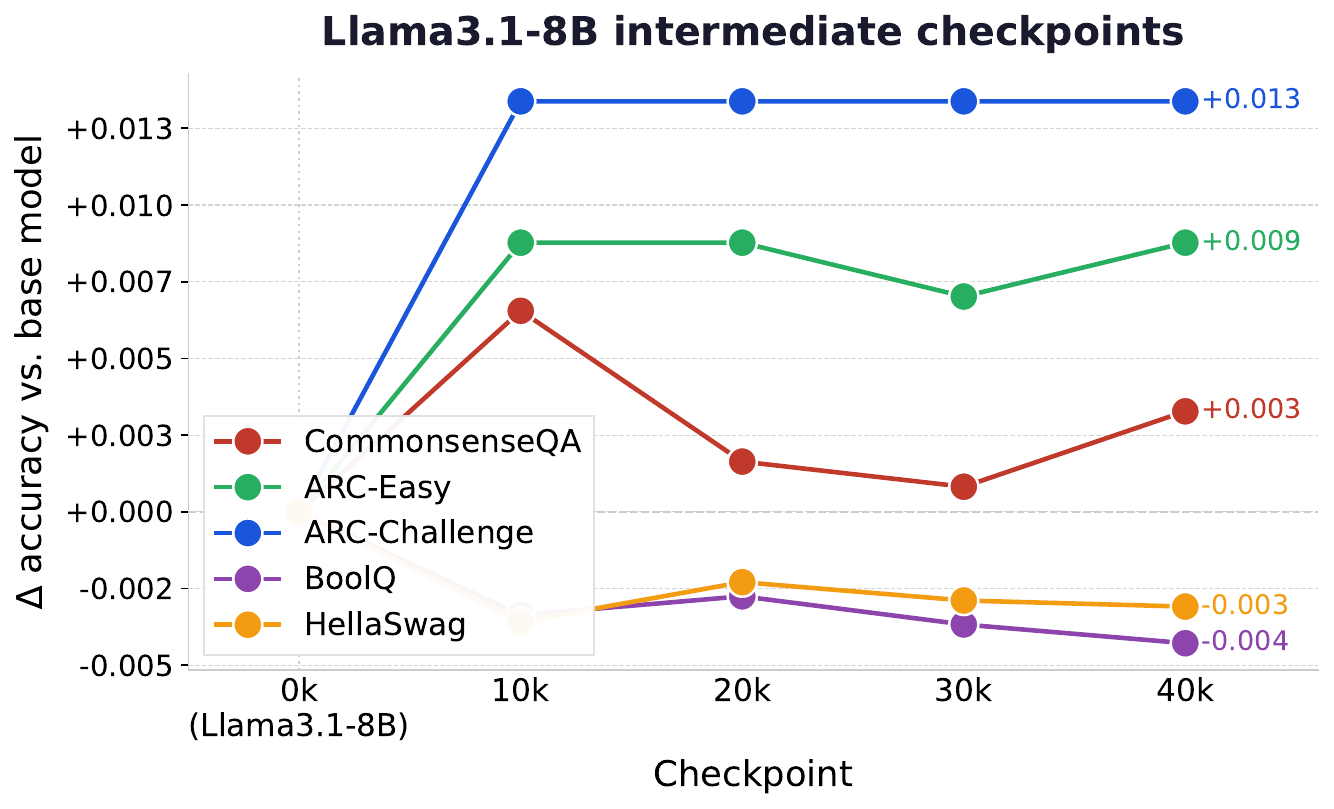}
    \caption{CPT converges rapidly on both knowledge acquisition and downstream tasks. Macro-F1 on Daily Oracle and accuracy on selected downstream benchmarks for \olmotwooneb (top) and \llamathreeoneeightb (bottom) across intermediate checkpoints. Both metrics plateau within the first third of training, suggesting that short, targeted CPT runs may suffice for effective temporal updates.}
    \label{fig:acc-vs-checkpoints}
\end{figure}

\begin{figure}
    \centering
    \includegraphics[width=0.98\linewidth, trim=0 0pt 0 0, clip]{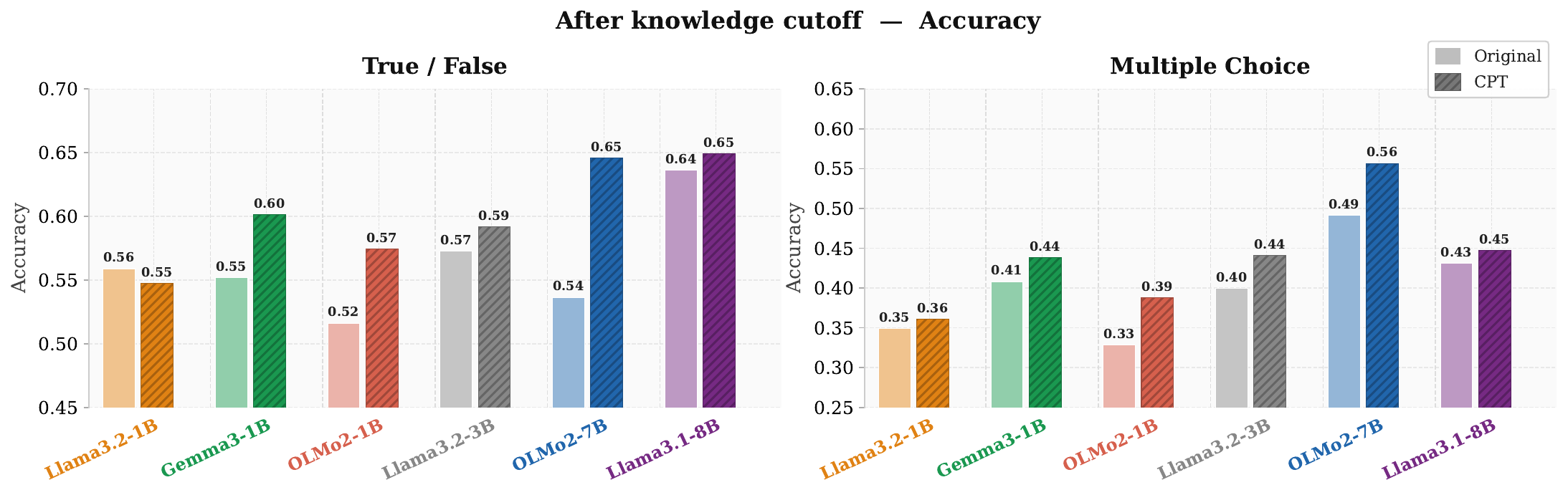}
    \caption{Accuracy counterpart to Figure~\ref{fig:main-avg} (after-cutoff split). Results confirm the family-level heterogeneity in knowledge acquisition: \olmotwo and \gemmathreeoneb models gain substantially on post-cutoff questions while \llama variants show limited improvement.}
    \label{fig:main-avg-supp}
\end{figure}

\begin{figure}
    \centering
    \includegraphics[width=0.98\linewidth, trim=0 0pt 0 0, clip]{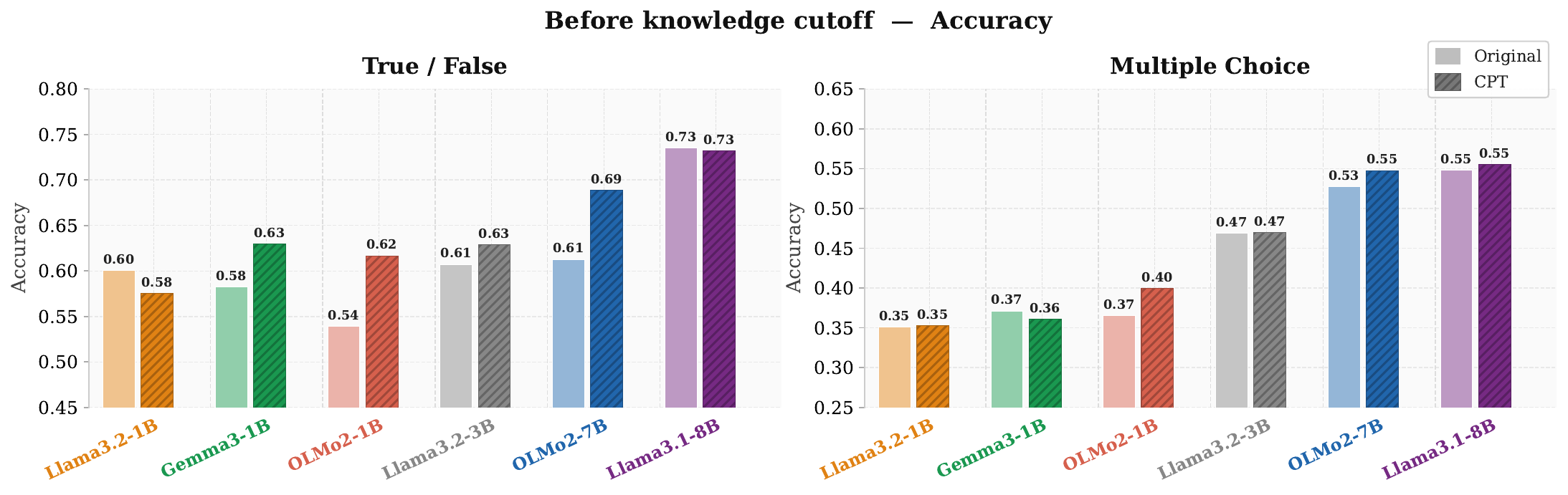}
    \caption{Accuracy counterpart to Figure~\ref{fig:main-avg} (before-cutoff split). Positive backward transfer is consistently observed across model families, with CPT improving or preserving accuracy on pre-cutoff questions for five of six models.}
    \label{fig:main-avg2-supp}
\end{figure}

\begin{figure}
    \centering
    \includegraphics[width=0.49\linewidth]{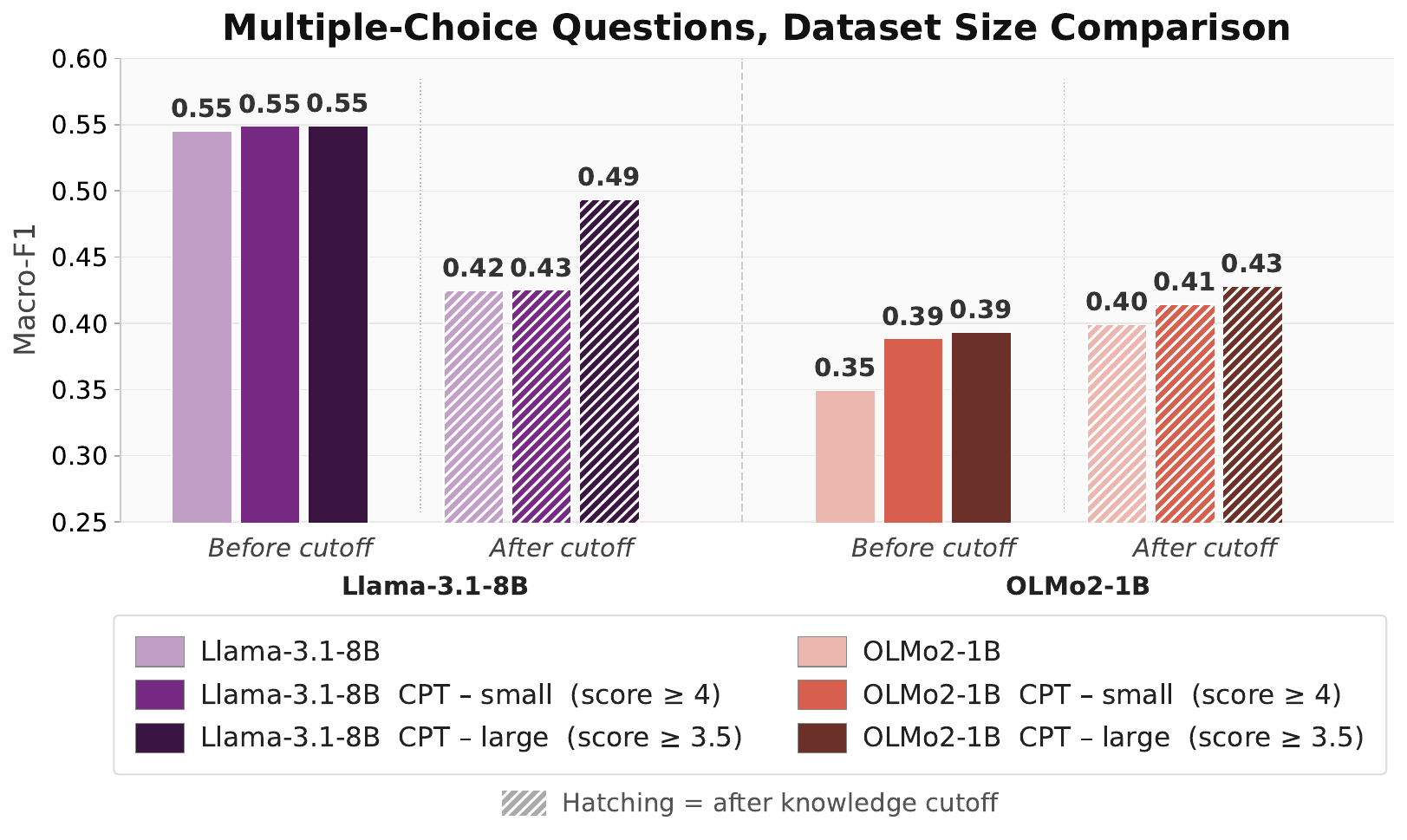}  \hfill
    \includegraphics[width=0.49\linewidth]{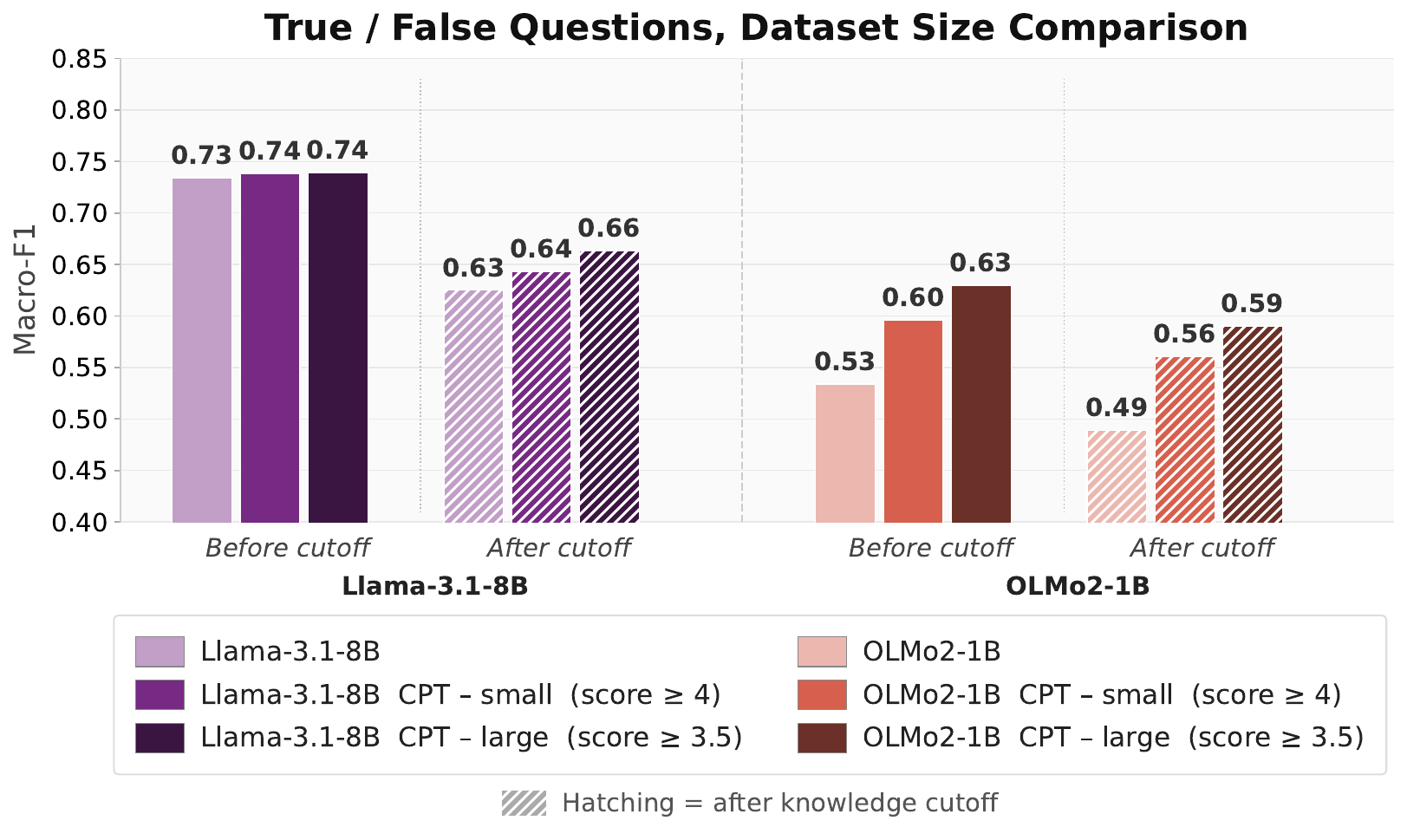}
    \includegraphics[width=0.49\linewidth]{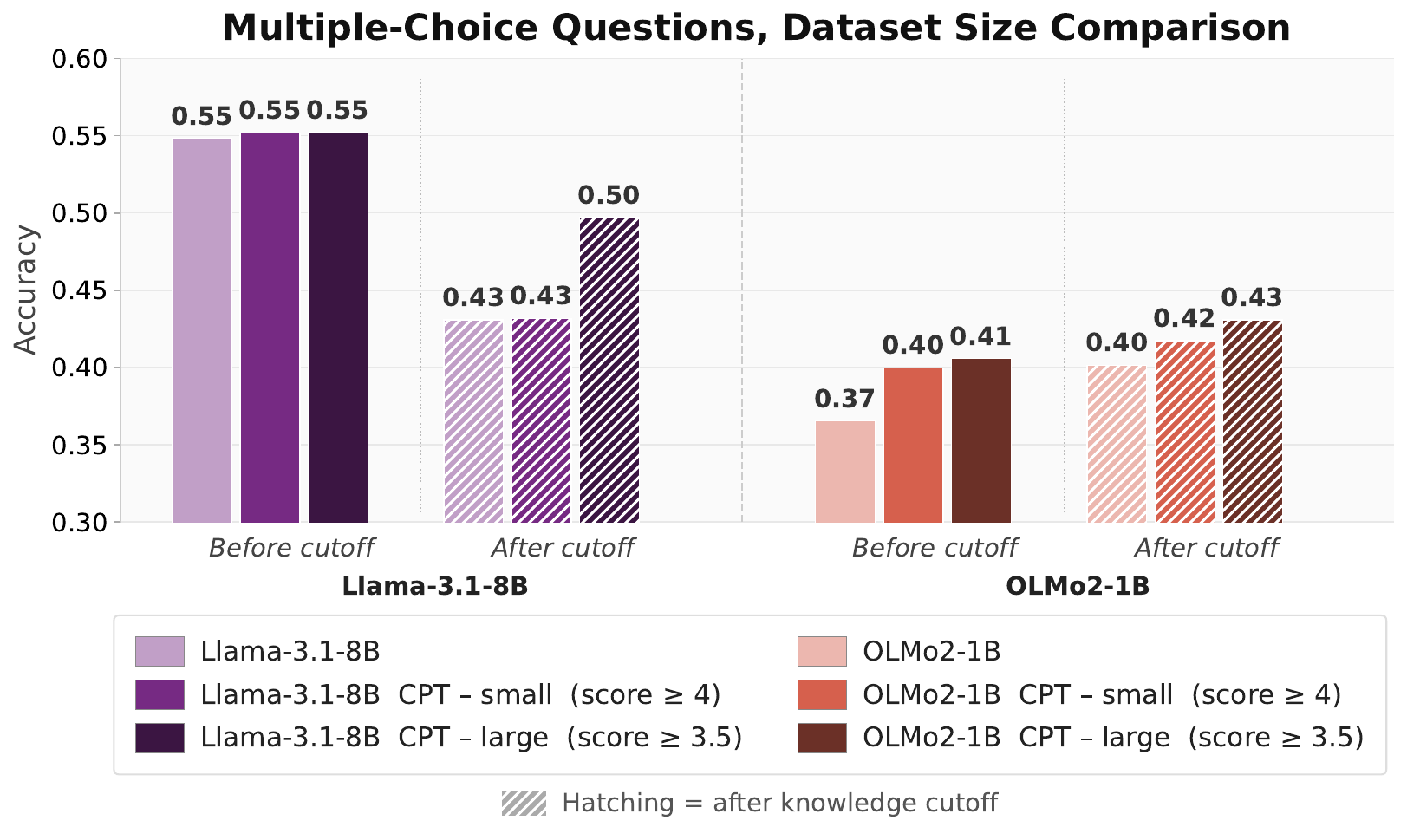}  \hfill
    \includegraphics[width=0.49\linewidth]{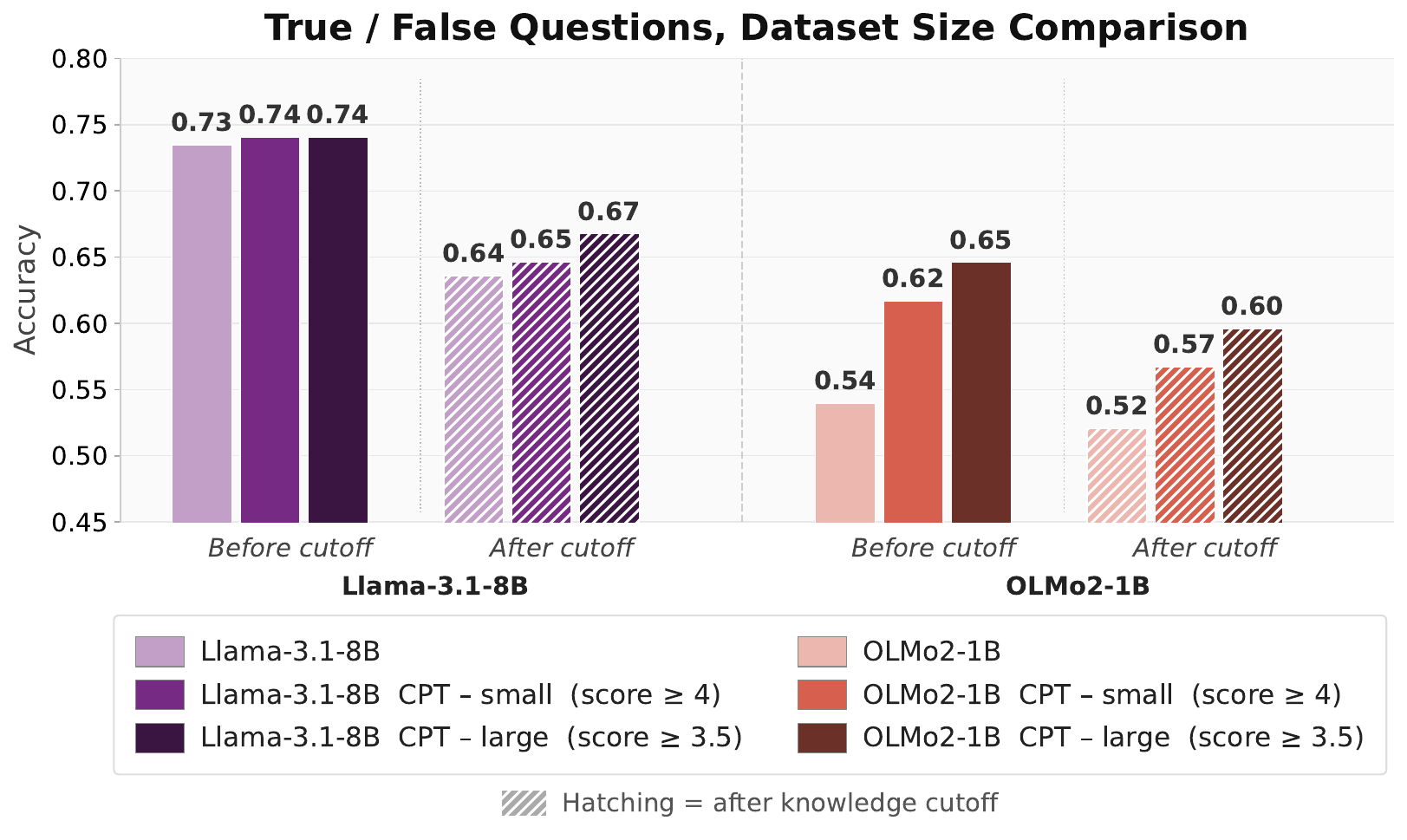}
    \caption{Accuracy and Multiple-Choice counterparts to Figure~\ref{fig:data-size}. Across both metrics and question formats, data quality dominates quantity: the high-quality 6B-token slice captures most of the available CPT signal, while the larger 40B-token slice yields only marginal additional gains and amplifies trade-offs on downstream tasks for plastic models.}
    \label{fig:data-size-supp}
\end{figure}

\begin{figure}
    \centering
    \includegraphics[width=0.53\linewidth]{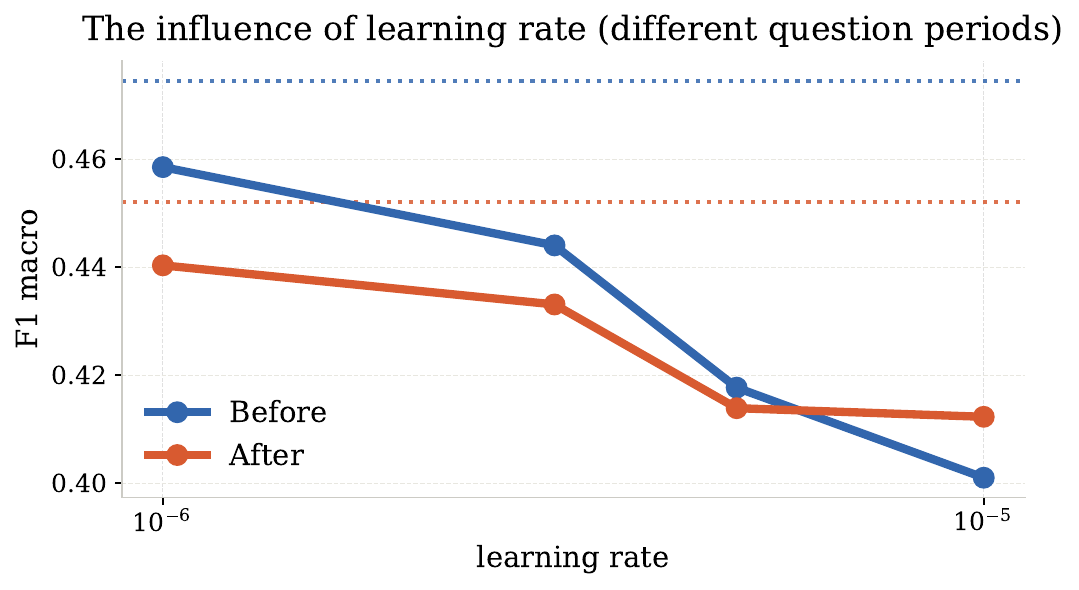}
    \caption{Learning rate modulates the stability--plasticity trade-off in CPT: lower rates preserve stability at the cost of limited plasticity. As expected, higher rates cause a larger degradation on the before-cutoff questions than after-cutoff since the pretraining set involves content about after-cutoff questions that the model has never seen before. These findings are aligned with Figure~\ref{fig:lr}.}
    \label{fig:lr-supp}
\end{figure}

\begin{figure}
    \centering
    \includegraphics[width=\linewidth]{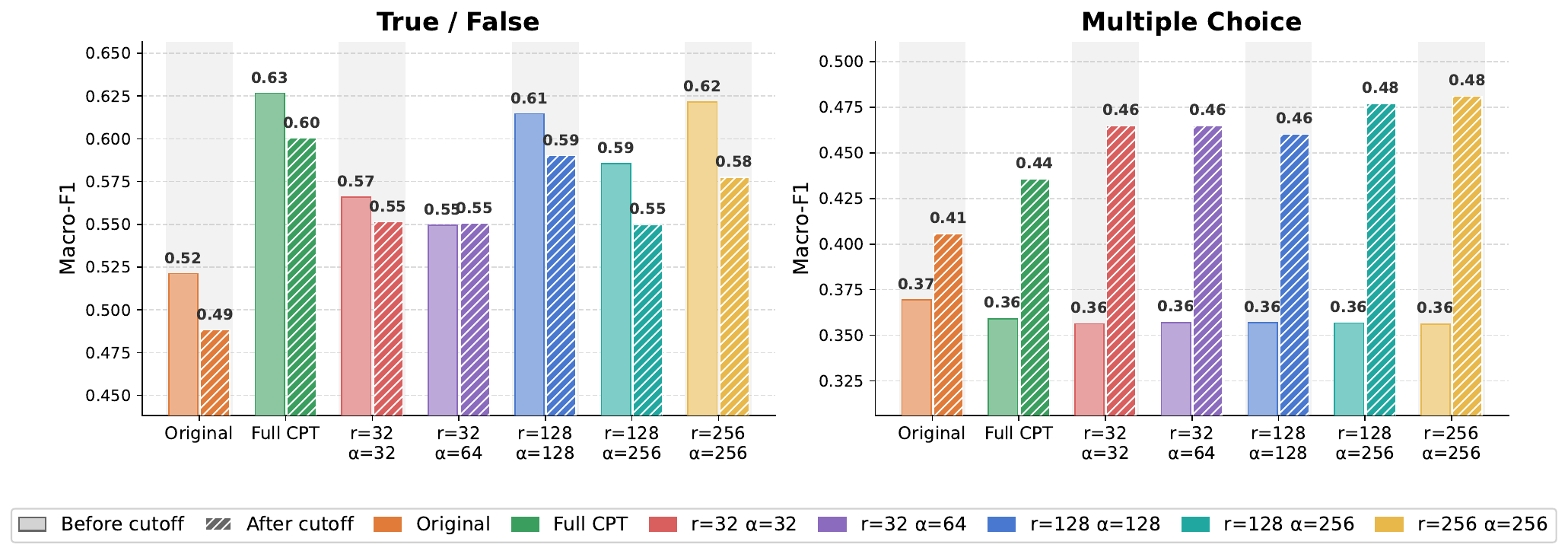}
    \includegraphics[width=\linewidth]{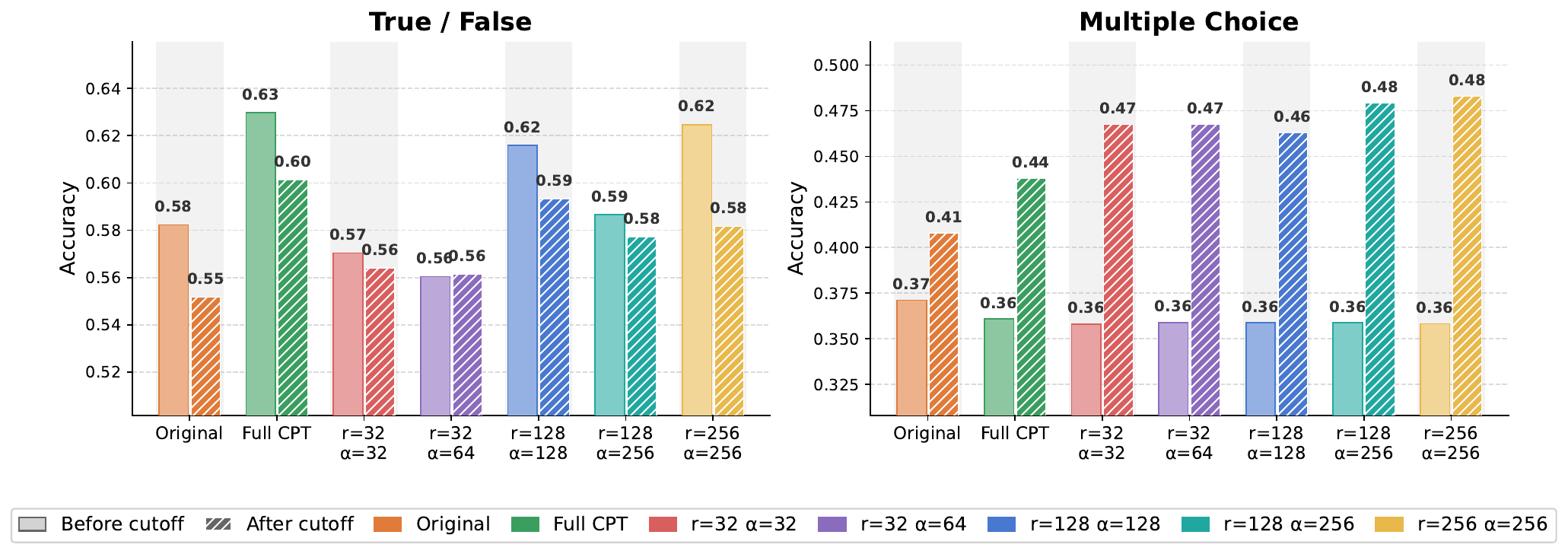}
    \caption{Separate True/False and Multiple-Choice breakdowns of the LoRA results summarised in Figure~\ref{fig:metric-lora}. Higher-rank configurations ($r{=}128$ and $r{=}256$) match or exceed full CPT on after-cutoff Macro-F1 across both question formats, while before-cutoff scores remain stable throughout.}
    \label{fig:metric-lora-supp}
\end{figure}

\begin{figure}
    \centering
    \includegraphics[width=\linewidth]{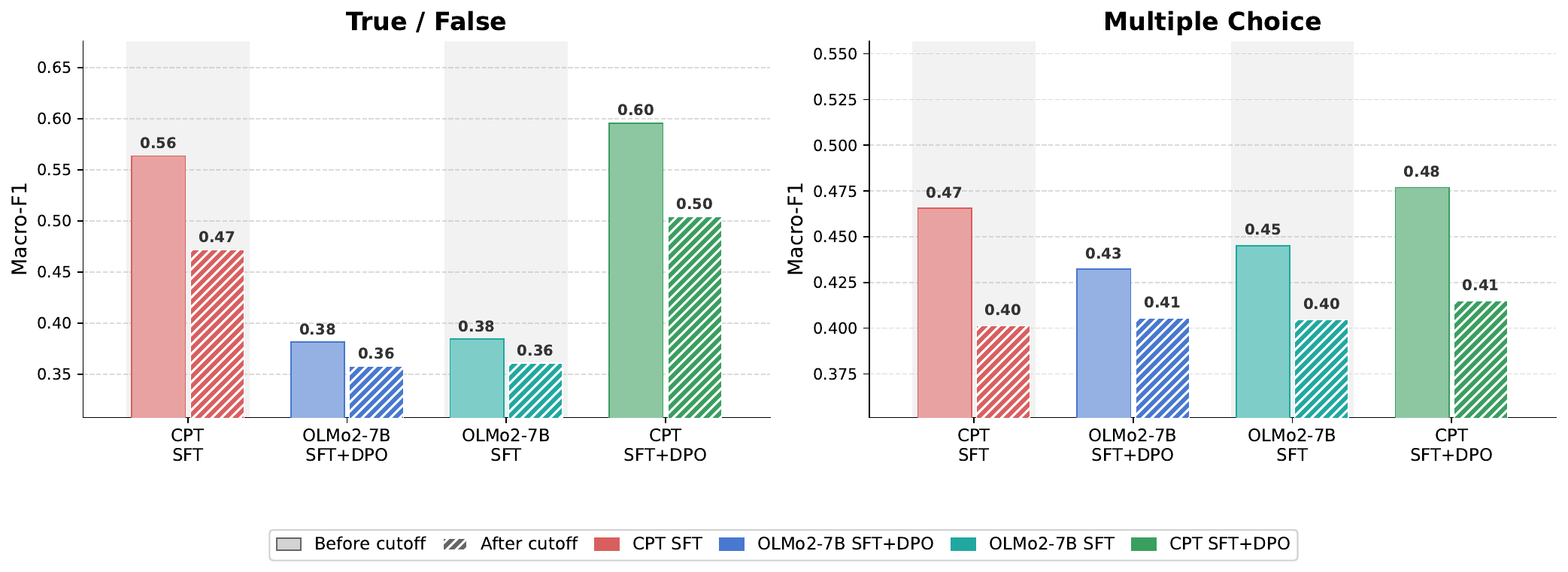}
    \includegraphics[width=\linewidth]{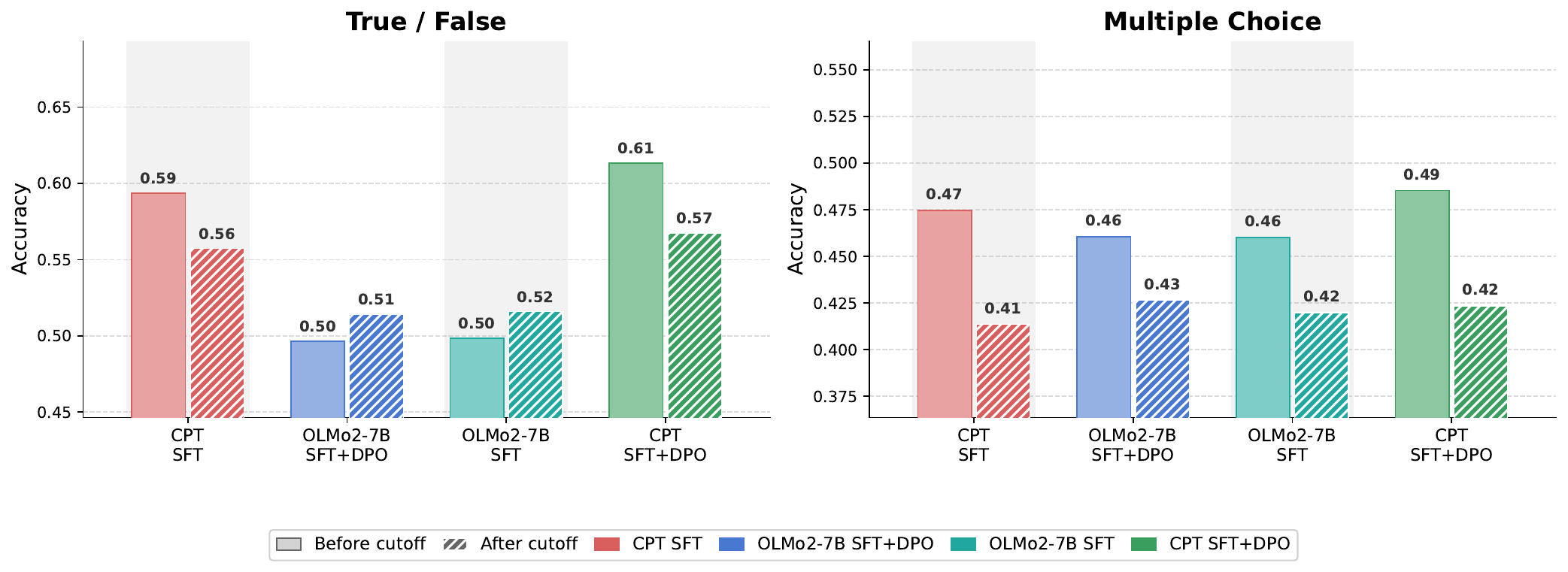}
    \caption{Separate True/False and Multiple-Choice breakdowns of the post-training results summarised in Figure~\ref{fig:post-train}. SFT and SFT+DPO variants consistently outperform the \olmotwosevenb baseline across both question formats.}
    \label{fig:post-train-olmo-supp}
\end{figure}

\begin{figure}
    \centering
    \includegraphics[width=.49\linewidth]{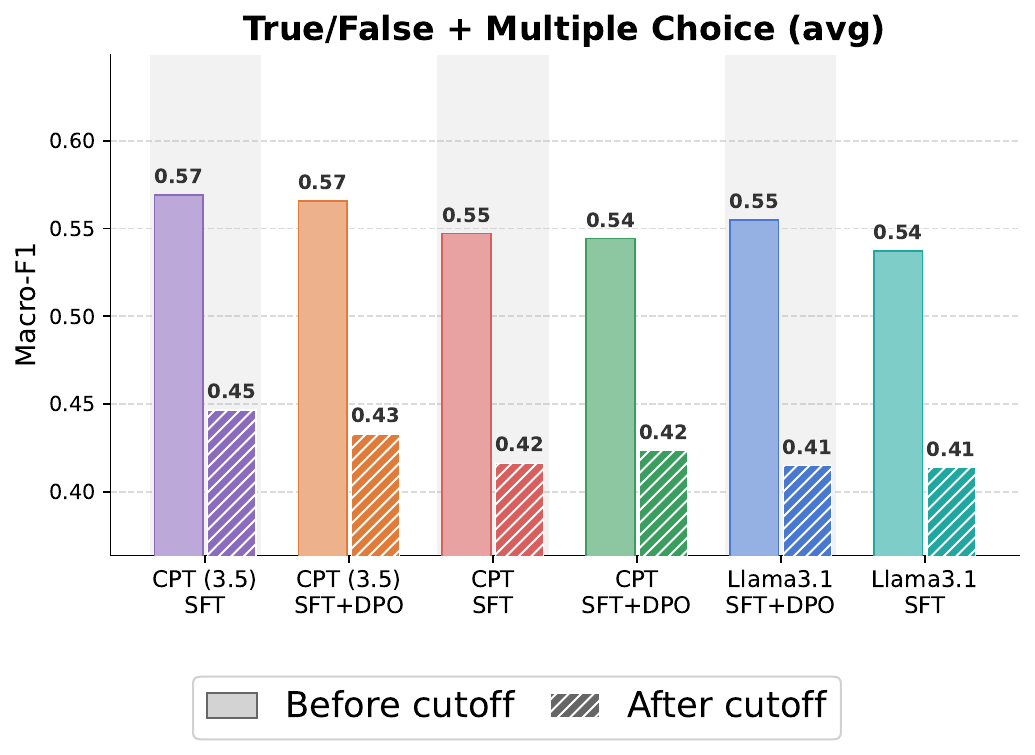} \hfill
    \includegraphics[width=.49\linewidth]{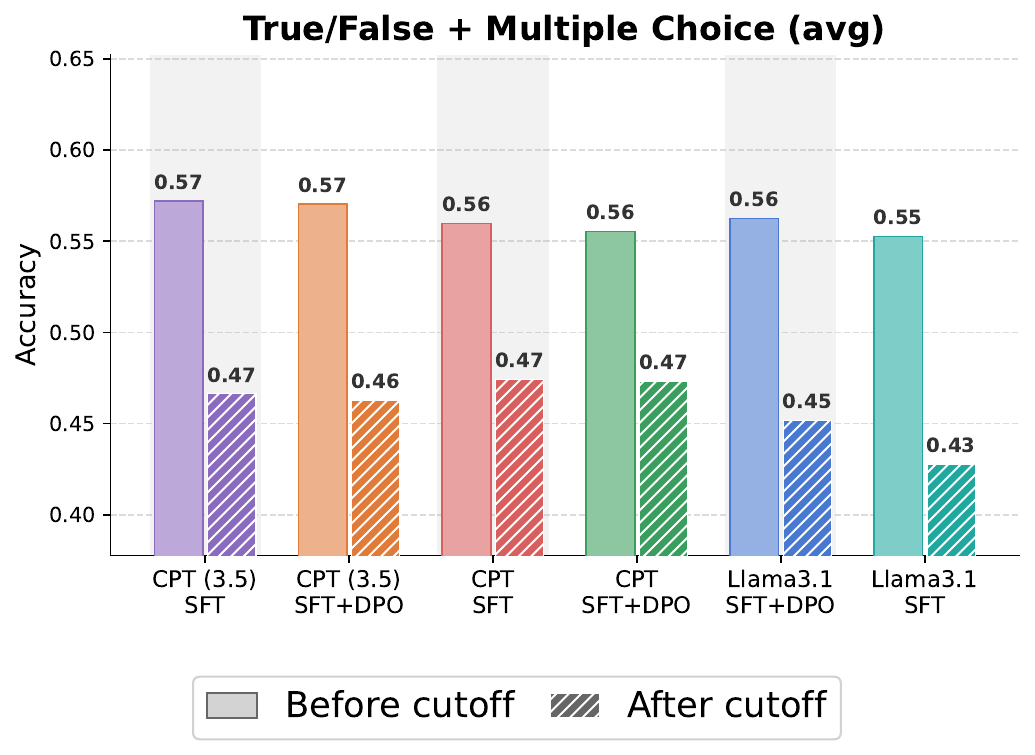} \\
    \vspace*{1cm}
    \includegraphics[width=\linewidth]{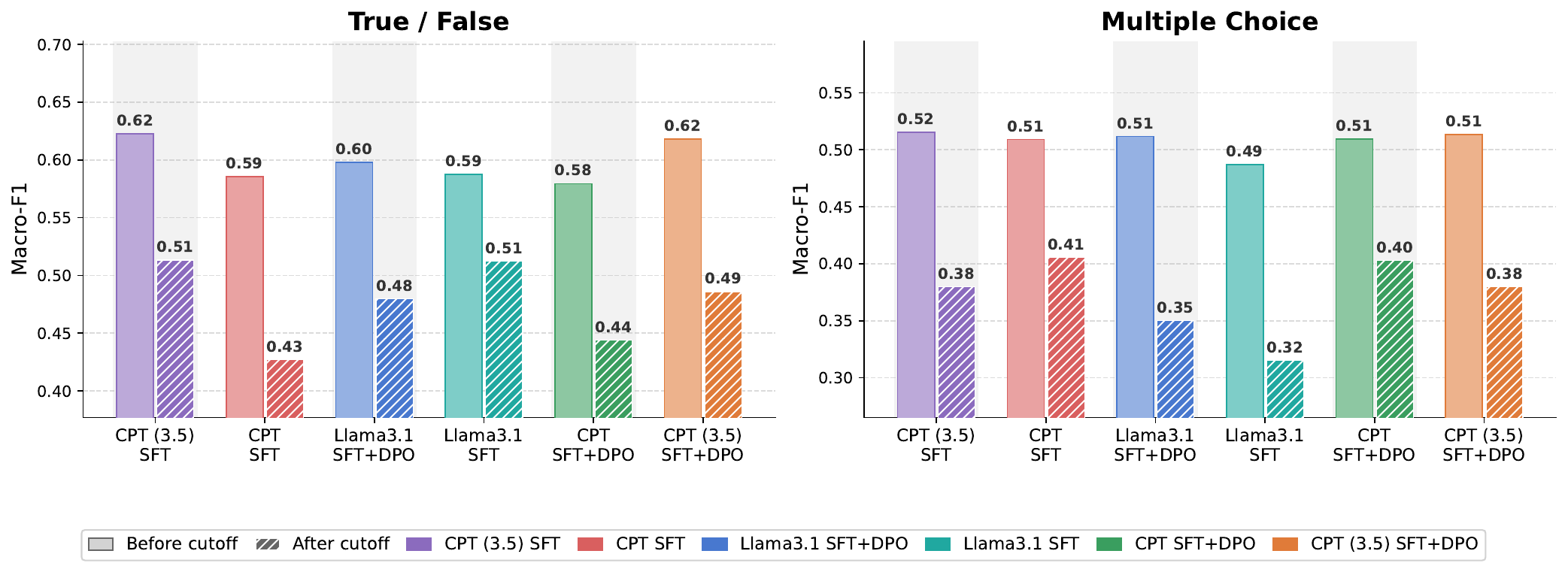} 
    \includegraphics[width=\linewidth]{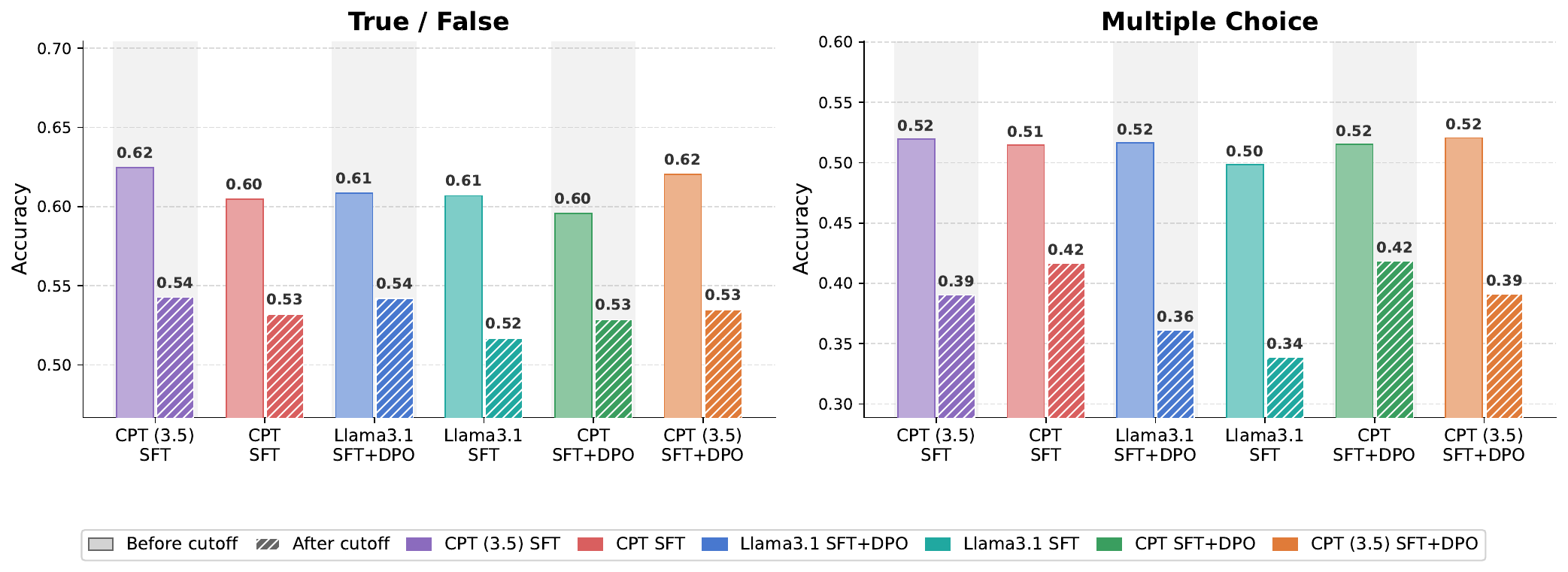}
    \caption{Daily Oracle performance of \llamathreeoneeightb model and its continued pretrained counterpart after SFT and DPO. We observe that CPT equips the model with knowledge that post-training preserves, leading to increased scores. CPT+SFT variants consistently outperform the \llamathreeoneeightb SFT baseline on after-cutoff questions across both formats, while adding DPO on top of CPT+SFT degrades performance in both splits.}
    \label{fig:post-train-llama-supp}
\end{figure}

%%%%%%%%%%%%%%%%%%%%%%%%%%%%%%%%%%%%%%%%%%%%%%%%%%%%%%%%%%
%%%%%%%%%%%%%%%  NEW APPENDIX MATERIAL  %%%%%%%%%%%%%%%%%%
%%%%%%%%%%%%%%%%%%%%%%%%%%%%%%%%%%%%%%%%%%%%%%%%%%%%%%%%%%
\clearpage

\section{Daily Oracle split statistics}
\label{app:split-stats}

Table~\ref{tab:split-stats} reports the per-class sample counts of the before- and after-cutoff partitions of Daily Oracle. The full question pool spans monthly slices from 2020-01 through 2025-06. The counts use the December 2023 boundary, which is the knowledge cutoff of the five \olmotwo and \llama models. \gemmathreeoneb has a later cutoff (August 2024), so its split is computed against that boundary instead, which yields TF before/after $= 15{,}496 / 2{,}821$ and MC before/after $= 13{,}374 / 3{,}112$. Both formats are close to class-balanced on the after-cutoff split; the before-cutoff TF split is the most imbalanced (53.4\% \emph{Yes}), which motivates our use of Macro-F1 rather than accuracy as the headline metric (Section~\ref{sec:findings}).

\begin{table}[htbp]
    \centering
    \caption{Per-class sample counts of the Daily Oracle splits, using the December 2023 cutoff.}
    \label{tab:split-stats}
    \small
    \begin{tabular}{l cc c c cccc c}
        \toprule
        & \multicolumn{3}{c}{\textbf{True/False}} & & \multicolumn{5}{c}{\textbf{Multiple Choice}} \\
        \cmidrule(lr){2-4} \cmidrule(lr){6-10}
        Split & Yes & No & Total & & A & B & C & D & Total \\
        \midrule
        Before-cutoff ($\leq$ 2023-12) & 7{,}175 & 6{,}267 & 13{,}442 & & 2{,}898 & 2{,}871 & 2{,}990 & 2{,}864 & 11{,}623 \\
        After-cutoff ($\geq$ 2024-01)  & 2{,}434 & 2{,}441 & \phantom{0}4{,}875 & & 1{,}212 & 1{,}179 & 1{,}199 & 1{,}273 & \phantom{0}4{,}863 \\
        \bottomrule
    \end{tabular}
\end{table}

\section{Pretraining saturation and CPT gains}
\label{app:saturation}

Section~\ref{sec:findings-know-acq} argues that plasticity tracks how saturated a checkpoint already is rather than its parameter count. Table~\ref{tab:saturation} makes the axis explicit, reporting for each model the parameter count $N$, the original pretraining-token budget $D$, the tokens-per-parameter ratio $D/N$, the approximate pretraining compute $6ND$, and the observed Macro-F1 gains on both splits. Rows are sorted by $D/N$.

\begin{table}[htbp]
    \centering
    \caption{Pretraining saturation versus CPT gain, sorted by tokens-per-parameter $D/N$. Setting \llamathreeoneeightb aside, the after-cutoff gains fall monotonically from the least-saturated to the most over-trained checkpoint.}
    \label{tab:saturation}
    \small
    \begin{tabular}{l rrrr rr}
        \toprule
        Model & $N$ & $D$ & $D/N$ & $6ND$ & $\Delta$ after-cutoff & $\Delta$ before-cutoff \\
        \midrule
        \olmotwosevenb       & 7.30B & $\sim$4.05T & $\sim$555\phantom{0,} & 1.8e23 & $+0.09$ & $+0.06$ \\
        \llamathreeoneeightb & 8.03B & 15T\phantom{$\sim$0.00} & $\sim$1{,}870 & 7.2e23 & $+0.02$ & $\phantom{+}0.00$ \\
        \gemmathreeoneb      & 1.00B & 2T\phantom{$\sim$00.00} & $\sim$2{,}000 & 1.2e22 & $+0.07$ & $+0.04$ \\
        \olmotwooneb         & 1.48B & $\sim$4.05T & $\sim$2{,}740 & 3.6e22 & $+0.05$ & $+0.05$ \\
        \llamathreetwothreeb & 3.21B & 9T\phantom{$\sim$00.00} & $\sim$2{,}800 & 1.7e23 & $+0.04$ & $+0.02$ \\
        \llamathreetwooneb   & 1.24B & 9T\phantom{$\sim$00.00} & $\sim$7{,}260 & 6.7e22 & $-0.01$ & $-0.01$ \\
        \bottomrule
    \end{tabular}
\end{table}

Setting \llamathreeoneeightb aside, the gains fall monotonically from the least-saturated checkpoint (\olmotwosevenb) to the most over-trained one (\llamathreetwooneb, which regresses); the rank correlation between $D/N$ and the overall gain is $\rho = -0.63$. Pretraining compute does not organize the spread ($6ND$ has correlation $0.20$ with the gain), which is why we frame the hypothesis in terms of tokens per parameter rather than total compute. Distance from compute-optimal, a natural alternative proxy, is $D/N$ divided by a constant and therefore gives the same ordering by construction.

Two caveats bound this reading. First, $D/N$ covaries with architecture, tokenizer and data quality, and we claim no causal isolation. Both within-family pairs hold $D$ fixed while varying $N$ --- \olmotwo at $\sim$4.05T ($+0.09$ vs.\ $+0.05$) and \llamathreetwo at 9T ($+0.04$ vs.\ $-0.01$) --- and both are consistent with the hypothesis, but neither separates saturation from parameter count, since $N$ covaries with $D/N$ by construction. Second, at $n=6$ the correlation is not significant ($p \approx 0.16$), and the \olmotwooneb\,/\,\llamathreetwothreeb gap ($+0.05$ vs.\ $+0.04$) lies within the $\pm 0.015$ rerun variation reported in Section~\ref{sec:findings}. We therefore state this as an ordinal relationship that holds across three families with one exception, not as an established quantitative law.

\section{Temporal holdout: questions postdating the training corpus}
\label{app:temporal-holdout}

The memorization probe in Section~\ref{sec:findings-know-acq} does not rule out event-level memorization from related articles covering the same news. To close that gap we extracted an additional question set following the Daily Oracle protocol, drawn exclusively from pages published \emph{after} our last training dump. Since the underlying events postdate the entire corpus, no source article can be present in training, so any improvement must reflect temporally-aligned knowledge and reasoning rather than recall.

\begin{table}[htbp]
    \centering
    \caption{Macro-F1 on a temporal holdout whose questions are drawn from pages published after the last training dump.}
    \label{tab:temporal-holdout}
    \small
    \setlength{\tabcolsep}{4pt}
    \begin{tabular}{l cccccc}
        \toprule
        & \llamathreeoneeightb & \llamathreetwooneb & \llamathreetwothreeb & \gemmathreeoneb & \olmotwooneb & \olmotwosevenb \\
        \midrule
        Base   & 0.531 & 0.477 & 0.493 & 0.494 & 0.504 & 0.463 \\
        CPT    & 0.524 & 0.500 & 0.542 & 0.522 & 0.521 & 0.522 \\
        $\Delta$ & $-0.007$ & $+0.023$ & $+0.049$ & $+0.028$ & $+0.017$ & $+0.059$ \\
        \bottomrule
    \end{tabular}
\end{table}

CPT improves the unseen split for five of six models (mean $+0.028$), with the model that was trained on the largest amount of data (\llamathreeoneeightb being flat) ($-0.007$). Importantly, the improvements are of comparable magnitude to the after-cutoff gains reported in Figure~\ref{fig:main-avg}, so the aggregate gains are not an artifact of overlap between the evaluation questions and the training corpus.

\section{LoRA replication on \llamathreetwothreeb}
\label{app:lora-llama}

The LoRA results in Section~\ref{sec:findings-lora} are obtained on \gemmathreeoneb. To check that the conclusion is not family-specific, we repeated the experiment on \llamathreetwothreeb, training with LoRA on the \texttt{score}$\geq\!4$ corpus at ranks $128$ and $256$ under otherwise identical settings. On before-cutoff questions all four models --- base, full CPT, LoRA-128 and LoRA-256 --- are indistinguishable, landing within $0.530$--$0.537$ Macro-F1. On after-cutoff questions the base model scores $0.484$, full CPT $0.540$, LoRA-128 $0.525$ and LoRA-256 $0.534$. Both LoRA variants therefore recover the large majority of the full-CPT acquisition gain, with the higher rank closer to full CPT, replicating the rank-dependent pattern of Figure~\ref{fig:metric-lora} on a second family and supporting the recommendation to use rank $\geq 128$ when memory-bound.

\section{\textsc{Qwen3} models and sequential updating}
\label{app:qwen}

\paragraph{Why \textsc{Qwen} is not in the main pool.} Our design is anchored on a documented knowledge cutoff: both the CPT corpus and the before/after question split are defined relative to that date. \textsc{Qwen} does not publish one, and an incorrect guess contaminates both sides at once --- post-cutoff training data that is in fact pre-cutoff, and after-cutoff questions the model has actually seen. We nonetheless ran the experiment, adopting 2023-12-31 as an assumed cutoff and performing \emph{no} learning-rate search (we reuse the setting selected for the comparable \llama models), so the numbers below should be read as a consistency check rather than as a tuned result.

\paragraph{\textsc{Qwen3} results.} CPT improves all three scales on both splits (Table~\ref{tab:qwen}), with the largest after-cutoff gain at 8B ($+0.05$). General capability is preserved at every scale: each model stays within $0.005$ of its base downstream average, matching the $\pm 0.01$ band reported for our main six models. Gains on questions extracted from articles absent from the training corpus follow suit (\textsc{Qwen3}-4B: $0.523 \to 0.539$; \textsc{Qwen3}-1.7B: $0.471 \to 0.531$), indicating that the improvement is not memorization. Across all three scales, then, CPT acquires post-cutoff knowledge, shows positive backward transfer on before-cutoff questions, and leaves general abilities intact, reproducing the picture from \olmotwo, \llama and \gemmathree on a fourth, independently developed family.

\begin{table}[htbp]
    \centering
    \caption{\textsc{Qwen3} under CPT (assumed cutoff 2023-12-31, no learning-rate search). Daily Oracle Macro-F1 by split, and macro-average accuracy over the thirteen downstream tasks.}
    \label{tab:qwen}
    \small
    \begin{tabular}{l cc cc cc}
        \toprule
        & \multicolumn{2}{c}{Before-cutoff F1} & \multicolumn{2}{c}{After-cutoff F1} & \multicolumn{2}{c}{Avg.\ downstream} \\
        \cmidrule(lr){2-3} \cmidrule(lr){4-5} \cmidrule(lr){6-7}
        Model & Base & CPT & Base & CPT & Base & CPT \\
        \midrule
        \textsc{Qwen3}-1.7B & 0.587 & 0.597 & 0.555 & 0.589 & 0.627 & 0.626 \\
        \textsc{Qwen3}-4B   & 0.639 & 0.657 & 0.614 & 0.642 & 0.696 & 0.694 \\
        \textsc{Qwen3}-8B   & 0.651 & 0.670 & 0.603 & 0.657 & 0.720 & 0.725 \\
        \bottomrule
    \end{tabular}
\end{table}

\paragraph{Sequential (incremental) updating.} The main study performs a single CPT round, whereas the realistic deployment setting is repeated updating. To test that setting directly, we split the post-cutoff data into three consecutive six-month eras (2024 H1, 2024 H2, 2025 H1) and continue-pretrained \llamathreetwothreeb sequentially, one era at a time, carrying the optimizer state and re-initializing across rounds. We evaluate every round on every era, so that forgetting on earlier slices and acquisition on later ones can be tracked as rounds accumulate (Table~\ref{tab:sequential}).

\begin{table}[htbp]
    \centering
    \caption{Sequential updating of \llamathreetwothreeb over three consecutive six-month eras (Macro-F1). Repeated updates neither compound forgetting nor fall behind a single bulk pass.}
    \label{tab:sequential}
    \small
    \begin{tabular}{l ccccc}
        \toprule
        Evaluation era & Base & 1-shot CPT & Seq.\ round I & Seq.\ round II & Seq.\ round III \\
        \midrule
        Before (2020--2023) & 0.536 & 0.538 & 0.552 & 0.550 & 0.535 \\
        2024 H1             & 0.482 & 0.509 & 0.506 & 0.512 & 0.503 \\
        2024 H2             & 0.505 & 0.554 & 0.535 & 0.538 & 0.545 \\
        2025 H1             & 0.507 & 0.591 & 0.566 & 0.576 & 0.589 \\
        \bottomrule
    \end{tabular}
\end{table}

Two points stand out. First, after three sequential rounds the before-cutoff score is essentially unchanged from base ($0.536 \to 0.535$) and every post-cutoff era remains above base, so the feared compounding of forgetting over rounds does not appear. Second, the final sequential model matches the one-shot CPT model: splitting the same data into three ordered updates reaches the same endpoint as a single bulk pass rather than degrading it. This is consistent with the explanation in Section~\ref{sec:findings-know-acq}, since per-snapshot re-crawl overlap keeps every round permanently in the replay regime.

\end{document}